\documentclass{article}

\usepackage{arxiv}

\usepackage[utf8]{inputenc} 
\usepackage[T1]{fontenc}    
\usepackage{hyperref}       
\usepackage{url}            
\usepackage{booktabs}       
\usepackage{amsfonts}       
\usepackage{amsmath}
\usepackage{amstext}
\usepackage{enumitem}
\usepackage{caption}
\usepackage{cite}
\usepackage{color}
\usepackage{colortbl}
\usepackage{epsfig}
\usepackage{eurosym} 
\usepackage{emptypage}
\usepackage{fancyhdr}
\usepackage{graphicx}
\usepackage{listings}
\usepackage{multirow}
\usepackage{microtype}      
\usepackage{nicefrac}       
\usepackage[intoc,prefix]{nomencl}
\usepackage{lipsum}
\usepackage{subcaption}
\usepackage{tabularx}
\usepackage{tikz,pgfplots,pgf}
\usetikzlibrary{matrix,shapes,arrows,positioning}
\usepackage{units}
\usepackage{upgreek}
\usepackage{microtype}      
\usepackage{wasysym}

\title{Identification of state functions by physically-guided neural networks with physically-meaningful internal layers}

\author{
Jacobo Ayensa-Jim\'enez \\
  Mechanical Engineering Department;\\
  Arag\'on Institute of Engineering Research (I3A);\\
  University of Zaragoza;\\
  Mariano Esquillor, S/N,  50018 Zaragoza, Spain. \\
  \texttt{jacoboaj@unizar.es} \\
\And
Mohamed H. Doweidar \\
  Mechanical Engineering Department;\\
  Escuela de Ingenier\'ia y Arquitectura (EINA);\\
  University of Zaragoza;\\
  Mar\'ia de Luna S/N, E. Betancourt, 50018 Zaragoza, Spain. \\
  \texttt{mohamed@unizar.es} \\
\And
Jose A. Sanz-Herrera \\
  Mechanical Engineering Department;\\
  School of Engineering;\\
  University of Sevilla;\\
  Camino de los Descubrimientos, S/N, 41092 Sevilla, Spain. \\
  \texttt{jsanz@us.es} \\
\And
Manuel Doblaré\thanks{Corresponding author} \\
  Mechanical Engineering Department;\\
  Arag\'on Institute of Engineering Research (I3A);\\
  University of Zaragoza;\\
  Mariano Esquillor, S/N,  50018 Zaragoza, Spain. \\
  \texttt{mdoblare@unizar.es} \\
}

\pgfplotsset{compat=1.16}
\begin{document}

\newcommand\norm[1]{\left\lVert#1\right\rVert}
\newcommand{\bs}[1]{\boldsymbol{#1}}
\newcommand{\ssf}[1]{\mathsf{#1}}
\tikzset{%
  every neuron/.style={
    circle,
    draw,
    minimum size=0.5cm
  },
  neuron missing/.style={
    draw=none, 
    scale=3,
    text height=0.2cm,
    execute at begin node=\color{black}$\vdots$
  },
}

\maketitle

\begin{abstract}
Substitution of well-grounded theoretical models by data-driven predictions is not as simple in engineering and sciences as it is in social and economic fields. Scientific problems suffer most times from paucity of data, while they may involve a large number of variables and parameters that interact in complex and non-stationary ways, obeying certain physical laws. Moreover, a physically-based model is not only useful for making predictions, but to gain knowledge by the interpretation of its structure, parameters, and mathematical properties. The solution to these shortcomings seems to be the seamless blending of the tremendous predictive power of the data-driven approach with the scientific consistency and interpretability of physically-based models.

We use here the concept of physically-constrained neural networks (PCNN) to predict the input-output relation in a physical system, while, at the same time fulfilling the physical constraints. With this goal, the internal hidden state variables of the system are associated with a set of internal neuron layers, whose values are constrained by known physical relations, as well as any additional knowledge on the system. Furthermore, when having enough data, it is possible to infer knowledge about the internal structure of the system and, if parameterized, to predict the state parameters for a particular input-output relation. We show that this approach, besides getting physically-based predictions, accelerates the training process, reduces the amount of data required to get similar accuracy, filters partly the intrinsic noise in the experimental data and provides improved extrapolation capacity.
\end{abstract}

\keywords{Physically Guided Neural Networks \and Explanatory Artificial Intelligence \and Data-Driven Simulation Based Engineering and Sciences \and Model Identification \and Parameter Identification \and NN prediction improvement}

\clearpage

\tableofcontents

\clearpage

\section{Introduction}
Science has progressed historically through the fruitful interaction between theory and experiments, or better, between hypotheses and data.  Additionally, since the appearance of computers in the fifties, and accelerated in the nineties of the XXst century, simulation has been progressively recognized as the third pillar of the scientific method \cite{skuse2019third}.

To describe a physical phenomenon, one states the mathematical equations that control the evolution of a set of variables (position, momentum, temperature, entropy, ...) that completely determine the state of the system. That evolution depends upon a set of external stimuli, that are assumed to be known, and upon the state itself. In this context, we distinguish between two kinds of equations: universal physical principles (conservation laws and physical inequalities), and the internal state equations that compile the averaged behavior of the system from its particular internal structure. The ability of any physico-mathematical model to accurately represent the reality is directly related to the quality of the simplification hypotheses that drive to those state equations and to the available experimental data required to identify the associated parameters.

This combination of universal physical principles and phenomenological state models under well-contrasted hypotheses has demonstrated to be highly effective to accurately predict the state and evolution of big and complex realistic problems, while keeping them mathematically tractable. 

A new paradigm is raising, however, based on our increasing ability to collect, store, analyze, and extract information from high volumes of data, a capability that is accelerating at an unprecedented rate \cite{Atzori2010,Manyika2011}. Based on the success of Data Science and Artificial Intelligence in fields like e-commerce \cite{Hill2006}, social sciences \cite{Aneshensel2013}, healthcare \cite{Raghupathi2014}, language recognition \cite{Lopez-Moreno2014}, image-based predictions \cite{Krizhevsky2012}, etc., they are also gaining prominence in simulation-based engineering and sciences (SBES).

However, data gathering in Physics is soaked by centuries of scientific knowledge and the associated human bias \cite{Berry2011, leonelli2012introduction, Gould1981, Kitchin2014}; so, a ``blind" algorithm without any information on that bias may lead to wrong predictions. Also, scientific problems suffer most times from paucity of data while involving a large number of variables that interact in complex and non-stationary ways. Therefore, we can expect poor predictive capability of purely data-based approaches in problems far from the training set. Finally, a physically-based model is not only useful for making predictions, but it is expected to help in gaining knowledge by the interpretation of its structure, parameters, and mathematical properties. In fact, physical interpretability is, in many cases, at least as important as predictive performance. It is not strange therefore the important efforts made in ``whitening" the ``black box" way of working of current machine-learning predictive algorithms \cite{Shwartz2017}. 

One possible solution to this shortcoming of data-only models is the seamless blending of their tremendous predictive power with the scientific consistency and interpretability of physically-based models. The term coined for this hybrid paradigm is physically-guided data science (PGDS) \cite{ayensa2019unsupervised, Karpatne2017, raissi2017physics, raissi2019physics, lu2019data}.  A straightforward application of PGDS techniques is dynamic data-driven systems (DDSBES) \cite{Darema2004, Schmidt2009, Brunton2015, Gonzalez2016, Kirchdoerfer2016, Ladeveze1989, Peherstorfer2015}. However, one of the most important drawbacks in current data-driven approaches is the need of explicitly defining the cloud of experimental values that identifies the internal state model (e.g. material constitutive equations in solid mechanics) with a sufficient number of points in the whole range of interest. This forces to perform extensive experimental campaigns that are costly in time and money and whose results rely on strong assumptions on the experimental model itself (e.g. uniform distribution of stresses in uniaxial tests), that cannot be overcome due to the non-observable (non-measurable) character of some of such state variables (e.g. stresses).

An opposite perspective is integrating physical knowledge into data science models, that is, to constrain the prediction domain of the standard data model by physical constraints. Of course, this approach may be extended to any known relation between the input-output variables. One simple example of this addition of physical knowledge was made in \cite{ayensa2019unsupervised} to fill incomplete data sets coming from experimental campaigns in a reliable way. Another more powerful approach is using physical knowledge to inform and improve the data prediction capability of neural networks \cite{karpatne2017physics, raissi2017physics, raissi2019physics, lu2019data, Karpatne2017}. However, in all these works, the physical information was introduced directly as relations between the input and output layers. Only in \cite{raissi2019physics, lu2019data} a first attempt was made to provide the network with some explanatory capacity by adding some of the parameters associated with the internal state model as output.

In this paper, we extend such explanatory capability by establishing a general approach in which we distinguish between the universal physical laws and the internal state equations. The former are treated as constraints imposed by the particular Physics between neuron values in the NN, with an appropriate topology. The latter are derived as a direct NN outcome. This will permit to identify some of the internal neurons with internal (in general non-observable) state variables. The objective of this work is therefore to introduce this new methodology of \emph{Physically Guided Neural Networks with Internal Variables} (PGNNIV) to predict both the input-output relation in a physical system from a sufficient set of data, as well as to infer knowledge on the system internal structure, but always considering the constraints imposed by Physics. The corresponding NN are trained by only observable data (e.g. displacements or velocities, forces, etc.), getting the internal non-observable (e.g. stresses) values from the NN output. 

We show that this methodology shows a better performance in terms of faster convergence, less need of data, data noise filtering and bias correction, and extrapolation capacity. 

\section{Physically-Guided Neural Networks with Internal Variables. Concept, formulation and types of applications} \label{secc::PGNNIV}

\subsection{General framework} \label{sssecc::framework}

We identify immediately two types of state variables in any averaged phenomenological theory: i) observable (measurable) variables that can be local such as the position, pressure or  temperature or integral as energy variations. These will be denoted as $\bs u$, and collect all possible \textbf{essential variables} of the problem, that is, the minimum set of independent, in general spatial and time-dependent variables that define the observable state of the system; ii) \textbf{internal variables} (not always directly measurable), that are model-specific. In general, these internal state variables collect the changes in the internal structure of the system that may be reversible or irreversible (examples are stresses, plastic or viscous strains, friction forces, damage and, in general variables associated with energy dissipation mechanisms). They will be denoted as $\bs{\eta}$.

In the same way, we stated already that there are two types of equations: i) \textbf{universal physical laws}, valid for any problem in a certain context (e.g. non-relativistic mechanics), such as conservation of mass, linear and angular momenta and energy. They define the time evolution or the static state of the system such as $\dot{\bs u} = \bs G(\bs u, \bs{\eta}, \bs f)$ ($\bs G(\bs u, \bs{\eta}, \bs f)=\bs 0$ in the non-transient case), with $\bs G$ a set of functions, universal for a particular family of problems (e.g. Continuum Mechanics) and $\bs f$ the external stimuli assumed to be known; ii) \textbf{state equations} that define the averaged evolution of the internal variables in terms of the current state of the system ($\bs u$ and $\bs \eta$) and a set of internal parameters $\bs \uplambda$, $\dot {\bs \eta} = \bs H(\bs \eta, \bs u, \bs \lambda)$ ($\bs \eta = \bs H(\bs u, \bs \lambda)$ in the non-transient case, redefining $\boldsymbol{H}$ if necessary). These equations are most times phenomenological, so they, and the associated parameters, have to be determined and validated from reasonable assumptions and experimental tests, being these latter, as commented, one of the main bottlenecks to model completely a physical system.

\subsection{Physically-Guided Neural Networks with Internal Variables} \label{ssecc::formulation}

\subsubsection{Mathematical formulation} \label{sssecc::math}

To predict the value of the essential variables, we define the architecture of our PGNNIV according to the following recipe:  

\begin{enumerate}
\item We identify the output layer with (some or all) the values of the essential variables, $\bs u$, or some directly related quantities of interest. 
\item The input layer corresponds to known values such as the external stimuli $\bs f$ including the associated boundary conditions. Conversely, we can choose as input the essential variables and as output the stimuli, following a similar approach.
\item Some predefined internal layers (PILs) are associated with the internal state variables $\bs \eta$. Of course, when required, the values of such neurons may be recovered after convergence, getting the values of the internal state variables from the solution of the system for a particular input. The difference between PILs and common internal layers is that mathematical constraints are applied to the neurons of the former driving the learning process.
\item The rest of internal layers, connecting the input, output and PILs, follows the standard approach in NN, that is, they are enough in number, size and with appropriate activation functions to "discover" the complex relations hidden in the function $\bs H$.
\item Finally,  the universal laws stated in $\bs G$, are established in the NN as constraints between layers. 
\end{enumerate}

Any additional physical knowledge of the system that may be expressed in mathematical terms relating state variables and stimuli can be imposed in a similar way. For example, physical inequalities such as the second principle of thermodynamics or any inequality constraint on the parameters (e.g. positivity of the elastic modulus or the physical range for Poisson coefficient in an isotropic material $-1 < \nu < 0,5$). Moreover, we can further supply partial or total information about the internal state model by defining a parametric state equation $
\bs \eta = \bs H(\bs u;\bs \lambda)$ (including the particular case of an explicit state model $\bs{\eta} = \bs H(\bs u)$), by means of an appropriate topology of the state model network. In those cases, additional constraints or conditions may be designed to include this knowledge.

As a result of the learning process, the relationship $\bs \eta  = \bs H(\bs u)$ is learned and so it is the input-output relationship. Therefore, Physically-Guided Neural Networks with Internal Variables have both predictive and explanatory capacity.

It is important to note that this representation opens a new paradigm in the characterization of a given constitutive model. The model is not anymore characterized by an explicit functional relationship $\boldsymbol{\eta} = \bs H(\bs u;\boldsymbol \lambda)$ depending on the field variables and additional parameters that have to be determined by classical fitting of a set of experimental tests (\emph{functional framework}), but by a neural network topology and metaparameters $\boldsymbol{\eta} = \mathsf{H}(\bs u) = \bs H(\bs u;\bs W)$, where $\bs W$ represents the weights and biases of the neural network that are tuned during the training process. The use of the sans serif notation for a functional dependence ($\bs{y} = \mathsf{Y}(\bs x)$, $\bs{\eta} = \mathsf{H}(\bs u)$) is here used to represent an implicit input-output relation in a NN. Indeed, the universal approximation theorem guarantees that a regular enough function may be approximated by a specific neural network with a sufficient number of layers and neurons and convenient activation functions \cite{cybenko1989approximations,hornik1991approximation,lu2017expressive,hanin2017universal}, so this second approach is, at least, as general as the first one, also unfolding the benefits of the neural network hardware (fast computation with GPU and TPU, cloud and distributed computing...) and software such as Keras and TensorFlow (modularity, plugability, fast generalization capability,...). All this allows for high performance computing capabilities and scalability \cite{strigl2010performance}, for major model flexibility that allows capturing strong non-linearities \cite{lee1993testing,sarle1994neural} and for soundness with respect to statistical data (heteroskedasticity, non-normality...) \cite{guh2002robustness,mcaleer2008neural,matias2010boosting}. 

\textbf{Remark 1:} We have to remark that these variables may be spatial and/or time fields depending on the location $\bs x$ and/or time $t$, being then $\bs G, \bs H$, respective functionals. In that cases, we shall consider that a previous discretization step has been applied so the time-position independent interpolating variables are the ones of the associated discretized problem. Therefore, the same approach can be used both for non-transient problems or for transient ones, using as variables $\bs u$ those algebraic values that define the approximated field at a certain time and interpolation point, following a step by step continuation approach. Of course, if some of the internal state variables depend on the history, this time-discretization is always required even if the problem is time-independent (e.g. rate-independent plasticity) using a pseudo-time during the loading-unloading process. 

\textbf{Remark 2:} It is also important to note that, if no constraint is applied and no PIL is defined, we recover the classical Neural Network framework. If the constraints are applied to the input or output layers, we recover the formulation developed by other authors for Physically-Guided Neural Networks \cite{Karpatne2017, raissi2019physics}. A similar approach to the one here presented was also addressed in \cite{raissi2019physics} for partial differential equations, but without using the PIL concept that is original, up to the authors' knowledge.

\textbf{Remark 3:} This framework allows the scientist to work only with directly measurable variables and fields, without the need of establishing any \emph{a priori} assumption on the expression of the internal variables, which is fundamental, since internal (non-measurable) variables are, indeed, mathematical constructs that are now determined as byproduct of the predictive problem.

\subsubsection{Construction of Physically-Guided Neural Networks with Internal Variables} \label{sssecc::comp}

Denoting the input variable $\mathbf{x}$ as $\mathbf{y}_0$, each hidden layer of $n_i$ neurons, $\mathbf{y}_i$,  $i=1,\cdots,L$ is defined by a functional relation:

\begin{equation}
\bs{y}_i = \phi(\bs{y}_{i-1}\bs{W}_i+\bs{b}_i).
\end{equation}
where $\bs{W}_i$ and $\bs{b}_i$, $i=1,\cdots,L$ are the weights and biases, the parameters of the model, and $\phi : \mathbb{R}^{n_i} \rightarrow \mathbb{R}^{n_i}$ is an activation function. With this notation, the output variable is $\bs{y} = \bs{y}_L$. The network is symbolically represented by the relationship $\bs{y} = \mathsf{Y}(\bs{x})$ or, denoting by $\bs{W}$ the whole set of weights and biases for a given network topology, $\bs{y} = \bs Y(\bs{x};\bs{W})$. Given a set of \emph{ground truth} data points $\mathcal{D} = \{(\bar{\bs{x}}^i,\bar{\bs{y}}^i)| i=1,\cdots,N\}$, a quadratic mean error is used to evaluate the network performance, $\mathrm{MSE}(\bs{W}|\mathcal{D}) = \frac{1}{N}\sum_{i=1}^N\norm{\bar{\bs{y}}^i-Y(\bar{\bs{x}}^i;\bs{W})}^2$.

Now, it is possible to train the neural network by minimizing the function $\mathrm{MSE}$, getting the optimal set of weights and biases $\bs W$. Let us now impose some constraints such that some neuron values satisfy some (physically-based) equations, including universal laws, manifold constraints and boundary conditions. Without loss of generality, we denote all these functions by $R_j, j=1,\ldots,r$.
\begin{equation}
\label{eq::constraints_R}
\bs R_j(\bs y_0,\ldots,\bs y_L) = \bs{0}, \quad j=1,\ldots,r.
\end{equation}

Given a ground truth set $\mathcal{D}$ that is considered to be physically based, we can reformulate the minimization problem as:
\begin{equation}
\label{eq::minimization_constrained_1}
\underset{\bs W}{\text{min}} \quad \mathrm{MSE}(\bs W |\mathcal{D}) \qquad  \text{s.t.} \qquad   R_j(\bs W |\mathcal{D}) = 0, \quad j=1,\ldots,r.
\end{equation}

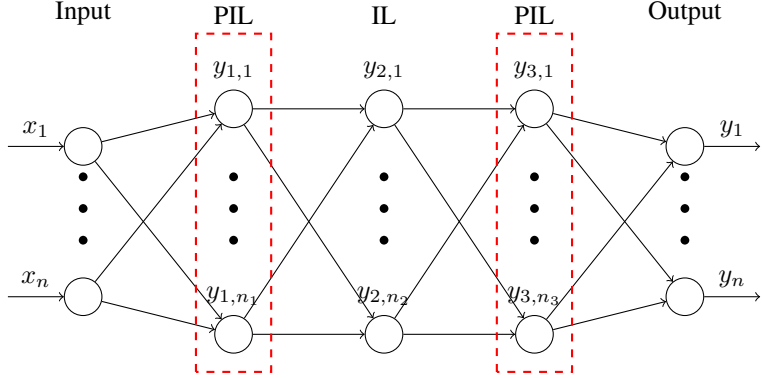
\begin{figure}
\centering
\begin{tikzpicture}

\foreach \m/\l [count=\y] in {1,missing,2}
  \node [every neuron/.try, neuron \m/.try] (input-\m) at (0,2-\y*1) {};
\foreach \m [count=\y] in {1,missing,2}
  \node [every neuron/.try, neuron \m/.try] (hidden_a-\m) at (2,3-\y*1.5) {};
\foreach \m [count=\y] in {1,missing,2}
  \node [every neuron/.try, neuron \m/.try] (hidden_b-\m) at (4,3-\y*1.5) {};
\foreach \m [count=\y] in {1,missing,2}
  \node [every neuron/.try, neuron \m/.try] (hidden_c-\m) at (6,3-\y*1.5) {}; 
\foreach \m [count=\y] in {1,missing,2}
  \node [every neuron/.try, neuron \m/.try] (output-\m) at (8,2-\y*1) {};

\foreach \l [count=\i] in {1}
  \draw [<-] (input-\l) -- ++(-1,0)
    node [above, midway] {$x_\l$};
\foreach \l [count=\i] in {2}
  \draw [<-] (input-\l) -- ++(-1,0)
    node [above, midway] {$x_n$};

\foreach \l [count=\i] in {1}
  \node [above] at (hidden_a-\i.north) {$y_{1,\l}$};
\foreach \l [count=\i] in {2}
  \node [above] at (hidden_a-\l.north) {$y_{1,n_1}$};
\foreach \l [count=\i] in {1}
  \node [above] at (hidden_b-\i.north) {$y_{2,\l}$};
\foreach \l [count=\i] in {2}
  \node [above] at (hidden_b-\l.north) {$y_{2,n_2}$};
\foreach \l [count=\i] in {1}
  \node [above] at (hidden_c-\i.north) {$y_{3,\l}$};
\foreach \l [count=\i] in {2}
  \node [above] at (hidden_c-\l.north) {$y_{3,n_3}$};

\foreach \l [count=\i] in {1}
  \draw [->] (output-\i) -- ++(1,0)
    node [above, midway] {$y_\i$};
\foreach \l [count=\i] in {2}
  \draw [->] (output-\l) -- ++(1,0)
    node [above, midway] {$y_n$};

\foreach \i in {1,2}
  \foreach \j in {1,2}
    \draw [->] (input-\i) -- (hidden_a-\j);   
\foreach \i in {1,2}
  \foreach \j in {1,2}
    \draw [->] (hidden_a-\i) -- (hidden_b-\j);
\foreach \i in {1,2}
  \foreach \j in {1,2}
    \draw [->] (hidden_b-\i) -- (hidden_c-\j);
\foreach \i in {1,2}
  \foreach \j in {1,2}
    \draw [->] (hidden_c-\i) -- (output-\j);

\draw[red,thick,dashed]  (1.5,-2) rectangle (2.5,2.5);
\draw[red,thick,dashed]  (5.5,-2) rectangle (6.5,2.5);

\foreach \l [count=\x from 0] in {Input, PIL, IL, PIL, Output}
  \node [align=center, above] at (\x*2,2.5) {\l};
\end{tikzpicture}
\caption{\textbf{Physically-Guided Neural Network for a three hidden-layered network.} The red dashed rectangles indicate the neurons in which a certain constraint is applied (PILs). The number of internal layers between the input layer and layer 1, layer 1 and layer 3 and between layer 3 and the output layer can be increased to allow more complex models. It is the physical constraint the one that provides the PILs $1$ and $3$ a physical interpretation as state variables.}
\label{fig_ConModel}
\end{figure}

Fig. \ref{fig_ConModel} illustrates a Physically-Guided Neural Network for a three hidden-layered neural network.

It is then possible to reformulate (\ref{eq::minimization_constrained_1}) using a penalty approach. Defining $r$ penalty parameters $p_j$, the problem is then expressed as:

\begin{equation}
\label{eq::minimization_constrained_3}
\begin{aligned}
& \underset{\bs{W}}{\text{min}}
& \mathrm{MSE}(\bs{W}|\mathcal{D})+\sum_{j=1}^r p_j \norm{\bs R_j(\bs{W}|\mathcal{D})}^2\\
\end{aligned}
\end{equation}

This approach allows the implementation of the problem in a standard Neural-Network framework (i.e. TensorFlow@Python) by just defining an adapted loss function that includes the penalty term $\mathrm{OF} = \mathrm{MSE} + \mathrm{PEN}$ with $\mathrm{PEN}(\bs{W}) = \sum_{j=1}^r p_j \norm{R_j(\bs{W}|\mathcal{D})}^2$.

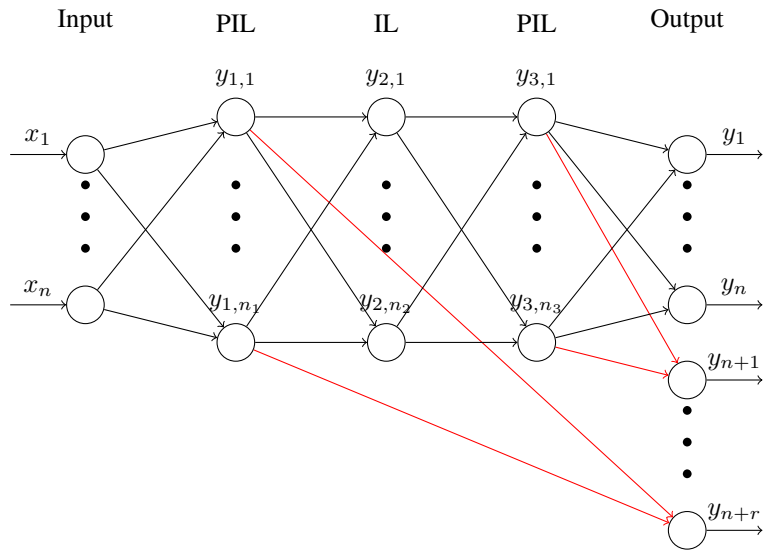
\begin{figure}
\centering
\begin{tikzpicture}

\foreach \m [count=\y] in {1,missing,2,3,missing,4}
  \node [every neuron/.try, neuron \m/.try] (output-\m) at (8,2-\y*1) {};
\foreach \i in {1,2}
  \foreach \j in {4}
    \draw [->,red] (hidden_a-\i) -- (output-\j);
\foreach \i in {1,2}
  \foreach \j in {3}
    \draw [->,red] (hidden_c-\i) -- (output-\j);
    
\foreach \m/\l [count=\y] in {1,missing,2}
  \node [every neuron/.try, neuron \m/.try] (input-\m) at (0,2-\y*1) {};
\foreach \m [count=\y] in {1,missing,2}
  \node [every neuron/.try, neuron \m/.try] (hidden_a-\m) at (2,3-\y*1.5) {};
\foreach \m [count=\y] in {1,missing,2}
  \node [every neuron/.try, neuron \m/.try] (hidden_b-\m) at (4,3-\y*1.5) {};
\foreach \m [count=\y] in {1,missing,2}
  \node [every neuron/.try, neuron \m/.try] (hidden_c-\m) at (6,3-\y*1.5) {}; 

\foreach \l [count=\i] in {1}
  \draw [<-] (input-\l) -- ++(-1,0)
    node [above, midway] {$x_\l$};
\foreach \l [count=\i] in {2}
  \draw [<-] (input-\l) -- ++(-1,0)
    node [above, midway] {$x_n$};

\foreach \l [count=\i] in {1}
  \node [above] at (hidden_a-\i.north) {$y_{1,\l}$};
\foreach \l [count=\i] in {2}
  \node [above] at (hidden_a-\l.north) {$y_{1,n_1}$};
\foreach \l [count=\i] in {1}
  \node [above] at (hidden_b-\i.north) {$y_{2,\l}$};
\foreach \l [count=\i] in {2}
  \node [above] at (hidden_b-\l.north) {$y_{2,n_2}$};
\foreach \l [count=\i] in {1}
  \node [above] at (hidden_c-\i.north) {$y_{3,\l}$};
\foreach \l [count=\i] in {2}
  \node [above] at (hidden_c-\l.north) {$y_{3,n_3}$};

\foreach \l [count=\i] in {1}
  \draw [->] (output-\i) -- ++(1,0)
    node [above, midway] {$y_\i$};
\foreach \l [count=\i] in {2}
  \draw [->] (output-\l) -- ++(1,0)
    node [above, midway] {$y_n$};
\foreach \l [count=\i] in {3}
  \draw [->] (output-\l) -- ++(1,0)
    node [above, midway] {$y_{n+1}$};
\foreach \l [count=\i] in {4}
  \draw [->] (output-\l) -- ++(1,0)
    node [above, midway] {$y_{n+r}$};

\foreach \i in {1,2}
  \foreach \j in {1,2}
    \draw [->] (input-\i) -- (hidden_a-\j);   
\foreach \i in {1,2}
  \foreach \j in {1,2}
    \draw [->] (hidden_a-\i) -- (hidden_b-\j);
\foreach \i in {1,2}
  \foreach \j in {1,2}
    \draw [->] (hidden_b-\i) -- (hidden_c-\j);
\foreach \i in {1,2}
  \foreach \j in {1,2}
    \draw [->] (hidden_c-\i) -- (output-\j);

\foreach \l [count=\x from 0] in {Input, PIL, IL, PIL, Output}
  \node [align=center, above] at (\x*2,2.5) {\l};
\end{tikzpicture}
\caption{\textbf{Augmented Neural Network equivalent to the Physically-Guided Neural Network.} Each constraint is replaced by an extra output representing the value of the constraint that, ideally, should be null.}
\label{fig::network_two}
\end{figure}

Note that (\ref{eq::minimization_constrained_3}) may be interpreted as an auxiliary neural network with input $\bs{x}$ and output $\hat{\bs{y}} = (\bs{y}; \bs{y}_P)$, being $\bs{y}_P$ a new set of output variables, $\bs{y}_P = (\bs R_1,\cdots,\bs R_r)$  with physical meaning, whose \emph{ground truth} value is always $0$, that is $\bar{\bs R_i} = \bs 0$. This latter neural network may be seen as the output augmentation of the former, with new connections between the hidden layers and the output variables, as  illustrated in Figure \ref{fig::network_two}. Conversely, a classical neural network is recovered when $\bs R_j(\bs{x};\bs{W}) = \bs{0}$.

\textbf{Remark 1:} This approach has two main advantages that match the two spearheads against Artificial Neural Networks (ANN) methods:

\begin{enumerate}
\item From a physical point of view, we postulate some extra conditions onto the hidden variables, which allow to interpret them as true physically-based features, that is, as state variables of the physical problem. This tries to overcome the \emph{black-box} problem of neural networks \cite{castelvecchi2016can,papernot2017practical,samek2017explainable}. \cite{castelvecchi2016can,papernot2017practical,samek2017explainable}.
\item As the search space is reduced via constraints, the optimization algorithm is expected to learn faster, with less information, to filter the noise incompatible with the problem physics and to discard solutions without physical sense. This could bring us convergence acceleration, less data needs, extra filtering capacities and physical correction when extrapolating. These improvements will be evaluated later.
\item Of course, if some of the constraints are expressed by an explicit equation, we can modify the problem (\ref{eq::minimization_constrained_1}) to make this constraint to disappear from the general formulation.
\end{enumerate}

\textbf{Remark 2:} The inclusion of inequalities in the presented framework is possible using the ReLU function. Indeed, the inclusion of a term in the penalty function with the structure $p\mathrm{ReLU}(f(\bs{W},\mathcal{D}))$ guarantees that if $p$ is high enough, $\mathrm{ReLU}(f(\bs{W},\mathcal{D}))$ has to be the smallest possible, ensuring that $\mathrm{ReLU}(f(\bs{W},\mathcal{D})) \rightarrow 0$ and therefore $f(\bs{W},\mathcal{D}) \leq 0$.

\clearpage

\subsection{Types of problems in which Physically-Guided Neural Networks with Internal Variables may be applied} \label{ssecc::types}

We can think of several families of problems, namely:

\begin{enumerate}
\item \textbf{Prediction problems:} The goal is to predict the value of a set of dependent output variables $\bs y$  from other independent measurable ones $\bs x$. The material constitutive model or state equation is assumed to be frozen and therefore there exists an (unknown) relationship, $\bs H$, whose functional form or properties have to be revealed. Usually $\bs y$ is identified with the essential variable $\bs u$, that corresponds to the solution field (continuum) or variable (discrete) in most physical problems and $\bs x$ corresponds to the values of the initial and boundary conditions ($\bs f^*$), together with the external stimuli ($\bs f$) (direct problem). However, these roles may be interchanged and the aim may be to obtain $\bs f$ from $\bs u$ after appropriate training (inverse problem). Consequently, the state equation is context-dependent and may be formulated as $\bs \eta = \bs H(\bs y)$ or $\bs \eta = \bs H(\bs x)$. We will use the first notation in the subsequent. Mixed formulations can be defined between the direct and the inverse problems, assuming the problem is well-posed.

When solving prediction problems, different objectives may be followed:

\begin{enumerate}

\item \textbf{Pure predictive problem}. We establish a direct correspondence between observable variables for a fixed system (same geometry and internal structure), without any constraint nor explicit establishment of PILs. In this case, we do not impose any constraint and use the NN in the standard black-box manner to get the correlation between the input and output variables to predict, after training, the latter for a particular input, without any knowledge of the physical system. The objective is, therefore, to solve with the NN the implicit relation $G(\bs u, \bs \eta(\bs{u}), \bs{f})$ as:

\begin{equation}
\bs y = \mathsf{Y}(\bs x) \\
\end{equation}
where $\mathsf{Y}$ is the representation of some appropriate neural network (architecture, connectivity and metaparameters). This has been done frequently in the last decades \cite{thibault1991neural,hambli2006real,pathak2005application}.

\item \textbf{Pure predictive problem with input-output constraints}. The only difference with the above is the assumption (or knowledge) of some input-output relations. Those relations are imposed via external constraints onto the objective function of the NN without using the concept of internal state variables or PILs.

\begin{align}
\bs y &= \mathsf{Y}(\bs x) \nonumber \\
& \text{s. t.} \; \;  \bs R(\bs{y}, \bs H(\bs{y}),\bs{x}) = \bs{0} 
\end{align}
being $\bs R$ a prescribed functions that acts between the input and output layers and $\bs H$ a model that is assumed implicitly. Some particular problems have been solved very recently using this methodology \cite{karpatne2017physics,raissi2019physics}.

\item \textbf{Predictive and explanatory problem with internal hidden variables and constraints}. We add, instead, a PIL with the physical meaning of internal state non-observable variables  and include, explicitly, the physical laws between them and the directly observable external stimuli and the boundary conditions by constraints in the NN, thus helping the NN to "know" that the system should fulfill such conservation laws. The unknown constitutive model is then obtained as an implicit relation between such internal state variables and the input ones.

\begin{align}
\bs{y} &= \mathsf{Y}(\bs x, \bs \eta); \;\;\; \bs{\eta} = \mathsf{H}(\bs{y}) \nonumber \\
& \text{s. t.} \; \;  \bs{R}(\bs y, \bs \eta, \bs x) = \bs{0} 
\end{align}
where $\bs R $ is a function encoding the universal principles driving the problem physics and geometry associated to the ambient space. The network learning capability is now located at the neural relations $\bs y = \mathsf{Y}(\bs x, \bs \eta )$ (predictive network) and $\bs{\eta} = \mathsf{H}(\bs{y})$ (explanatory network).

\item \textbf{Predictive and explanatory problem to identify fixed internal parameters}. We add to the previous problem additional information on the structure of the constitutive model via new constraints between PILs, adding (or not) information on the values or ranges of the constitutive parameters. In this case, it is possible to ask the NN to predict the particular values of the parameters for a certain predefined explicit constitutive model structure if enough training data are provided. In this family of problems, the model has to be postulated (partially or totally) \emph{a priori} and, if some of the parameters are known, the rest are obtained as output, reaching some explanatory capacity. We have thus the same mathematical formulation as in the previous case except for the fact that the network relationship $\bs \eta  = \mathsf{H}(\bs{y})$ is not general, but some material symmetries and properties are postulated. One possible way is to define $\mathsf{H}(\bs{y}) = \bs H(\bs{y};\bs{\lambda})$ where $\boldsymbol \lambda$ are parameters characterizing the state equation/constitutive model that have to be learned. After the training process, we obtain estimated values for these physical parameters $\boldsymbol \lambda$ relative to the state equation. As a particular case, the model may be perfectly defined, $\mathsf{H}(\bs{y}) = \bs H(\bs{y})$, so the network has no explanatory capacity, only predictive. In that case, the methodology presented may be seen as a pure dimensionality reduction technique or an offline calculator (response surface) for posterior real time evaluations. The model is therefore formulated as:

\begin{align}
\bs y &= \mathsf{Y}(\bs x); \;\;\; \bs{\eta} = \bs H(\bs{y};\boldsymbol \lambda) \nonumber \\
& \text{s. t.} \; \;  \bs R(\bs y, \bs \eta, \bs x) = \bs{0} 
\end{align}

When the constraints are applied directly to the input-output layers we have 

\begin{align}
\bs y &= \mathsf{Y}(\bs x) \nonumber \\
& \text{s. t.} \; \;  \bs R(\bs y,\bs H(\bs y;\bs \lambda),\bs x) = \bs{0} 
\end{align}
This approach has already been explored by some authors but always in the latter form, which limits it to using only measurable variables \cite{raissi2019physics}.

\item \textbf{Model selection}. The idea is to test a set of potential constitutive models as in the previous case, getting the optimal parameters for each of them and then identifying the most likely of them as such with the lowest loss function. The mathematical formulation is, therefore:
\begin{align}
\bs y &= \mathsf{Y}(\bs x); \nonumber \\
& \text{s. t.}\; \;  \bs R(\bs y, \bs H(\bs y;\boldsymbol \lambda), \bs x) = \bs{0} \quad \bs H \in \mathcal{C} 
\end{align}

Here $\mathcal{C}$ denotes \emph{a catalog} of models. The one finally selected will be the one showing the best performance in terms of the error function. As commented, a specific case is when $\bs H$ is totally specified, that is, $\boldsymbol \lambda$ are not learned from the data. These two later examples are similar to classical model regression, but using NN software, hardware and methods, with their advantages in terms of computational cost and distributed computation.

\end{enumerate}
\item \textbf{Characterization problems}: The goal here is to characterize the parameters of a pre-established constitutive model or state equation for different macroscopic materials. The material constitutive model or parameters are assumed to vary from one training data to another. Therefore, for the problem to make sense, enough physical information about the material response must be provided. The inputs are, consequently, the stimulus and the response of the material, $\mathbf{x}= (\mathbf{u},\mathbf{f})$. The output are variables related to the state model. Two classes of descriptors may be provided as output variables. 
\begin{itemize}
\item \textbf{Structural descriptors:} Any functional descriptor of the model characteristics (e.g. spatial homogeneity or time invariance, anisotropy or symmetries, linearity, memoryless, damage accumulation, ...). In general, this can be addressed, from a theoretical point of view, by using Lie Group theories and Noether's theorem. As an illustrative example, the spatial covariance of the elasticity tensor is an estimator of the material heterogeneity, and the autocorrelation function is an estimator of time invariance.  In that case, the output has to be defined accordingly to these considerations. This type of problems is out of the scope of this paper.
\item \textbf{Prescribed model parameters:} parameters appearing in the mathematical expression of a given empirical equation. For instance, the Young modulus $E$, and the Poisson ratio $\nu$, the particle attraction $a$ and the mole size exclusion parameter $b$ in the Van Der Waals equation.
\end{itemize} 

In this second family, we can distinguish the same types of problem than in the pure predictive one, with analog mathematical formulations, but the particular state model, defined by an appropriate subset of the complete NN or by a set of parameters associated with an \emph{a priori} defined parametric model, are also part of the output, considering that the training set includes a sufficiently complete, in number and variety, dataset derived from problems with different state models within the family of interest.
\end{enumerate}

These types of problem may be combined in mixed ones. For example, we can include a PIL associated to gradients of the essential variable (e.g. strains in Mechanics) with the corresponding defining constraint with the input/output layer (displacements) and another one associated to internal hidden variables (stresses) constrained by the conservation law (conservation of linear momentum). These two layers are connected by a subset of the whole NN that identifies the state model defined between both (stress-strain constitutive model).

\clearpage

\section{Examples of application} \label{secc::App}

Next, we consider several examples to illustrate the methodology and the different types of applications. We present the two approaches described: prediction and characterization. The fundamental results are shown, while different aspects about the method performance will be delayed to the Discussion.

\subsection{Problem statement}

\begin{figure}
\centering
\includegraphics[clip=true,trim=100pt 180pt 300pt 80pt,width=0.8\textwidth]{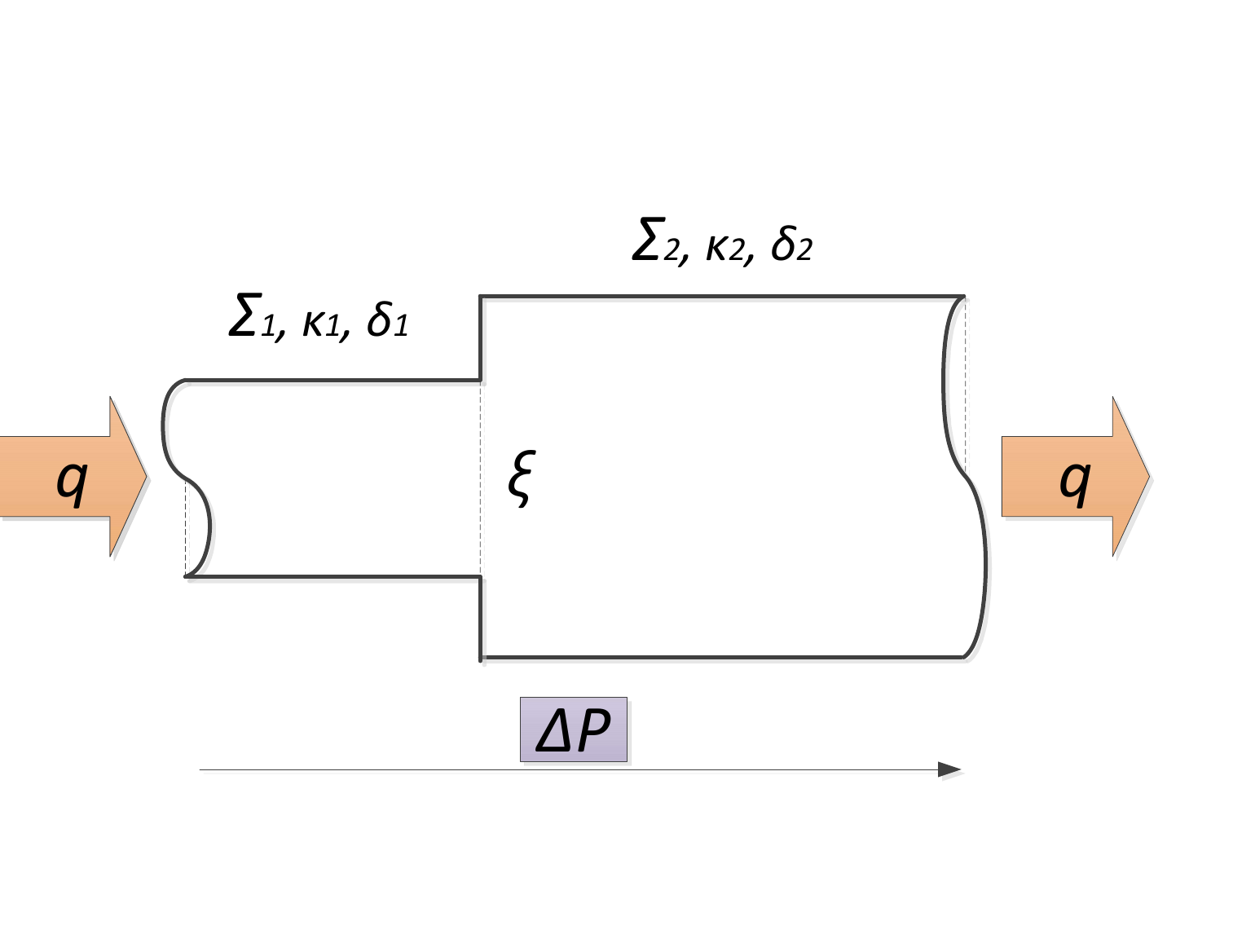}
\caption{\textbf{Scheme of the illustrative example.} Hydraulic head loss along a piecewise cross-section pipe.}
\label{fig::pipe}
\end{figure}

Let us consider a pipe segment of length $\delta_1$, with circular cross section with diameter $D_1$, and a sudden change in its circular cross-section to another segment of length $\delta_2$, with the same shape but with bigger diameter $D_2$ (Fig. \ref{fig::pipe}). The initial objective is to compute the head pressure loss, $\Delta p$, along the length of the whole pipe in the steady state regime, assuming fluid incompressibility. Using the Bernoulli equation, along a streamline, the hydraulic head is defined as $h = \frac{v^2}{2g} + z  + \frac{p}{\gamma}$, with $\gamma = \rho g$, being $\rho$ the density of the fluid, $g$ the gravity acceleration, $z$ the elevation and $\Delta h$ the hydraulic head loss due to eddy dissipation and wall friction that corresponds to the state equation of the problem, so it has to be characterized by means of: i) additional assumptions, usually with poor accuracy; ii) experimental tests with the corresponding fitted phenomenological equations; iii) simulations with complex fluid flow models. Here, we shall assume that the hydraulic head loss is associated with two physical phenomena: i) viscous dissipation distributed along the pipe and ii) localized dissipation at the pipe expansion. 
 
For the first physical phenomenon, we have distributed losses along a streamline, so assuming the flow as unidimensional and considering a horizontal pipe, we can write:
\begin{equation}
\label{eq::Bernouilli_2}
\frac{d}{dx}\left(\frac{v^2}{2g} + \frac{p}{\gamma}\right)=\frac{dh}{dx}=-i
\end{equation}
with $i$ the hydraulic head slope. 

It is common to express the hydraulic head slope in terms of the fluid velocity by using the Darcy-Weisbach expression \cite{weisbach1845lehrbuch}:
\begin{equation}
i = f_D\frac{1}{2g}\frac{v^2}{\Phi}
\end{equation}
where $\Phi$ is the hydraulic diameter of the pipe and $f_D$ the Darcy friction factor, an empirical coefficient that, again, is determined by additional hypotheses or semi-empirical expressions. Here, for illustrative purposes, we adopt a common empirical model: the Hazen-Williams expression for the hydraulic head slope that considers the fluid viscosity and the pipe roughness simultaneously \cite{williams1908hydraulic}:
\begin{equation}
i = \lambda\left(\frac{q}{\kappa}\right)^\alpha\Phi^\beta
\end{equation}
with $q=\Sigma v$ the flow rate ($\Sigma$ is the cross section area), $\lambda=10.67$, $\alpha = 1.8520$, $\beta = -4.8704$, $\kappa$ the roughness parameter of the pipe wall and $\Phi$ its hydraulic diameter.

If the pipe has the same properties along a certain distance $\delta$, the previous expression can be immediately integrated, getting:
\begin{equation}\label{eq_mia1}
\Delta h = \frac{dh}{dx}\delta = -i \delta = -\lambda\left(\frac{q}{\kappa}\right)^\alpha\Phi^\beta \delta
\end{equation}

If we now move to the second term of the energy loss, due to the sudden pipe expansion, a typical approach is the so-called Borda-Carnot equation \cite{batchelor2000introduction}:
\begin{equation}
\label{eq::Borda-Carnot}
\Delta h = \xi\frac{1}{2g}\left(1-\frac{\Sigma_1}{\Sigma_2}\right)^2v_1^2
\end{equation}
where $\Sigma_1$ and $v_1$ are the cross-section area and flow velocity before the expansion and $\Sigma_2$ the cross-section area after expansion. $\xi$ is, again, an empirical coefficient accounting for the magnitude of the viscous eddy dissipation. 

Assuming an almost uniform flow velocity profile, which is the case for fully developed flows, it is possible to derive from the mass and momentum conservation equation that $\xi \simeq 1$ \cite{batchelor2000introduction}. 

Combining these two hydraulic head loss phenomena, we finally write:
\begin{subequations}
\label{eq::model_1}
\begin{align}
(\Delta h)_1 &= \lambda\left(\frac{q}{\kappa_1}\right)^\alpha\Phi_i^\beta \delta_1 \label{eq::model_1a} \\
(\Delta h)_e &= \xi\frac{1}{2g}\left(1-\frac{\Sigma_1}{\Sigma_2}\right)^2v_1^2 \label{eq::model_1b} \\
(\Delta h)_2 &= \lambda\left(\frac{q}{\kappa_2}\right)^\alpha\Phi_2^\beta \delta_2  \label{eq::model_1c}
\end{align}
\end{subequations}
where $\delta_1$ and $\delta_2$ are the lengths of the two tubes. In terms of the pressure drop:
\begin{subequations}
\label{eq::model_2}
\begin{align}
(\Delta p)_1 &= \lambda \gamma \left(\frac{q}{\kappa_1}\right)^\alpha\Phi_i^\beta \delta_1 \label{eq::model_2a} \\
(\Delta p)_e &= \frac{1}{2}\rho q^2\left[\left(\frac{1}{\Sigma_2^2} - \frac{1}{\Sigma_1^2}\right) + \xi\left(\frac{1}{\Sigma_1}-\frac{1}{\Sigma_2}\right)^2\right] \label{eq::model_2b} \\
(\Delta p)_2 &= \lambda \gamma\left(\frac{q}{\kappa_2}\right)^\alpha\Phi_2^\beta \delta_2  \label{eq::model_2c}
\end{align}
\end{subequations}

Eq. (\ref{eq::model_2}) describes the whole physics of the model when Hazen-Williams and Borda-Carnot loss models are assumed. Besides the empirical equation relating the energy losses with the velocity, there is another underlying universal physics inherent to the problem, although some of it has been already used even if it has been masked: the momentum conservation equation has been used in the derivation of the Bernoulli equation and the energy conservation in Eq (\ref{eq_mia1}). Another universal equation is the mass conservation (constant flow equation), that was used only in Eq. (\ref{eq::model_2b}) but will be used again later.

Let us now use the two different approaches, prediction or characterization, to solve this simple problem.

\begin{itemize}
\item \textbf{Prediction problem:} 

The aim here is to predict a pressure drop $\Delta p$ from a given flow $q$ through the pipe. This problem is illustrative in the sense that we know a conserved quantity of the problem (the mass) and we want to know the functional dependence between another physical variable (pressure drop) and this conserved quantity (equivalent to velocity). Taking into account that for circular-based cylindrical pipes $\Phi_i = \sqrt{\frac{4\Sigma_i}{\pi}}$, and renaming the parameters, we can write the equation (\ref{eq::model_2}) as:
\begin{equation}
\label{eq::model_pQ}
\Delta p = \lambda_1 q^2 + \lambda_2q^{\lambda_3}
\end{equation} 
where:
\begin{subequations}
\label{eq::lambda}
\begin{align}
\lambda_1 &= \frac{1}{2}\rho \left[\left(\frac{1}{\Sigma_2^2} - \frac{1}{\Sigma_1^2}\right) + \xi \left(\frac{1}{\Sigma_1}-\frac{1}{\Sigma_2}\right)^2\right]  \label{eq::lambda_a}\\
\lambda_2 &= \frac{2\lambda\gamma}{\sqrt{\pi}} \sum_{i=1}^2\frac{\Sigma_i^{\beta/2}}{\kappa_i^\alpha}\delta_{i} \label{eq::lambda_b}\\
\lambda_3 &= \alpha \label{eq::lambda_c}
\end{align}
\end{subequations}

This model may be seen as a physically-based mathematical relation, relating one input, $x = q$ to an output, $\bs{y} = \Delta p$ variable, by a function $\bs{y} = \bs{Y}(\mathbf{x};\boldsymbol \lambda)$, that includes three formal parameters, $\lambda_1$, $\lambda_2$ and $\lambda_3$, easily obtained from:
\begin{itemize}
\item The cross section areas of the pipe: $\Sigma_i$, $i=1,2$.
\item The roughness of the pipe wall: $\kappa_i$, $i=1,2$.
\item The lengths of different sections of the pipe: $\delta_i$, $i=1,2$.
\item Some physical parameters: density of the water $\rho$ and gravitational acceleration $g$.
\item The model parameters: the exponents $\alpha = 1.8520$ and $\beta = -4.8704$ and the coefficient $\lambda = 10.67$. The values are taken in I.S. units.
\end{itemize}

Note that this model is highly nonlinear and, despite its conceptual simplicity, is not that easy to solve using data-based approaches.

Depending on the selected approach:

\begin{enumerate}
\item We try to learn the relationship $\Delta p = \mathsf{Y}(q)$.
\item We learn $\Delta p$ from the velocities, $\Delta p = Y(v_0,v_1,v_2)$, where $v_i = \frac{q}{\Sigma_i}$. This is a simple but illustrative example of defining a new state variable from the input variable $q$.
\item We learn $\Delta p$ from the flow $q$ but by means of local pressure gradients, $\Delta p = Y(w_1,w_2) = \delta_1w_1 + (\Delta p)_e + \delta_2w_2$, where $w_i = \left.\frac{dp}{dx}\right|_i$ is the local pressure drop gradient along the segment $i$, $\delta_i$ is the length of this segment and $(\Delta p)_e$ is the pressure drop due to the expansion. This equation corresponds to momentum conservation. Besides, we have to postulate the relation $(w_1,w_2, (\Delta p)_e) = \bs H(q)$.
\item The combination of the two previous ones. We try to learn $\Delta p$ from flow $q$ but by means of local pressure gradients as before. Besides, we postulate $(w_1,w_2, (\Delta p)_e) = \bs H(v_0,v_1,v_2)$ and $(w_1,w_2, (\Delta p)_e) = \mathsf{H}(v_0,v_1,v_2)$ where we have defined a new set of internal variables, $v_i$, that must satisfy mass conservation equation, $v_i \Sigma_i = q$.
\item We try to learn $\Delta p$ from the flow $q$ as in the previous example. The only difference is that, now, we define $(w_1,w_2,(\Delta p)_e) = \bs H(v_0,v_1,v_2)$ with:

\begin{subequations}
\label{eq::tubeModel_1}
\begin{align}
H_1(v_0,v_1,v_2) &= \lambda\gamma\left(\frac{v_0\Sigma_1}{\kappa_1}\right)^\alpha \Phi_1^\beta \label{eq::tubeModel_1a} \\
H_2(v_0,v_1,v_2) &= \frac{1}{2}\rho\left[(v_2^2-v_1^2) + \xi(v_2-v_1)^2\right] \label{eq::tubeModel_1b} \\
H_3(v_0,v_1,v_2) &= \lambda\gamma \left(\frac{v_2\Sigma_2}{\kappa_2}\right)^\alpha \Phi_2^\beta \label{eq::tubeModel_1c}
\end{align}
\end{subequations}

Now, we can establish as unknown parameters $\lambda_1 = \xi \nonumber$, $\lambda_2 = \frac{\lambda\Phi_1^\beta}{\kappa_1^\alpha}$ and $\lambda_3 = \frac{\lambda\Phi_2^\beta}{\kappa_2^\alpha}$. We have therefore $\bs H(v_0,v_1,v_2) = \bs H(v_0,v_1,v_2;\bs \uplambda)$. This is quite common, as it means that we do not know the roughness of each pipe segment (equivalently, the value of $1/\kappa_i^\alpha$), but we do know the model and the geometry. The rest of the parameters $\lambda$, $\Phi_i^\beta$ act as fixed multiplicative constants.
\item The same problem as before, except for the fact that, now, we have several possible models and we want to select the best. For example, using the Darcy-Weisbach model instead of Hazen-Williams' for the distributed losses, the $\bs H$ model writes:

\begin{subequations}
\label{eq::tubeModel_2}
\begin{align}
H_1(v_0,v_1,v_2) &= \frac{1}{2}\rho f_{D_1}\frac{v_1^2}{\Phi_1} \label{eq::tubeModel_2a} \\
H_2(v_0,v_1,v_2) &= \frac{1}{2}\rho\left[(v_2^2-v_1^2) + \xi(v_2-v_1)^2\right] \label{eq::tubeModel_2b} \\
H_3(v_0,v_1,v_2) &=  \frac{1}{2}\rho f_{D_2}\frac{v_2^2}{\Phi_2} \label{eq::tubeModel_2c}
\end{align}
\end{subequations}

For that case, the unknown parameters are $\lambda_1 = \xi$, $\lambda_2 = f_{D_1}$ and $\lambda_3 = f_{D_2}$. As $f_{D_i}$ depend on the flow regime, assuming, for example, laminar regime, we have $f_{D_i} = 64 \nu/(v_i \Phi_i)$.

\end{enumerate}
\item \textbf{Characterization problem:}

Now the aim is to characterize some of the parameters of the pipe segments for a given set of values $(q, p_0, p_1, p_2)$. That is, the parameters of the constitutive equation vary from one sample to another, and the final goal is to predict those parameters defining the intrinsic behavior of the system, assumed a given state model structure. Let us suppose, for instance, that we want to characterize the roughness of the pipe in terms of the two parameters $\kappa_1$ and $\kappa_2$. For the sake of simplicity, let us fix a constant area $\Sigma_1 = \Sigma_2 = 1 \, \mathrm{m^2}$, equivalent to assume $\xi = 0$ and $\rho = 0$ at Eqs. (\ref{eq::tubeModel_1}) and (\ref{eq::tubeModel_2}). Note that this example is very illustrative in the sense that it characterizes a spatially variable property of a given material. This may be extrapolated to obtain the profile of a material parameter $\kappa = \kappa(\bs{x})$ for heterogeneous materials, when monitoring its behavior under certain actions.

As we are in the heterogeneous case, for discovering the roughness parameters, it is necessary to measure the pressure drop at the two segments, otherwise, the problem would be undetermined. For the present problem, the relation to be learned is $(q,p_0,p_1,p_2) \rightarrow (\kappa_1,\kappa_2)$. For the Hazen-Williams model, the parameters $\kappa_1$ and $\kappa_2$ are related to the above variables by:

\begin{equation} \label{eq::true_kappa_i}
\kappa_i = \left(\gamma \lambda \Phi_i^\beta \delta_i\right)^{1/\alpha}q \left(p_{i-1}-p_{i}\right)^{-1/\alpha}
\end{equation}
that is $\kappa_i = \lambda_1 q \left(p_{i-1}-p_{i}\right)^{\lambda_2}$ where $\lambda_1 = \left(\gamma \lambda \Phi_i^\beta \delta_i\right)^{1/\alpha}$ and $\lambda_2 = -\frac{1}{\alpha}$. Note that the parameter dependence is one to one, that is, $\kappa_1 = Y(p_0,p_1)$ and $\kappa_2 = Y(p_2,p_1)$. This is not the general case, but could be exploited in the design of the state equation model $\boldsymbol \upkappa = \mathsf{H}(p_0,p_1,p_2)$, as pointed out in Section \ref{ssecc::types}. However, this discussion is important and will be elaborated in upcoming papers. Here, conventional multilayer perceptrons $\mathsf{H}$ will be used to model the state equation $\bs H$.

Depending on the selected approach, we would like:

\begin{enumerate}
\item To learn the relationship $(\kappa_1,\kappa_2) = \mathsf{Y}(q,p_0,p_1,p_2)$.
\item To learn the variables $\kappa_1$ and $\kappa_2$ from the velocities and pressures, $(\kappa_1,\kappa_2) = \mathsf{Y}(v_1,v_2,p_0,p_1,p_2)$, with velocities satisfying the conservation equation, $q = \Sigma_i v_i$.
\item To learn $\kappa_1$ and $\kappa_2$ from the pressures and flow velocities, $(\kappa_1, \kappa_2) = \mathsf{Y}(v_1,v_2,w_1,w_2)$ where $(w_1,w_2) = \mathsf{H}(p_0,p_1,p_2)$, with $w_1, w_2$ the head pressure drops (related to viscous forces) and $R(w_1,w_2,p_0,p_1,p_2) =(\delta_1 w_1 -(p_1-p_0),\delta_2 w_2 -(p_2 - p_1)$ comes from the momentum conservation.
\end{enumerate}
\end{itemize}

\subsection{Results of the prediction approach}

\subsubsection{Direct input-ouput prediction. Comparison between the classical unconstrained and the constrained neural networks }

\begin{figure}
\begin{subfigure}{1.0\textwidth}
\centering
\begin{tikzpicture}

\foreach \m/\l [count=\y] in {1}
  \node [every neuron/.try, neuron \m/.try] (input-\m) at (0,0) {};
\foreach \m [count=\y] in {1,2,3}
  \node [every neuron/.try, neuron \m/.try] (velocity-\m) at (2,3-\y*1.5) {};
\foreach \m [count=\y] in {1,missing,2}
  \node [every neuron/.try, neuron \m/.try] (hiddena-\m) at (4,4-\y*2) {};
\foreach \m [count=\y] in {1,missing,2}
  \node [every neuron/.try, neuron \m/.try] (hiddenb-\m) at (6,4-\y*2) {}; 
\foreach \m [count=\y] in {1}
  \node [every neuron/.try, neuron \m/.try] (output-\m) at (8,0) {};

\foreach \l [count=\i] in {1}
  \draw [<-] (input-\i) -- ++(-1,0)
    node [above, midway] {$q$};

\foreach \l [count=\i] in {1,2,3}
  \node [above] at (velocity-\i.north) {$y_{1,\l}$};
\foreach \l [count=\i] in {1,15}
  \node [above] at (hiddena-\i.north) {$y_{2,\l}$};
\foreach \l [count=\i] in {1,15}
  \node [above] at (hiddenb-\i.north) {$y_{3,\l}$};

\foreach \l [count=\i] in {1}
  \draw [->] (output-\i) -- ++(1,0)
    node [above, midway] {$\Delta p$};

\foreach \i in {1}
  \foreach \j in {1,2,3}
    \draw [->] (input-\i) -- (velocity-\j);   
\foreach \i in {1,2,3}
  \foreach \j in {1,2}
    \draw [->] (velocity-\i) -- (hiddena-\j);
\foreach \i in {1,2}
  \foreach \j in {1,2}
    \draw [->] (hiddena-\i) -- (hiddenb-\j);
\foreach \i in {1,2}
  \foreach \j in {1}
    \draw [->] (hiddenb-\i) -- (output-\j);

\foreach \l [count=\x from 0] in {Input, Hidden, Hidden, Hidden, Output}
  \node [align=center, above] at (\x*2,3) {\l \\ layer};
\end{tikzpicture}
\caption{Unconstrained neural network.}
\label{fig_unconstrainedNN}
\end{subfigure} \\%

\begin{subfigure}{1.0\textwidth}
\centering
\begin{tikzpicture}
\foreach \m [count=\y] in {1,2,3}
  \node [every neuron/.try, neuron \m/.try,red] (outputP-\m) at (8,1.75-\y*1.25) {};
\foreach \i in {1,2,3}
    \draw [->,red] (velocity-\i) -- (outputP-\i);
\foreach \i in {1,2,3}
    \draw [->,red] (input-1) -- (outputP-\i);

\foreach \m/\l [count=\y] in {1}
  \node [every neuron/.try, neuron \m/.try] (input-\m) at (0,0) {};
\foreach \m [count=\y] in {1,2,3}
  \node [every neuron/.try, neuron \m/.try] (velocity-\m) at (2,3-\y*1.5) {};
\foreach \m [count=\y] in {1,missing,2}
  \node [every neuron/.try, neuron \m/.try] (hiddena-\m) at (4,4-\y*2) {};
\foreach \m [count=\y] in {1,missing,2}
  \node [every neuron/.try, neuron \m/.try] (hiddenb-\m) at (6,4-\y*2) {};
\foreach \m [count=\y] in {1}
  \node [every neuron/.try, neuron \m/.try] (output-\m) at (8,2) {};

\foreach \m [count=\y] in {2}
  \node [every neuron/.try, neuron \m/.try,fill=white] (hiddenb-\m) at (6,-2) {};
\foreach \m [count=\y] in {2}
  \node [every neuron/.try, neuron \m/.try,fill=white] (velocity-\m) at (2,0) {};

\foreach \l [count=\i] in {1}
  \draw [<-] (input-\i) -- ++(-1,0)
    node [above, midway] {$q$};

\foreach \l [count=\i] in {1,2,3}
  \node [above] at (velocity-\i.north) {$v_\l$};
\foreach \l [count=\i] in {1,15}
  \node [above] at (hiddena-\i.north) {$y_{2,\l}$};  
\foreach \l [count=\i] in {1,15}
  \node [above] at (hiddenb-\i.north) {$y_{3,\l}$};

\foreach \l [count=\i] in {1}
  \draw [->] (output-\i) -- ++(1,0)
    node [above, midway] {$\Delta p$};  
\foreach \l [count=\i] in {1,2,3}
  \draw [->] (outputP-\i) -- ++(1,0)
    node [above, midway] {$0$};

\foreach \i in {1}
  \foreach \j in {1,2,3}
    \draw [->] (input-\i) -- (velocity-\j);
\foreach \i in {1,2,3}
  \foreach \j in {1,2}
    \draw [->] (velocity-\i) -- (hiddena-\j);
\foreach \i in {1,2}
  \foreach \j in {1,2}
    \draw [->] (hiddena-\i) -- (hiddenb-\j);  
\foreach \i in {1,2}
  \foreach \j in {1}
    \draw [->] (hiddenb-\i) -- (output-\j);

\foreach \l [count=\x from 0] in {Input, Velocity, Hidden, Hidden, Output}
  \node [align=center, above] at (\x*2,3) {\l \\ layer};
\end{tikzpicture}
\caption{Physically-Guided neural network.}
\label{fig_constrainedNN}
\end{subfigure}
\caption{\textbf{Comparison of unconstrained and constrained neural network.} The constrained neural network (Physically-Guided Neural Network) is illustrated by its physically augmented network, where constraints have been replaced by extra outputs.}
\end{figure}
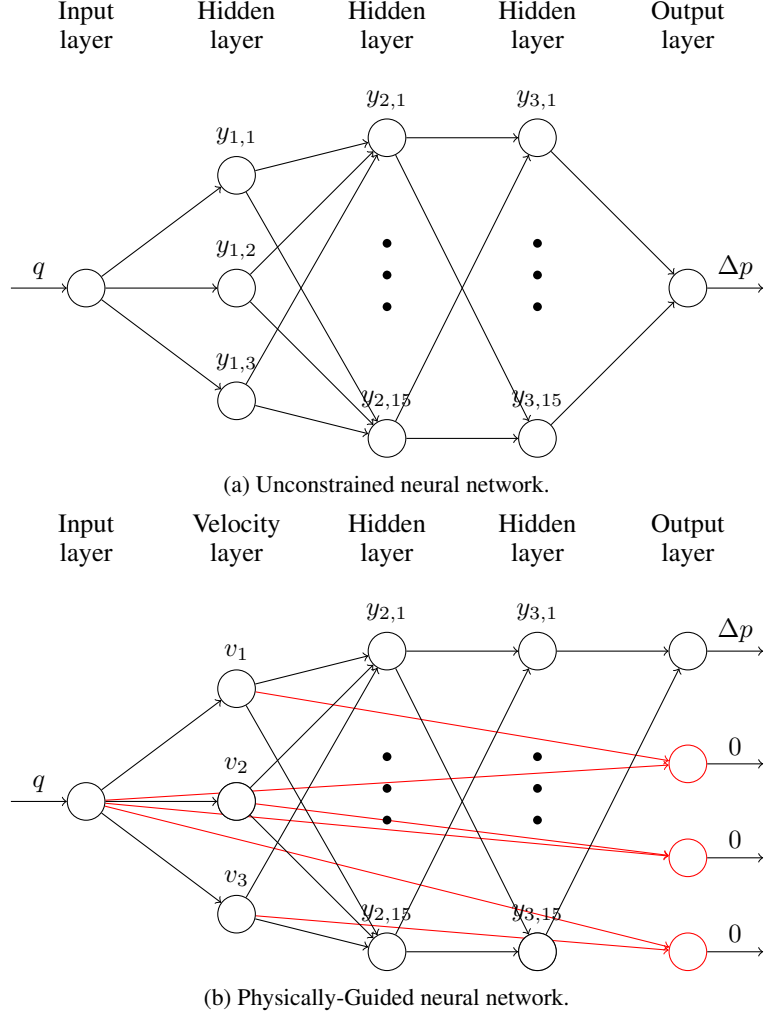

Let us first consider the problem of predicting directly the nonlinear relationship $q \rightarrow \Delta p$ without any additional constraint, that is a standard NN approach. A neural network is established to solve the \emph{single input - single output} $q \rightarrow \Delta p$ problem proposed. We choose a neural network with only three hidden layers of $n_1=3$, $n_2 = 15$ and $n_3 = 15$ neurons, respectively. The network is illustrated in Figure \ref{fig_unconstrainedNN}

The neural layers are defined mathematically as follows. Given $\bs{x}=\bs{y}_0=q$, $\bs{y}= \bs{y}_4 = \Delta p$ and

\begin{align}
\bs{y}_1 &= \bs{x}\bs{W}_1+\bs{b}_1, \qquad \bs{y}_2 = \bs{y}_1\bs{W}_2+\bs{b}_2 \nonumber \\
\bs{y}_3 &= \mathrm{ReLU}(\bs{y}_2\bs{W}_3+\bs{b}_3) \nonumber \\
\bs{y}_4 &= \bs{y}_3\bs{W}_4+\bs{b}_4
\end{align}
with $\mathrm{ReLU}$ a Rectified Linear Unit activation function. 

Now we define an analogous neural network in which we impose mass conservation, that is, constant flow in the three reference points of the pipe:

\begin{equation}
\label{eq::flow_cons}
v_i\Sigma_i = q, \quad i=0,1,2
\end{equation}

The imposition of Eqs. (\ref{eq::flow_cons}) gives the hidden variables $\bs{y}_1$ a clear physical interpretation: the flow velocities $v_i$. Eq. (\ref{eq::flow_cons}) is imposed in the neural network system via constraints between the values of the corresponding neurons by a penalty approach. Fig. \ref{fig_constrainedNN} illustrates the interpretation of this physically guided network. The network topology is identical to the one of the unconstrained network, but the objective function includes now additional terms, accounting for the constraints associated with the physics. 

For the training process, the data input was randomly generated with a uniform distribution, using the state model presented in equation (\ref{eq::model_pQ}) for $q \in [1.0;5.0] \,  (\mathrm{m^3/s})$. The physical parameters used for the data generation are shown in Table \ref{table::param_Ph}.

\begin{table}[!htbp]
\caption{Physical parameters for the problem with fixed geometry.}
\centering
\begin{tabular}{cccccccccc}
\toprule
Parameter & $\Sigma_1$ & $\Sigma_1$ & $\rho$ & $\xi$ & $g$ & $\kappa_1$ & $\kappa_2$ & $\delta_1$ & $\delta_2$\\
\midrule
Value & $1.0 \, \mathrm{m^2}$ & $2.0 \, \mathrm{m^2}$ & $1.0 \, \mathrm{m^3/kg}$ &$1.0$ & $9.81 \, \mathrm{m/s^2}$ & $140$ & $140$ &  $10 \, \mathrm{m}$ & $10 \, \mathrm{m}$ \\
\bottomrule
\end{tabular}
\label{table::param_Ph}
\end{table}

As a learning algorithm, a gradient descent optimizer was selected with learning rate $\beta=0.001$. At each training step, $n=4$ data points were selected, enough for our purpose. For the constrained network, we chose a penalty parameter $p=0.01 \, \mathrm{Pa^2 m^6 / s^2}$. $N_\mathrm{test} =  1000$ samples were randomly generated for the testing procedure. 

\begin{figure}
\centering
\begin{subfigure}{.6\textwidth}
  \includegraphics[width=\linewidth]{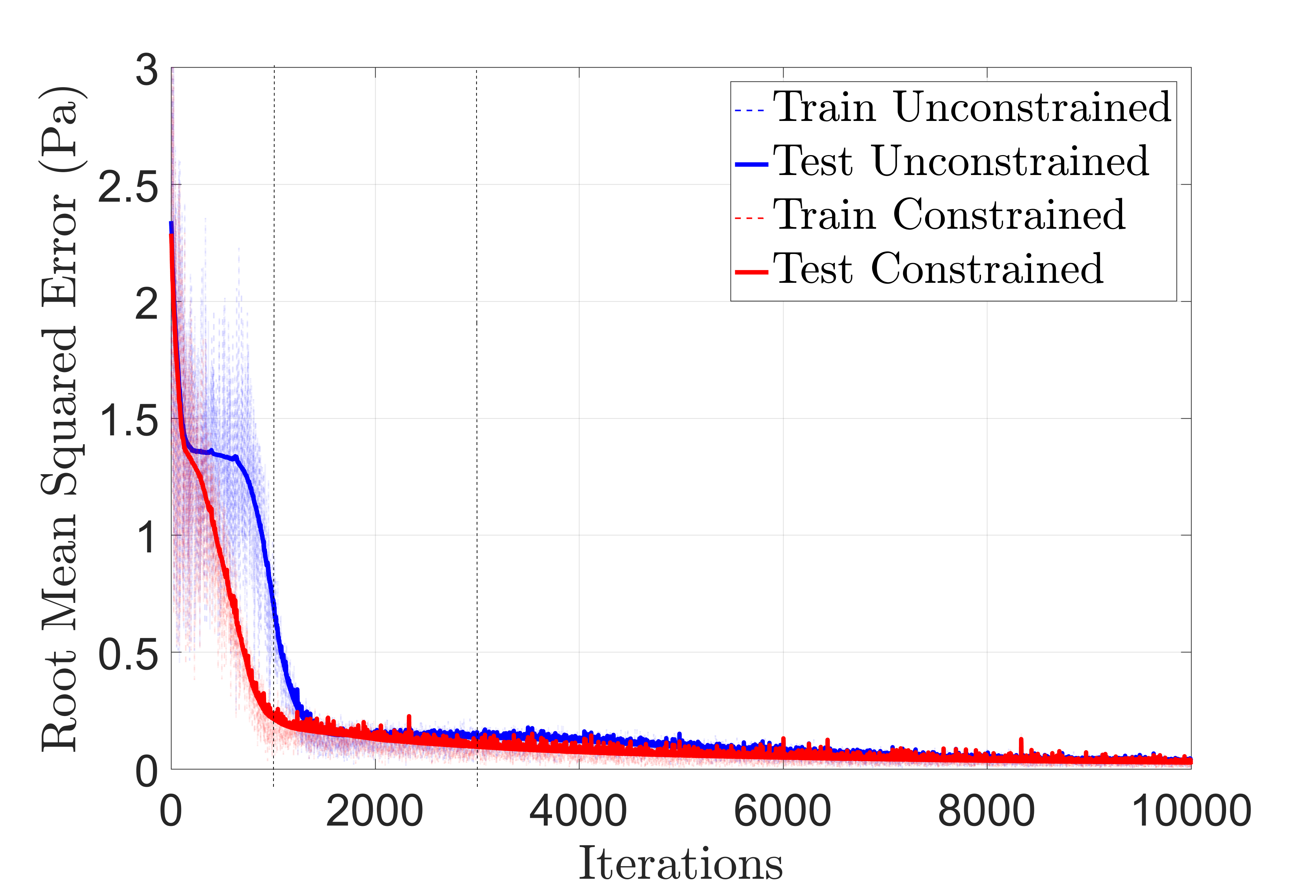}
  \caption{Root Mean Squared Error ($\mathrm{RMSE} = \sqrt{\mathrm{MSE}}$).}
  \label{fig::convergence_1b}
\end{subfigure} \\
\begin{subfigure}{.6\textwidth}
  \includegraphics[width=\linewidth]{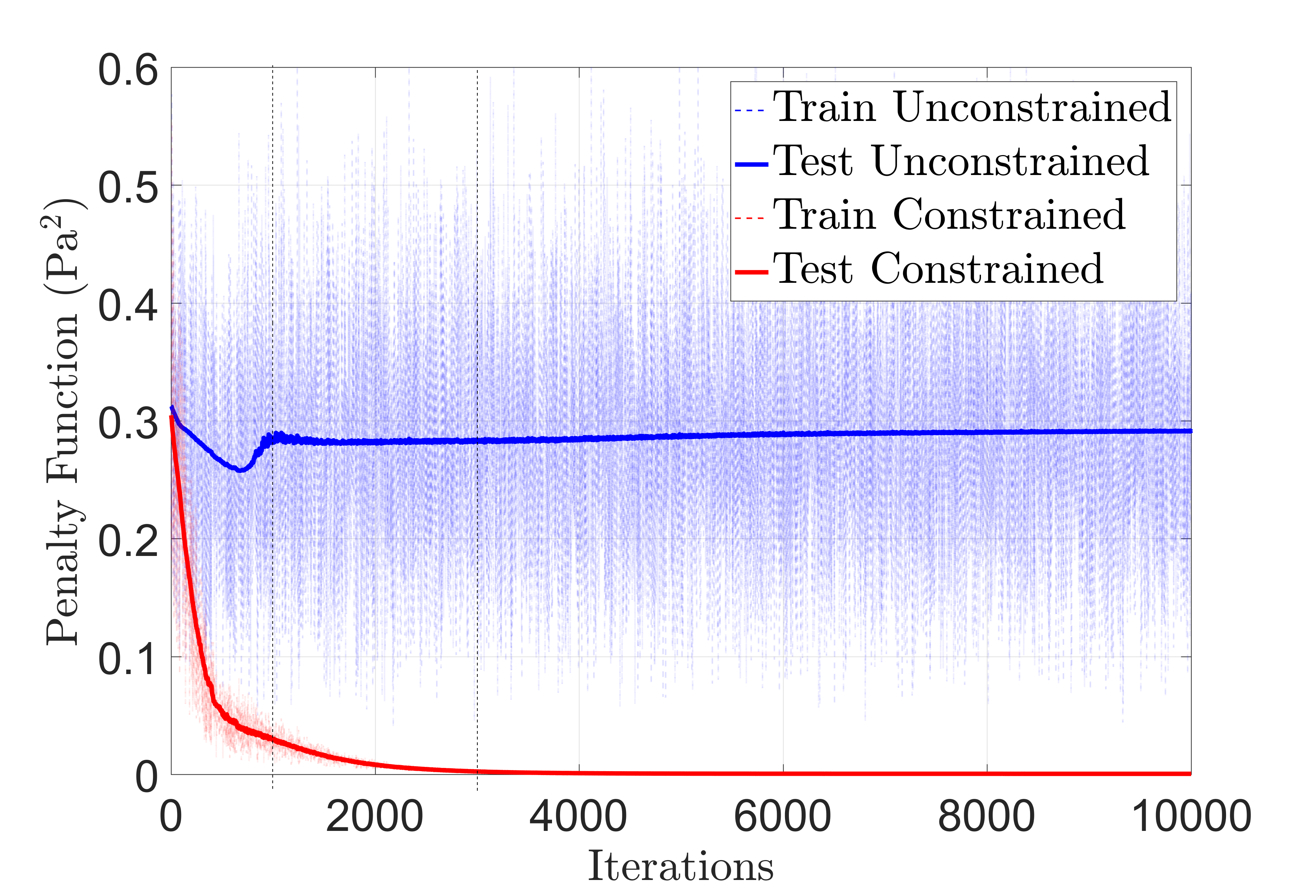}
  \caption{Penalty function (PEN).}
  \label{fig::convergence_1c}
\end{subfigure}
\caption{\textbf{Root Mean Squared Function and penalty function.} Note that, as the topology and parameters of the network between the $v$ and $\Delta p$ layers are the same, the asymptotic trend of both RMSE functions is the same. Only a modification in the network topology associated with the model, that is, in $\bs H$, would improve the accuracy of the method.}
\label{fig::convergence_1}
\end{figure}

Fig. \ref{fig::convergence_1} shows the value of the Root Mean Square Error (RMSE) and the Penalty (PEN) functions along the training iterations. The effect of including the penalty term (related to the physics of the problem) is clearly illustrated in Fig \ref{fig::convergence_1} that shows that the RMSE has a faster decay in the early learning steps. Indeed, the error converges to the same value when the number of iterations increases. Although there is not a general recipe for the model improvement and each benchmark problem requires its own strategy, as discussed in the broad bibliography on neural networks \cite{nielsen2015neural}, the fundamental conclusion here is that the constrained networks does not necessarily improve the accuracy of the model if they both have the same network topology and metaparameters, but the introduction of physical constraints does speed-up the network convergence. This speed-up is also explained by the evolution of the penalty term value PEN (Fig. \ref{fig::convergence_1c}) for both neural networks: for the unconstrained one, this term is not included in the penalty function and therefore it is non necessarily decreasing. Of course, since the unconstrained network does not force its fulfillment and the error MSE goes also to zero, it is clear that with this network topology, there is not a global minimum solution but many local minima. However, the convergence to one of these minimal solutions is accelerated in the constrained case and for, let's say, $N=600$ iterations, the behavior of the unconstrained network is suboptimal. Unless the whole underlying parametric model is assumed as known, $H = H(\cdot,\boldsymbol \lambda)$, what would place us in a case analog to classical parametric fitting via optimization procedures, there is no simple way of defining a reduced enough network to guarantee both the required abstraction capability (generalization) and global minimal requirements (specificity), reducing the computational cost.

\begin{figure}
\centering
\begin{subfigure}{.48\textwidth}
  \centering
  \includegraphics[width=\linewidth]{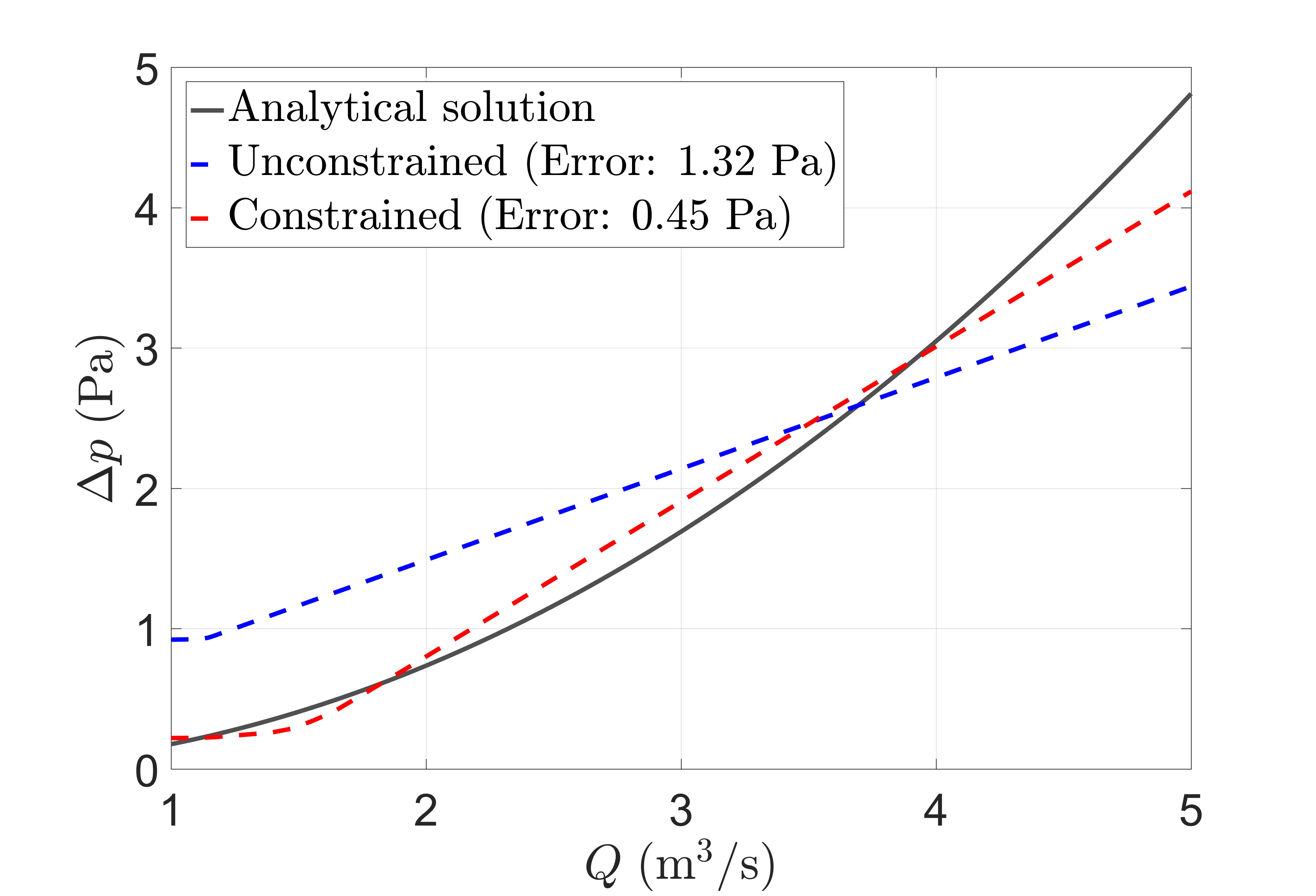}
  \caption{$N=1000$.}
  \label{fig::error_1a}
\end{subfigure}
\begin{subfigure}{.48\textwidth}
  \centering
  \includegraphics[width=\linewidth]{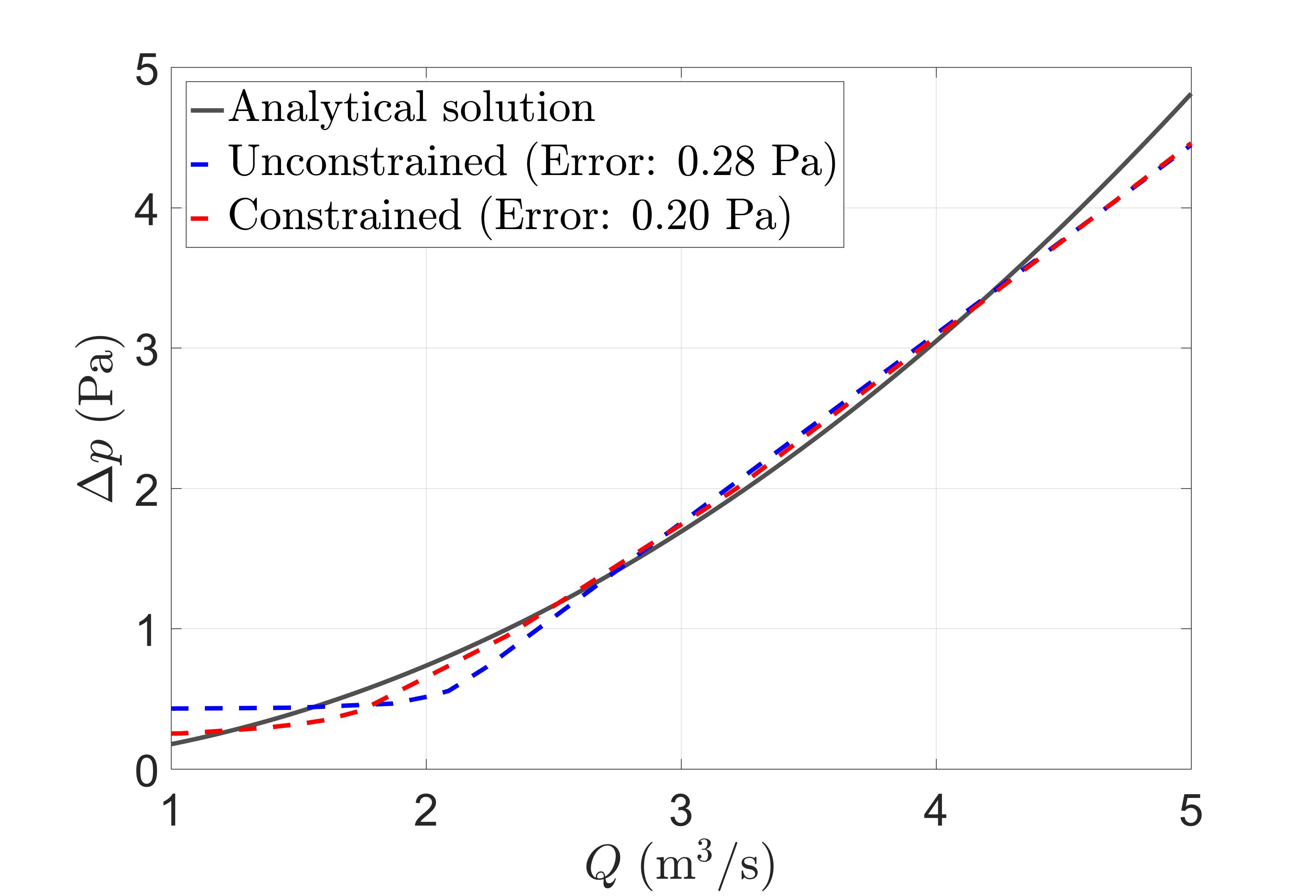}
  \caption{$N=3000$.}
  \label{fig::error_1b}
\end{subfigure} \\
\begin{center}
\begin{subfigure}{.48\textwidth}
  \centering
  \includegraphics[width=\linewidth]{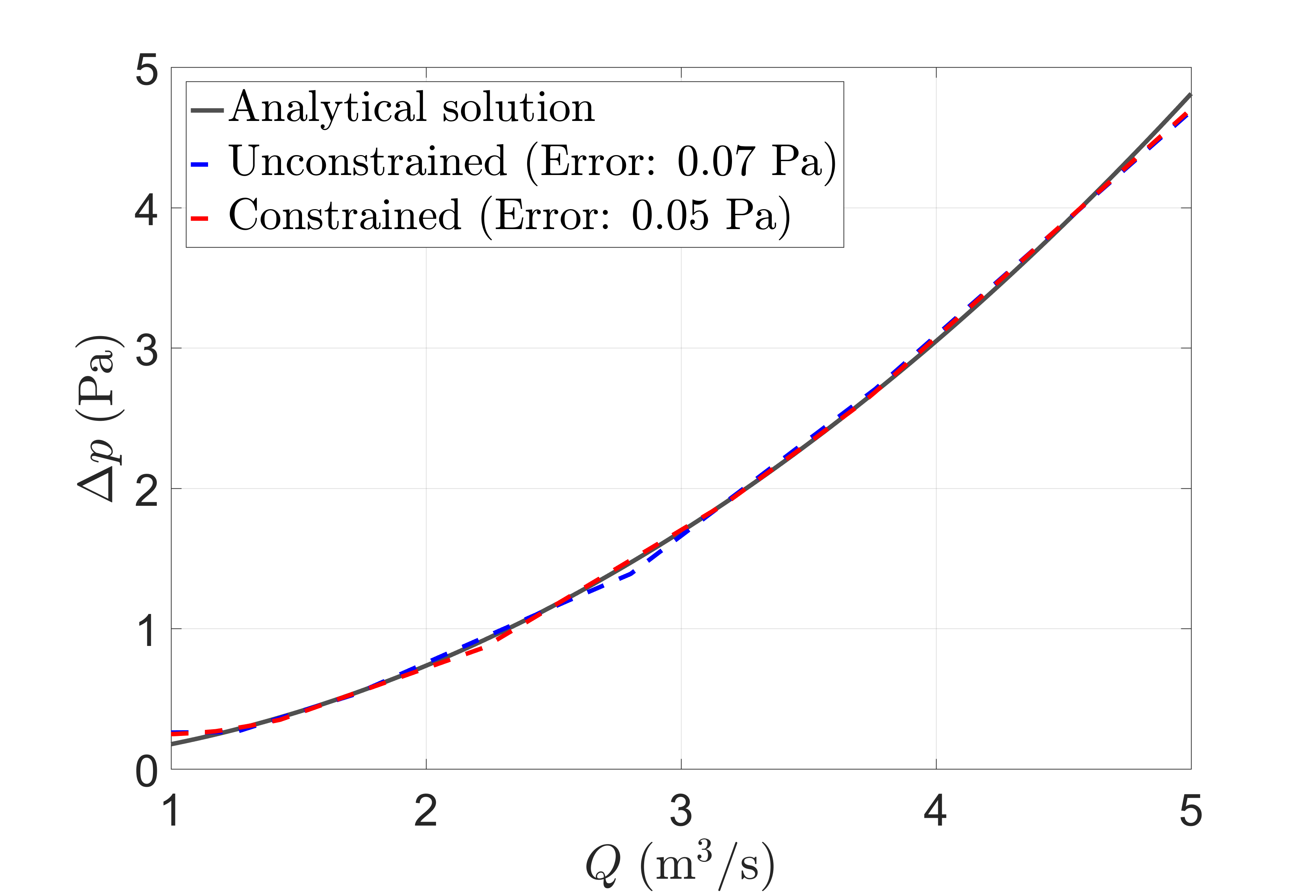}
  \caption{$N=10000$.}
  \label{fig::error_1c}
\end{subfigure}%
\end{center}
\caption{\textbf{Comparison of the network output with the analytical model.} Both models present a similar error. The effect of the constraints is not the improvement of the accuracy, but to speed-up the convergence, besides the physical interpretation of some of the internal layers.}
\label{fig::error_1}
\end{figure}

Fig. \ref{fig::error_1} shows the accuracy of constrained and unconstrained neural networks after $N=1000,3000,10000$ iterations (marked with a dashed bar in Fig. \ref{fig::convergence_1})) when compared to the analytical solution. As explained before, the performance of both networks, if we assume convergence, is similar and only the learning rate, not the accuracy, is improved by the constrained network. The error included in Fig. \ref{fig::error_1} was computed as:
\begin{equation}
E_{L_2}  =\left(\int_{0}^{10} (Y(q) - Y_m(q))^2 \, dq\right)^{1/2}
\end{equation}
where $Y(q)$ is the network predicted pressure drop and $Y_m(q)= \lambda_1 q^2 + \lambda_2q^{\lambda_3}$ the analytical solution.

The problem may be enriched by taking into account some geometrical aspects. For example, we can consider the two pipe lengths, $l_1$ and $l_2$  (until now, they were denoted as $\delta_1$ and $\delta_2$ because they were constant parameters) as extra input variables. This adds a double benefit: (i) it allows us to consider variable geometries and (ii), if some problem parameters are known, the neural network may be simply adapted and simplified to include more physical knowledge. Indeed, for $\xi = 0$, we know that all pressure drop is associated with the distributed head loss along the two stretches of lengths $l_1$ and $l_2$ so that the hidden layer may be replaced by a layer with two neurons, whose relationship with the output neuron will be $\Delta p = l_1y_{3,1} +l_2y_{3,2}$. With these considerations, the third hidden layer acquires also a physical meaning (the local pressure drop per unit length at stretches 1 and 2). It is important to note that the constraint in the first hidden layer is now crucial, because the input variables have different dimension and the normalization and the ReLU activation function acting between the first and second layer are complemented by the constraint indicating that the lengths $l_1$ and $l_2$ do not influence the flow velocity.

Neural layers are defined mathematically as follows. Given $\bs{x}=(q,l_1,l_2)$:
\begin{align}
\bs{y}_1 & = \mathrm{ReLU}(\bs{x}\bs{W}_1+\bs{b}_1), \qquad \bs{y}_2 = \mathrm{ReLU}(\bs{y}_1\bs{W}_2+\bs{b}_2) \nonumber \\
\bs{y}_3 & = \mathrm{ReLU}(\bs{y}_2\bs{W}_3+\bs{b}_3), \qquad \bs y_4 =  y_{3,1}l_1 + y_{3,2}l_2 = \Delta p
\end{align}

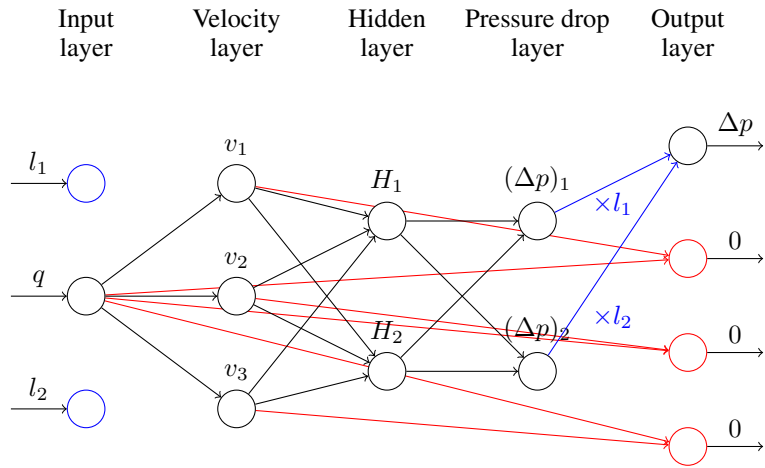
\begin{figure}
\centering
\begin{tikzpicture}

\foreach \m [count=\y] in {1,2,3}
  \node [every neuron/.try, neuron \m/.try,red ] (outputP-\m) at (8,1.75-\y*1.25) {};
\foreach \i in {1,2,3}
    \draw [->,red] (velocity-\i) -- (outputP-\i);
\foreach \i in {1,2,3}
    \draw [->,red] (input-1) -- (outputP-\i);
  
\foreach \m/\l [count=\y] in {1}
  \node [every neuron/.try, neuron \m/.try] (input-\m) at (0,0) {};
\foreach \m [count=\y] in {1,2,3}
  \node [every neuron/.try, neuron \m/.try,fill=white] (velocity-\m) at (2,3-\y*1.5) {};
\foreach \m [count=\y] in {1,2}
  \node [every neuron/.try, neuron \m/.try,fill=white] (hidden-\m) at (4,3-\y*2) {};
\foreach \m [count=\y] in {1,2}
  \node [every neuron/.try, neuron \m/.try,fill=white] (dp-\m) at (6,3-\y*2) {};
\foreach \m [count=\y] in {1}
  \node [every neuron/.try, neuron \m/.try,fill=white] (output-\m) at (8,2) {};
\foreach \m [count=\y] in {1}
  \node [every neuron/.try, neuron \m/.try,blue] (inputL-\m) at (0,3-\y*1.5) {};
\foreach \m [count=\y] in {2}
  \node [every neuron/.try, neuron \m/.try,blue] (inputL-\m) at (0,3-\y*4.5) {};

\foreach \l [count=\i] in {1}
  \draw [<-] (input-\i) -- ++(-1,0)
    node [above, midway] {$q$};
\foreach \l [count=\i] in {1,2}
  \draw [<-] (inputL-\i) -- ++(-1,0)
    node [above, midway] {$l_\i$};

\foreach \l [count=\i] in {1,2,3}
  \node [above] at (velocity-\i.north) {$v_\l$};
\foreach \l [count=\i] in {1,2}
  \node [above] at (hidden-\i.north) {$H_{\l}$};  
\foreach \l [count=\i] in {1,2}
  \node [above] at (dp-\i.north) {$(\Delta p)_{\l}$};

\foreach \l [count=\i] in {1}
  \draw [->] (output-\i) -- ++(1,0)
    node [above, midway] {$\Delta p$};  
\foreach \l [count=\i] in {1,2,3}
  \draw [->] (outputP-\i) -- ++(1,0)
    node [above, midway] {$0$};

\foreach \i in {1}
  \foreach \j in {1,2,3}
    \draw [->] (input-\i) -- (velocity-\j);
\foreach \i in {1,2,3}
  \foreach \j in {1,2}
    \draw [->] (velocity-\i) -- (hidden-\j);
\foreach \i in {1,2}
  \foreach \j in {1,2}
    \draw [->] (hidden-\i) -- (dp-\j);
\foreach \i in {1}
  \foreach \j in {1}
    \draw [->,blue] (dp-\i) -- node[below] {$\times l_\i$}(output-\j);
\foreach \i in {2}
  \foreach \j in {1}
    \draw [->,blue] (dp-\i) -- node[below=0.5cm] {$\times l_\i$}(output-\j);

\foreach \l [count=\x from 0] in {Input, Velocity, Hidden, Pressure drop, Output}
  \node [align=center, above] at (\x*2,3) {\l \\ layer};
\end{tikzpicture}
\caption{\textbf{Physically augmented neural network for the geometry-dependent problem.} The red lines illustrate the velocity definition in terms of flow, while the blue lines represent the geometry inclusion by means of the momentum conservation equation.}
\label{fig_geometrydNN}
\end{figure}

The constrained network includes the same constraint as before, relating flow and velocities. The physically guided representation of this new constrained neural network is illustrated in Fig. \ref{fig_geometrydNN}.

All physical and geometrical parameters are the same than in the preceding example, except that $\xi = 0$, $\Sigma_1 = \Sigma_2 = 1 \; \mathrm{m^2}$, since the effect of the pipe expansion is not the relevant phenomenon here, $\kappa_1 = 140$ and $\kappa_2 = 100$. $l_1$ and $l_2$ were uniformly generated between $0$ and $10$. As before, a gradient descent optimizer was selected with learning rate $\beta=0.003$. At each training step, $n=100$ data points were selected. The same value for the penalty parameter ($p=0.01 \, \mathrm{Pa^2 m^6 / s^2}$) was selected and $N_\mathrm{test} =  10000$ samples were randomly generated for the testing procedure. 

\begin{figure}
\centering
\begin{subfigure}{.6\textwidth}
  \centering
  \includegraphics[width=\linewidth]{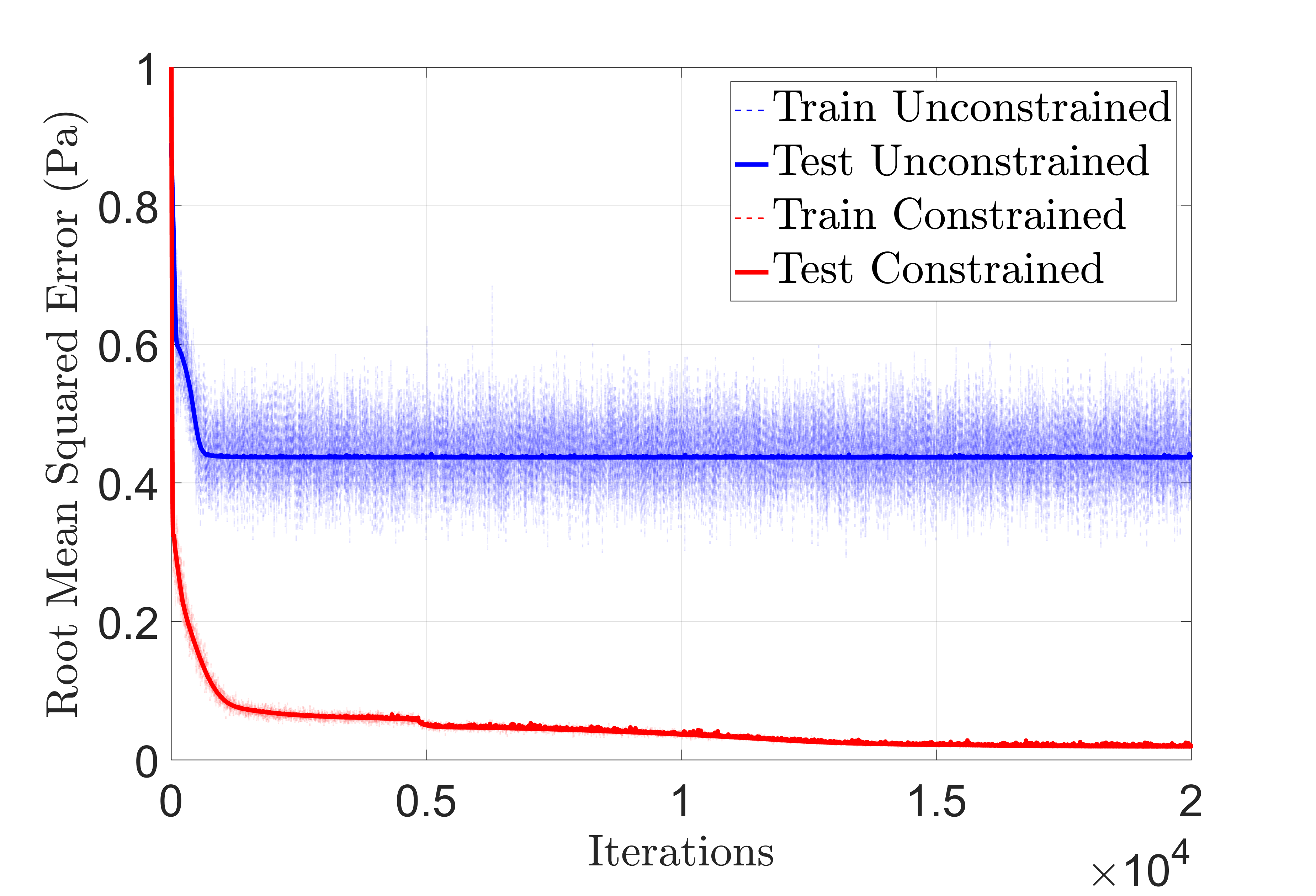}
  \caption{Root Mean Squared Error ($\mathrm{RMSE} = \sqrt{\mathrm{MSE}}$).}
  \label{fig::convergence_2b}
\end{subfigure} \\
\begin{subfigure}{.6\textwidth}
  \centering
  \includegraphics[width=\linewidth]{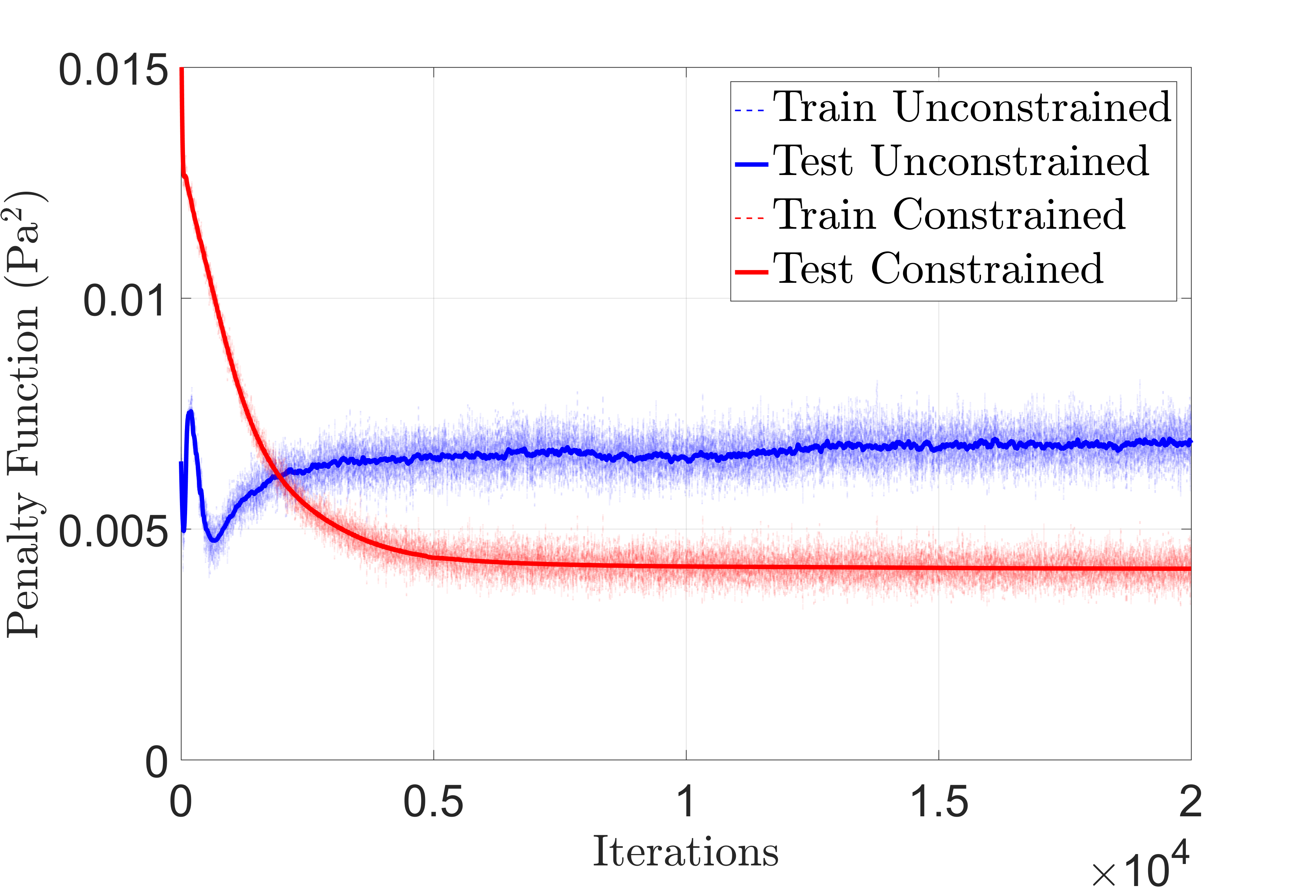}
  \caption{Penalty function (PEN).}
  \label{fig::convergence_2c}
\end{subfigure}%
\caption{\textbf{Root Mean Squared Function and penalty function for the network including geometry.} The physical constraints give the hidden layers the correct physical interpretation, such as the integration constraint, $\Delta p = w_1 l_1 + w_2 l_2$ is correctly formulated. This is achieved thanks to the effect of the penalty term, which gives the PILs layers their correct interpretation.}
\label{fig::convergence_2}
\end{figure}

\begin{figure}
\centering
\includegraphics[clip=true,trim=0pt 0pt 0pt 0pt,width=0.8\textwidth]{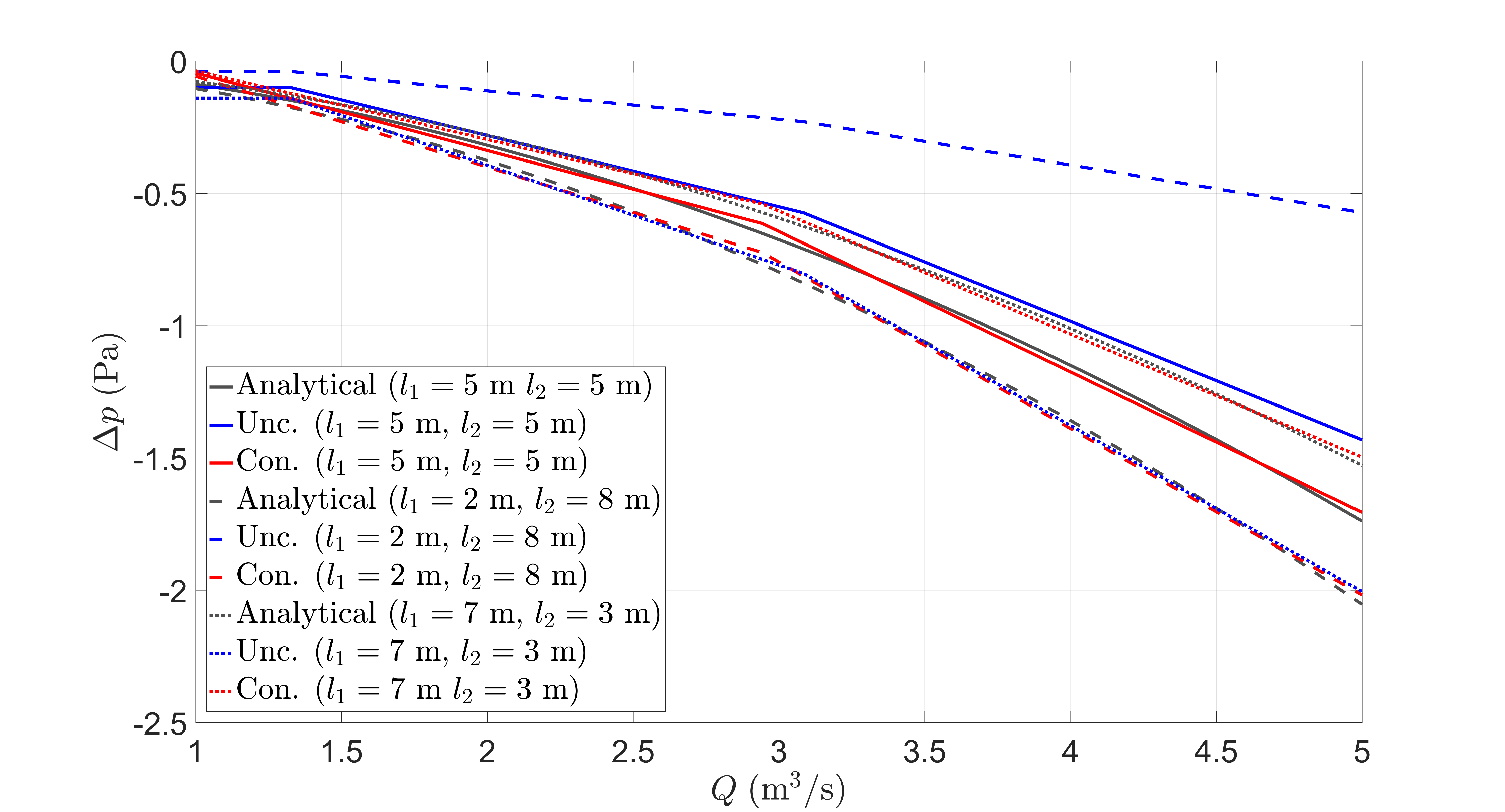}
\caption{\textbf{Exact and predicted solution for different lengths of the segments.} The specified network topology and the introduction of the physical constraints are responsible for the network convergence. Results are given for different segment lengths.}
\label{fig::error_2}
\end{figure}

As in the previous case, Fig. \ref{fig::convergence_2} shows the convergence curves for the RMSE and PEN functions demonstrating good convergence and no overfitting may be observed. Fig. \ref{fig::error_2} shows the accuracy of the constrained and unconstrained neural networks after $N=20000$ iterations. In that case, in addition to the acceleration of the convergence, the PGNNIV shows a better accuracy, as expected, because the topology of the network was thought in a physical sense, with the last hidden layer having a physical interpretation (the internal variable $w_i$, $i=1,2$).

For comparison purposes, Table \ref{table::geometry_comparison} shows the statistics of the relative error $\varepsilon_r = \frac{(\Delta p)_\mathrm{predicted} - (\Delta p)_\mathrm{true}}{(\Delta p)_\mathrm{true}}$ obtained for both the unconstrained and constrained networks, when $100 \times 100 \times 100$ values of $q$, $l_1$ and $l_2$ were sampled in $[1;5] \times [0;10] \times [0;10]$. It is clear that the effect of the constraints is to reduce the relative error together with its variability.

\begin{table}[!htbp]
\caption{Statistics of the relative error $\varepsilon_r$ for the two networks.}
\centering
\begin{tabular}{lcccccc}
\toprule
Network & Minimum & $Q_1$ & $Q_2$ & $Q_3$ & Maximum & Mean ( $\pm$ Std. error) \\
\midrule
Unconstrained & $-0.97$ & $-0.55$ & $-0.17$ & $0.38$ & $2.8 \times 10^2$ & $25.0 \times 10^{-2} \quad (\pm 0.3 \times 10^{-2})$ \\
Constrained   & $-0.58$ & $-0.02$ & $0.01$ & $0.02$ & $0.08$ & $-1.123 \times 10^{-2} \quad (\pm 0.009 \times 10^{-2})$ \\
\bottomrule
\end{tabular}
\label{table::geometry_comparison}
\end{table}

\clearpage

\subsubsection{Prediction of the internal variables and identification of the state model}

Now, and besides the constraints associated with physical principles, we evaluate the effect of adding model constraints to the network. We shall also discuss the explanatory capacity of the presented method in learning the internal physical variables and, if it is the case, the model parameters. As previously explained, there are two kinds of equations used in the formulation of the hydraulic head loss in a pipe:

\begin{itemize}
\item \textbf{Fundamental principles}: Mass ($\Sigma_1 v_1 = \Sigma_2 v_2$), linear momentum ($\Delta p = w_1 \delta_1 + (\Delta p)_e +  w_2 \delta_2$) conservation and energy (hydraulic head) balance ($\Delta h = (\Delta h)_1 + (\Delta h)_e + (\Delta h)_2$) (the subscripts indicate the corresponding segment of the pipe).
\item \textbf{Constitutive equation}: These equations relate the hydraulic head loss (which is directly related to pressure drop by means of Bernoulli equation) to the fluid velocity along the streamline. E.g. the Hazen-Williams or Darcy-Weisbach's for the losses associated with the pipe roughness and Borda-Carnot's for the eddy energy dissipation due to the pipe expansion.
\end{itemize}

It is clear that there is no general need of including constraints related to the constitutive equation, following the approach of the two previous examples. This is, indeed, contraindicated if there is no knowledge about the underlying behavior of the fluid (physical nature, regime...). However, it might be interesting in at least two circumstances:

\begin{enumerate}
\item Model selection: We want to select among many candidate models able to capture, from a macroscopic point of view, the fluid behavior. For instance, in the present example, we shall choose between the Darcy-Weisbach and the Hazen-Williams models for the hydraulic losses. Indirectly, this may give us information about the fluid regime, since the relationship between the Darcy factor $f_D$ and the fluid velocity is fixed ($f_D = \frac{64 \nu}{v \Phi}$ for instance for laminar regime). 
\item Structure physical discovering: Usually, the model parameters are related to some physical properties that give us insight on the nature, structure, or geometry of the problem. In the present example, for the Hazen- Williams model, $\kappa_1$ and $\kappa_2$ are related to the roughness of the pipe segments and in the Borda-Carnot model, $\xi$ is related to the gradualness of the expansion.
\end{enumerate}

As the aim is to predict the state model, it is clear that now the output for each data point must be a triplet of values $((\Delta p)_1,(\Delta p)_e, (\Delta p)_2))$ corresponding to the pressure drop at segment 1, expansion and segment 2, respectively. Without this multiple output consideration, it should be impossible to distinguish between effects in the whole pressure drop. In what follows, three neural networks (with and without constraints) are compared:

\begin{enumerate}
    \item Model-free approach: Physically-Guided Neural Network where the physics (fundamental laws) are imposed via appropriate constraints in certain layers. This occurs when we add the constraints by means of functions $v_i = E_i(q) = \frac{q}{\Sigma_i}, \,  i=1,2$.
    \item Model-based approach: Physically-Guided Neural Network where both, physical and empirical (constitutive/state equations) laws are imposed.
\begin{enumerate}    
    \item Hazen-Williams model: This corresponds to the constraints:
    
\begin{subequations}
\label{eq::MB_HW}
\begin{align}
(\Delta p)_1 &= \lambda \left(\frac{v_1 \Sigma_1}{\kappa_1}\right)^\alpha \Phi_1^\beta \delta_1, \label{eq::MB_HWa} \\
(\Delta p)_2 &= \lambda \left(\frac{v_2 \Sigma_2}{\kappa_2}\right)^\alpha \Phi_2^\beta \delta_2, \label{eq::MB_HWb} \\
(\Delta p)_e &= \frac{1}{2}\rho q^2\left[\left(\frac{1}{\Sigma_2^2} - \frac{1}{\Sigma_2^2}\right) + \xi\left(\frac{1}{\Sigma_1} - \frac{1}{\Sigma_2}\right)^2\right]. \label{eq::MB_HWc}
\end{align}
\end{subequations}

    \item Darcy-Weisbach model: This corresponds to the constraints:
    
\begin{subequations}
\label{eq::MB_DW}
\begin{align}
(\Delta p)_1 &= \frac{\rho}{2}f_{D_1} \frac{v_1^2}{\Phi_1}, \label{eq::MB_DWa} \\
(\Delta p)_2 &= \frac{\rho}{2}f_{D_2} \frac{v_2^2}{\Phi_2}, \label{eq::MB_DWb} \\
(\Delta p)_e &= \frac{1}{2}\rho q^2\left[\left(\frac{1}{\Sigma_2^2} - \frac{1}{\Sigma_2^2}\right) + \xi\left(\frac{1}{\Sigma_1} - \frac{1}{\Sigma_2}\right)^2\right]. \label{eq::MB_DWc}
\end{align}
\end{subequations}

Note that in that case, for the laminar regime, $f_{D_i} = \frac{64 \nu}{v_i \Phi_i}$ and $f_{D_i}$ are constant, while they depend on the pipe roughness in the rough turbulent regime.
    \end{enumerate}
\end{enumerate}

In the model-free network, the network topology is prescribed as:
\begin{align}
\bs{y}_1 & = \bs{y}_0\bs{W}_1+\bs{b}_1, \qquad \bs{y}_2 = \mathrm{ReLU}(\bs{y}_1\bs{W}_2+\bs{b}_2) \nonumber \\
\bs{y}_3 & = \mathrm{ReLU}(\bs{y}_2\bs{W}_3+\bs{b}_3), \qquad \bs{y}_4 = \mathbf{y}_3\bs{W}_4+\bs{b}_4
\end{align}
with $\bs{y}_0 = \bs{x} = q$ and $\bs{y}_4 = \bs{y} = ((\Delta p)_1, (\Delta p)_e, (\Delta p)_2)$. As before, a PIL is prescribed for the variables $\bs{y}_1$, that will be identified with $v_1$ and $v_2$ while the mass conservation is imposed via the constraint $v_i - \frac{q}{\Sigma_i} = 0$. Layers 2 and 3 are composed of $n_1 = n_2 = 15$ neurons.

Similarly, in the model-based network, we propose the following topology:
\begin{equation}
\bs{y}_1 = \bs{y}_0\bs{W}_1+\bs{b}_1, \qquad \bs{y}_2 = \bs H(\bs{y}_1;\boldsymbol \lambda)
\end{equation}
where $\bs H$ is the model equation, formulated in terms of the model parameters $\boldsymbol \lambda$, which are defined as $\lambda_1 = \xi$, $\lambda_2 = \Phi_1^\beta/\kappa_1$ and $\lambda_3 = \Phi_2^\beta/\kappa_2$ for the Hazen-Williams model and $\lambda_1 = \xi$, $\lambda_2 = \nu/\Phi_1$ and $\lambda_3 = \nu/\Phi_2$ for the Darcy-Weisbach model (in the laminar regime).

\begin{figure}
\centering
\begin{subfigure}{1.0\textwidth}
\centering
\begin{tikzpicture}

\foreach \m/\l [count=\y] in {1}
  \node [every neuron/.try, neuron \m/.try] (input-\m) at (0,0) {};
\foreach \m [count=\y] in {1,2,3}
  \node [every neuron/.try, neuron \m/.try] (velocity-\m) at (2,3-\y*1.5) {};
\foreach \m [count=\y] in {1,missing,2}
  \node [every neuron/.try, neuron \m/.try] (hiddena-\m) at (4,4-\y*2) {};
\foreach \m [count=\y] in {1,missing,2}
  \node [every neuron/.try, neuron \m/.try] (hiddenb-\m) at (6,4-\y*2) {}; 
\foreach \m [count=\y] in {1,2,3}
  \node [every neuron/.try, neuron \m/.try] (output-\m) at (8,3-\y*1.5) {};

\foreach \l [count=\i] in {1}
  \draw [<-] (input-\i) -- ++(-1,0)
    node [above, midway] {$q$};

\foreach \l [count=\i] in {1,2,3}
  \node [above] at (velocity-\i.north) {$y_{1,\l}$};
\foreach \l [count=\i] in {1,15}
  \node [above] at (hiddena-\i.north) {$y_{2,\l}$};
\foreach \l [count=\i] in {1,15}
  \node [above] at (hiddenb-\i.north) {$y_{3,\l}$};

\draw [->] (output-1) -- ++(1,0)
    node [above, midway] {$(\Delta p)_1$};
\draw [->] (output-2) -- ++(1,0)
    node [above, midway] {$(\Delta p)_e$};
\draw [->] (output-3) -- ++(1,0)
    node [above, midway] {$(\Delta p)_2$};

\foreach \i in {1}
  \foreach \j in {1,2,3}
    \draw [->] (input-\i) -- (velocity-\j);   
\foreach \i in {1,2,3}
  \foreach \j in {1,2}
    \draw [->] (velocity-\i) -- (hiddena-\j);
\foreach \i in {1,2}
  \foreach \j in {1,2}
    \draw [->] (hiddena-\i) -- (hiddenb-\j);
\foreach \i in {1,2}
  \foreach \j in {1,2,3}
    \draw [->] (hiddenb-\i) -- (output-\j);

\foreach \l [count=\x from 0] in {Input, Velocity, Hidden, Hidden, Output}
  \node [align=center, above] at (\x*2,3) {\l \\ layer};
  
\draw[red,thick,dashed]  (1.5,-2) rectangle (2.5,2.5);
\end{tikzpicture}
\caption{Model-free neural network.}
\label{fig::MF_NN}
\end{subfigure} \\%

\begin{subfigure}{1.0\textwidth}
\centering
\begin{tikzpicture}

\foreach \m/\l [count=\y] in {1}
  \node [every neuron/.try, neuron \m/.try] (input-\m) at (0,0) {};
\foreach \m [count=\y] in {1,2,3}
  \node [every neuron/.try, neuron \m/.try] (velocity-\m) at (4,3-\y*1.5) {};
\foreach \m [count=\y] in {1,2,3}
  \node [every neuron/.try, neuron \m/.try] (output-\m) at (8,3-\y*1.5) {};

\foreach \l [count=\i] in {1}
  \draw [<-] (input-\i) -- ++(-1,0)
    node [above, midway] {$q$};

\foreach \l [count=\i] in {1,2,3}
  \node [above] at (velocity-\i.north) {$y_{1,\l}$};

\draw [->] (output-1) -- ++(1,0)
    node [above, midway] {$(\Delta p)_1$};
\draw [->] (output-2) -- ++(1,0)
    node [above, midway] {$(\Delta p)_e$};
\draw [->] (output-3) -- ++(1,0)
    node [above, midway] {$(\Delta p)_2$};

\foreach \i in {1}
  \foreach \j in {1,2,3}
    \draw [->] (input-\i) -- (velocity-\j);   
\foreach \i in {1,2,3}
    \draw [blue,->] (velocity-\i) -- (output-\i);

\foreach \l [count=\x from 0] in {Input, Velocity, Output}
  \node [align=center, above] at (\x*4,3) {\l \\ layer};
  
\draw[red,thick,dashed]  (3.5,-2) rectangle (4.5,2.5);
\end{tikzpicture}
\caption{Model-based neural network.}
\label{fig::MB_NN}
\end{subfigure}
\caption{\textbf{Model-free and model-based PGNNIV.} The universal constraints are illustrated using the red dashed boxes. Model-based constraints are illustrated using blue lines.}
\label{fig::enrichment_NN}
\end{figure}
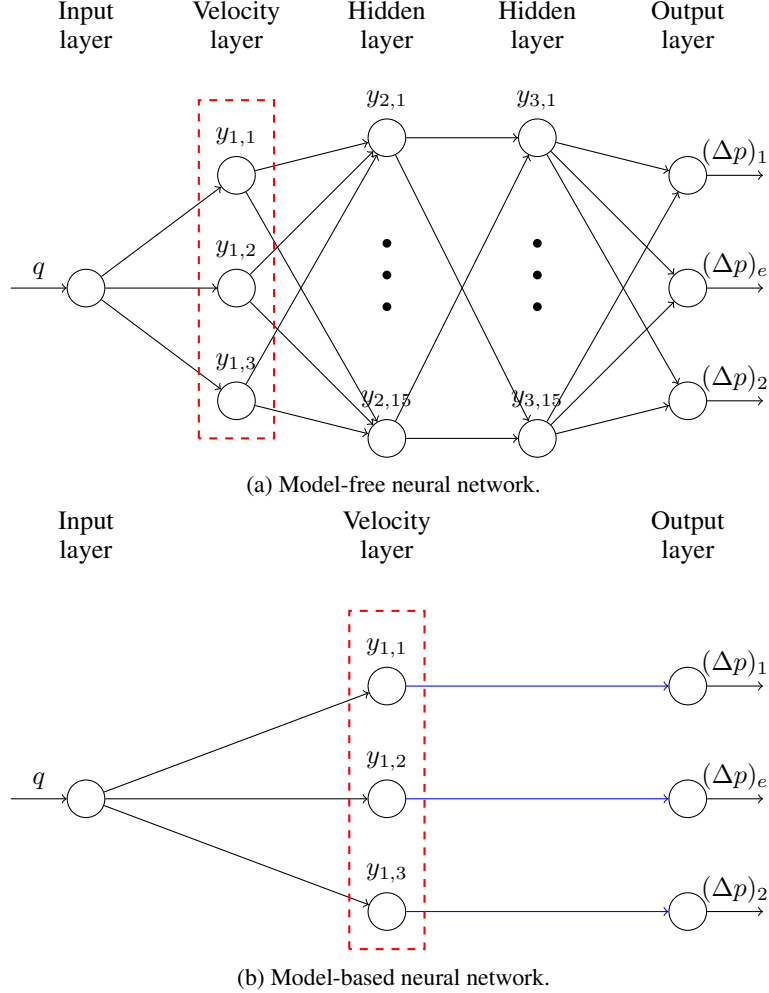

Both neural networks are illustrated in Fig. \ref{fig::enrichment_NN}.

For the training process, the data input was randomly generated with a uniform distribution using the state model presented in equation (\ref{eq::model_pQ}) for $q \in [1.0;5.0] \,  (\mathrm{m^3/s})$. The physical parameters used for the data generation are shown in Table \ref{table::param_Ph}. As a learning algorithm, a gradient descent optimizer was selected with learning rate $\beta=0.0001$. At each training step, $n=4$ data points are selected. For the constrained network, we chose a penalty parameter of $p=0.01 \, \mathrm{Pa^2 m^6 / s^2}$. $N_\mathrm{test} =  1000$ samples were randomly generated for the testing procedure.

To evaluate the performance of all neural networks, we illustrate in Table \ref{table::enrichment_1} the statistics of the relative error of the predicted value (when compared to the analytical one) for the different variables involved in the problem: i) Measurable variables (output variables), that is, the pressure drops $(\Delta p)_1$, $(\Delta p)_e$ and $(\Delta p)_2$; ii) Non-measurable variables (internal variables), that is, the flow velocity at each segment, $v_1$ and $v_2$.

\begin{figure}
\centering
\begin{subfigure}{.6\textwidth}
  \includegraphics[width=\linewidth]{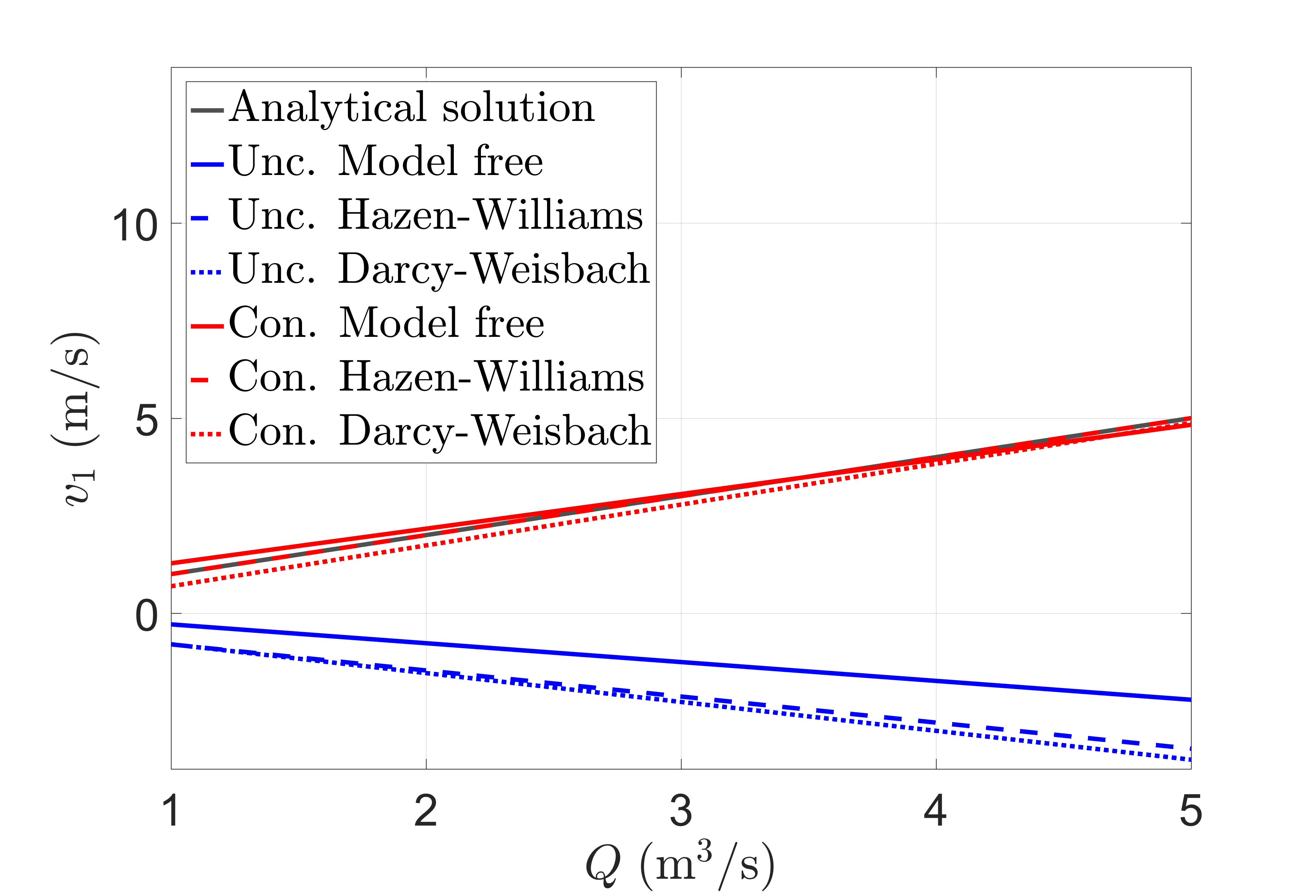}
  \caption{$v_1$.}
  \label{fig::enrichment_1a}
\end{subfigure} \\
\begin{subfigure}{.6\textwidth}
  \includegraphics[width=\linewidth]{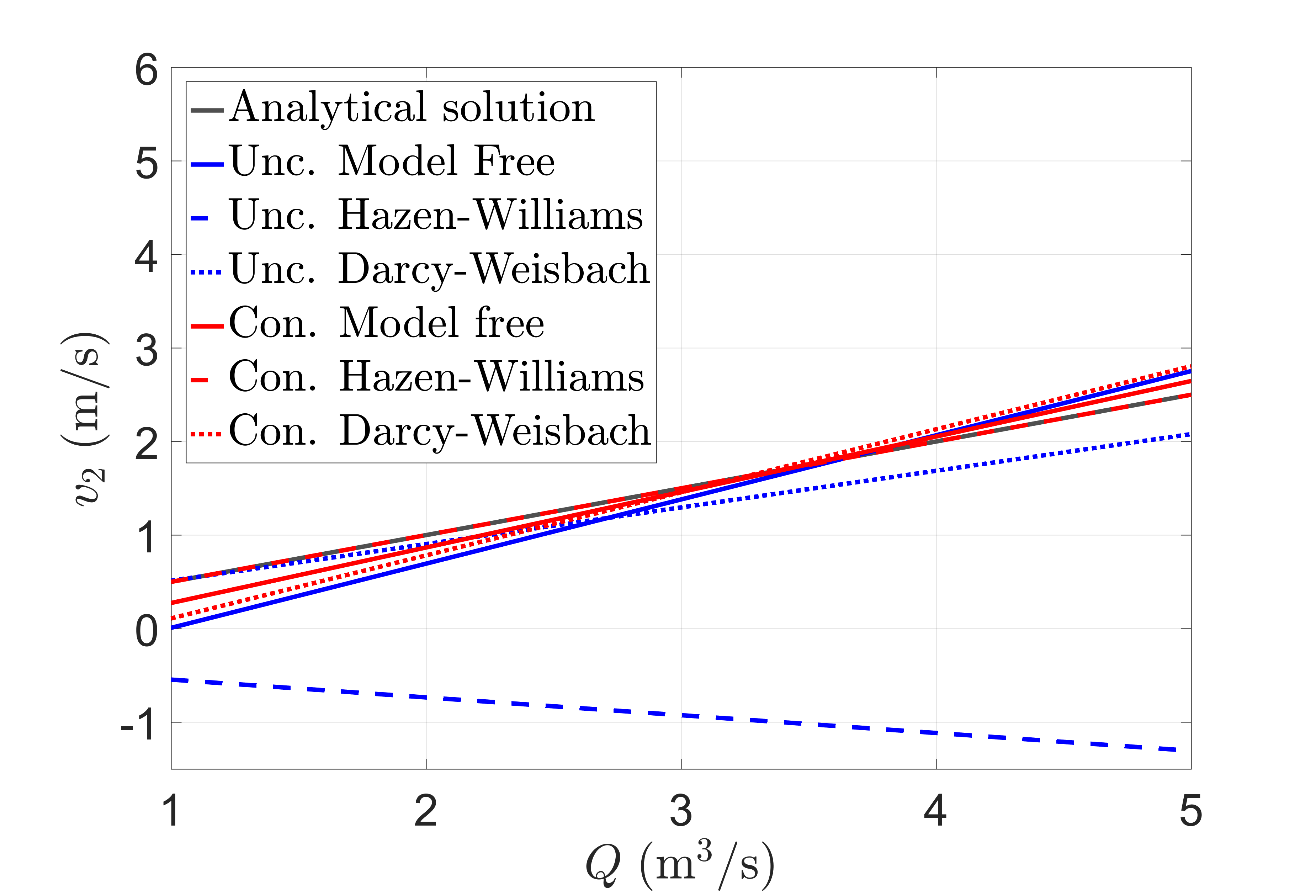}
  \caption{$v_2$.}
  \label{fig::enrichment_1b}
\end{subfigure}
\caption{\textbf{Predictive capacity of each neural network in estimating the internal variables.} The constrained network is the only one able to predict accurately the internal variables. Model specification improves the accuracy only when the model assumed is the correct one.}
\label{fig::enrichment_1}
\end{figure}

\begin{figure}
\centering
\begin{subfigure}{.48\textwidth}
  \includegraphics[width=\linewidth]{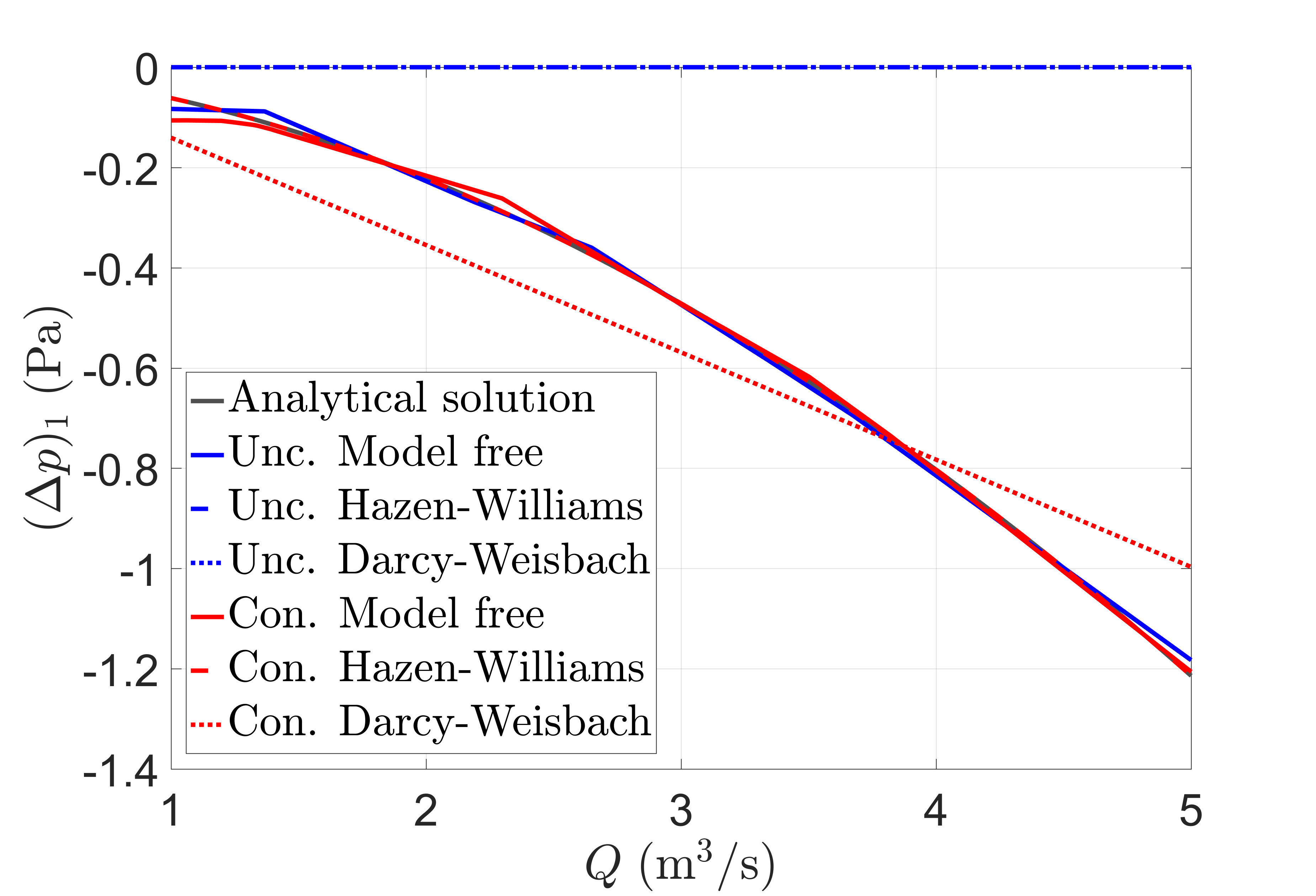}
  \caption{$(\Delta p)_1$.}
  \label{fig::enrichment_2a}
\end{subfigure} 
\begin{subfigure}{.48\textwidth}
  \includegraphics[width=\linewidth]{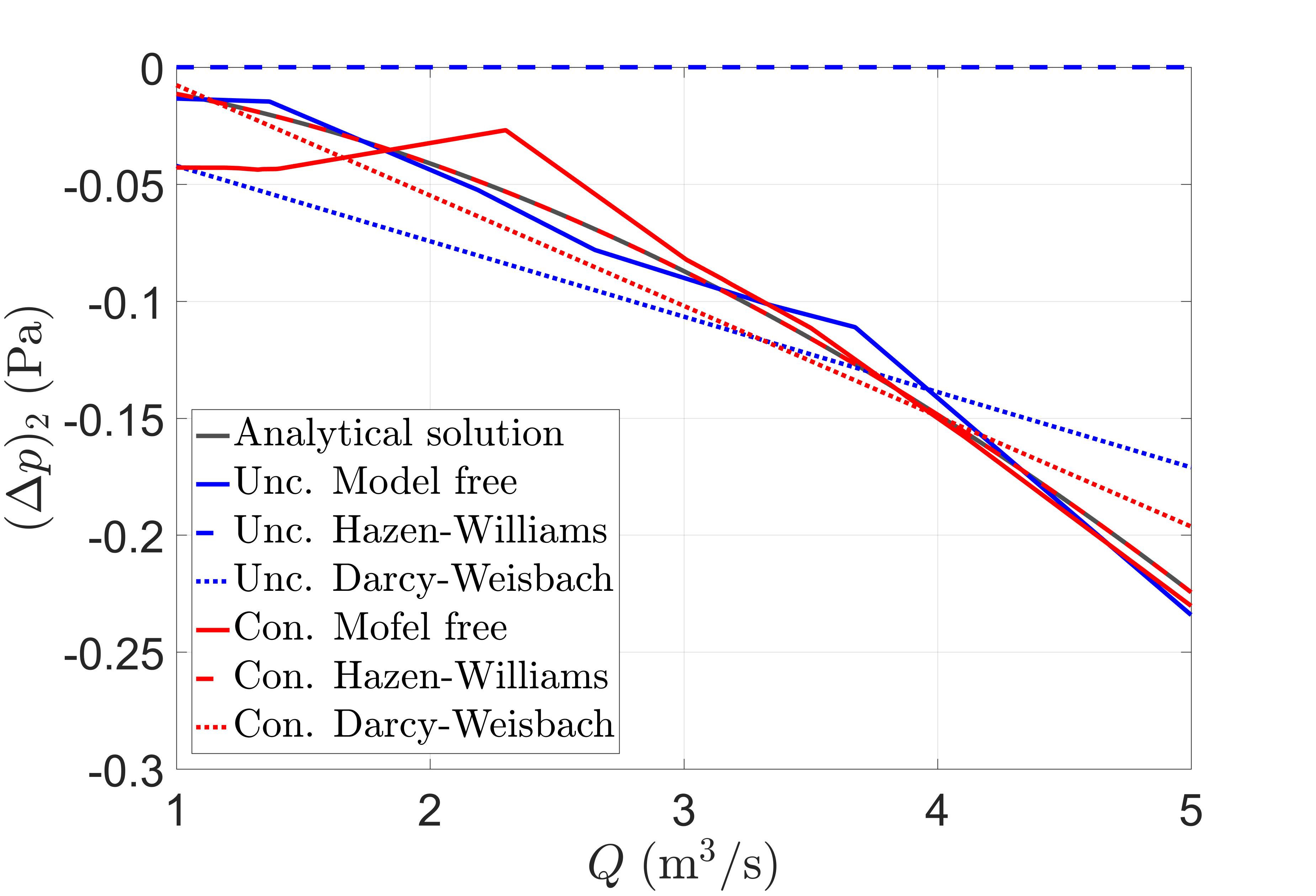}
  \caption{$(\Delta p)_2$.}
  \label{fig::enrichment_2c}
\end{subfigure} \\
\begin{center}
\begin{subfigure}{.48\textwidth}
  \includegraphics[width=\linewidth]{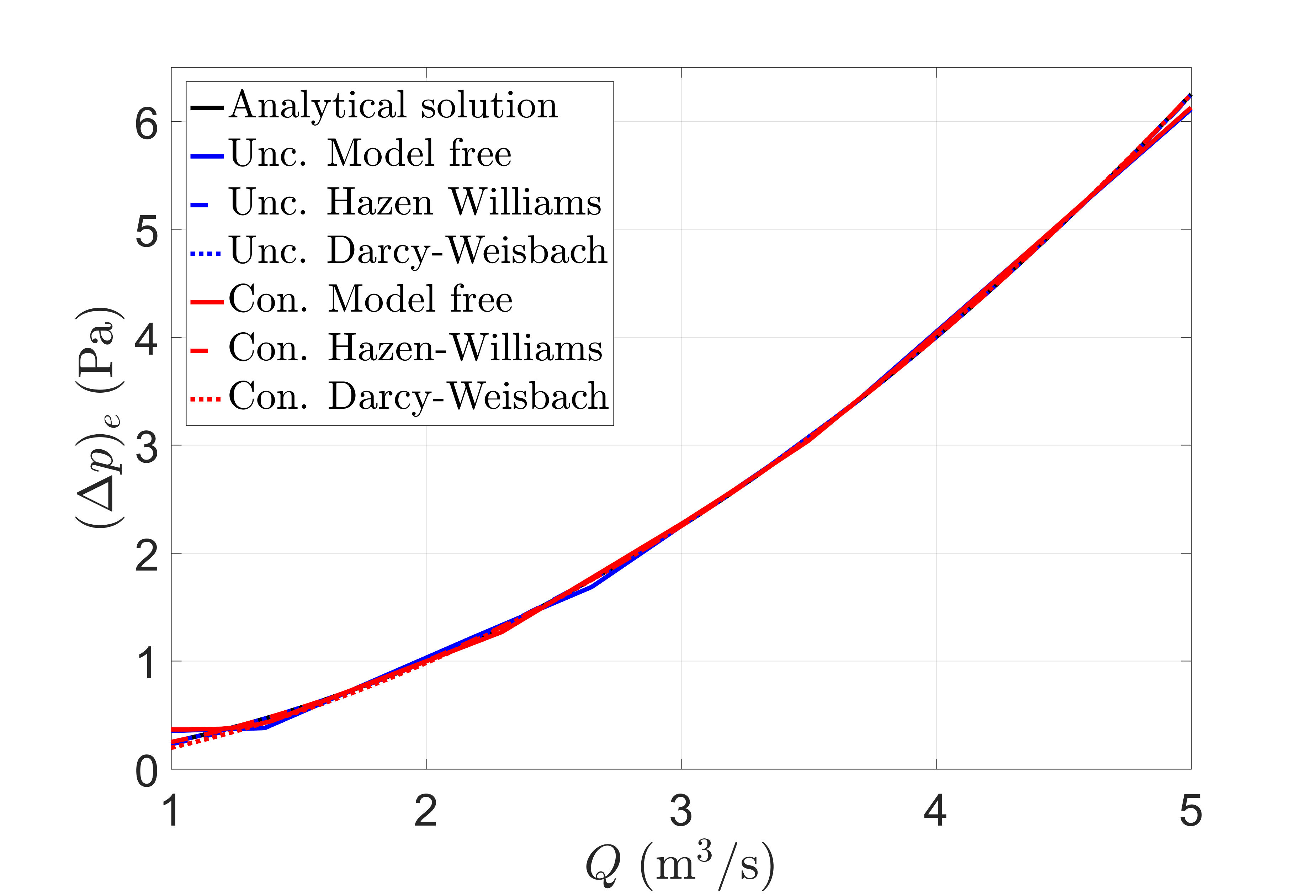}
  \caption{$(\Delta p)_e$.}
  \label{fig::enrichment_2b}
\end{subfigure}
\end{center}
\caption{\textbf{Predictive capacity of each neural network in estimating the measurable variables.} The unconstrained and constrained networks have a similar capacity in estimating the measurable variables. Model specification improves the accuracy only when the model assumed is the correct one.}
\label{fig::enrichment_2}
\end{figure}

\begin{figure}
\centering
\begin{subfigure}{.6\textwidth}
  \includegraphics[width=\linewidth]{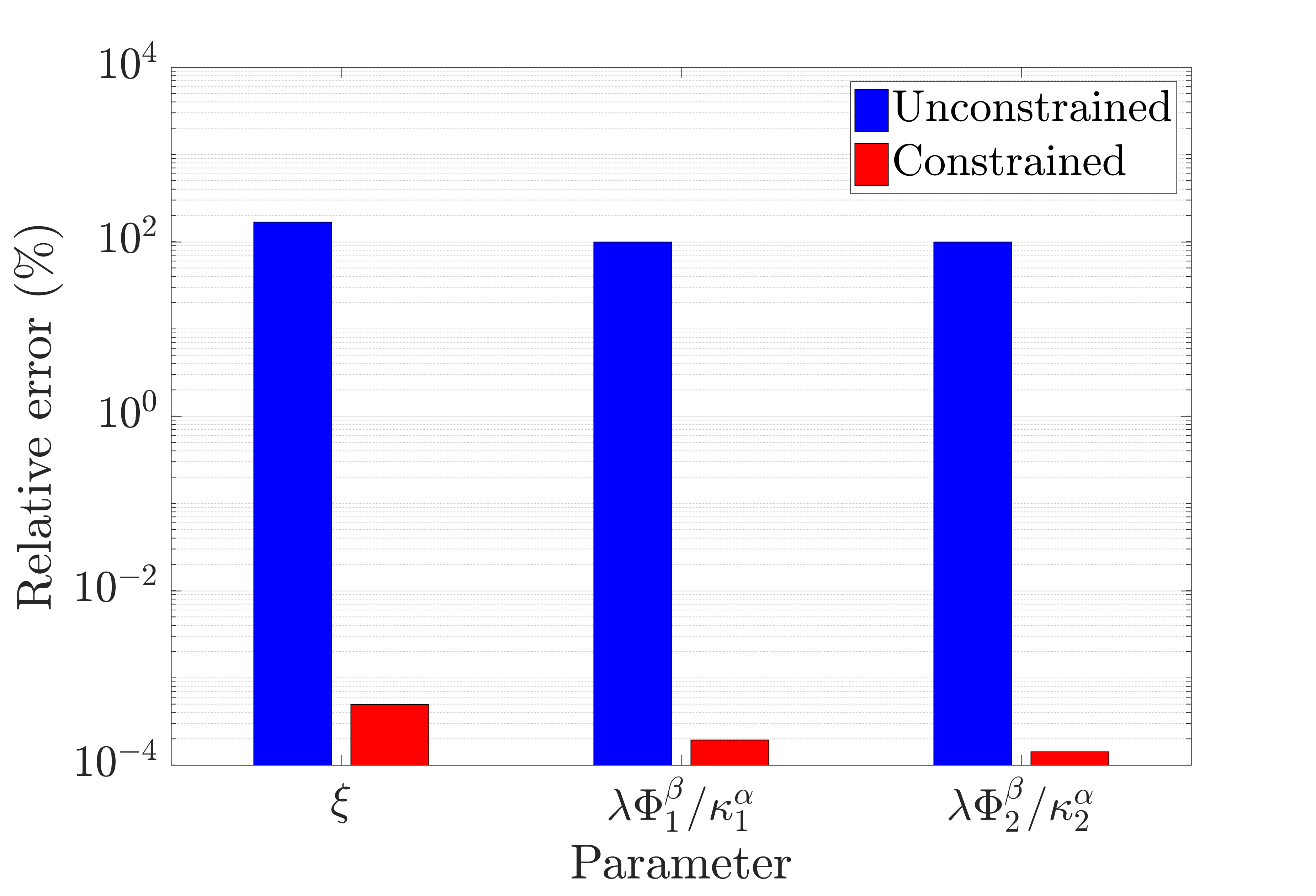}
  \caption{Hazen-Williams model-based.}
  \label{fig::enrichment_3a}
\end{subfigure} \\
\begin{subfigure}{.6\textwidth}
  \includegraphics[width=\linewidth]{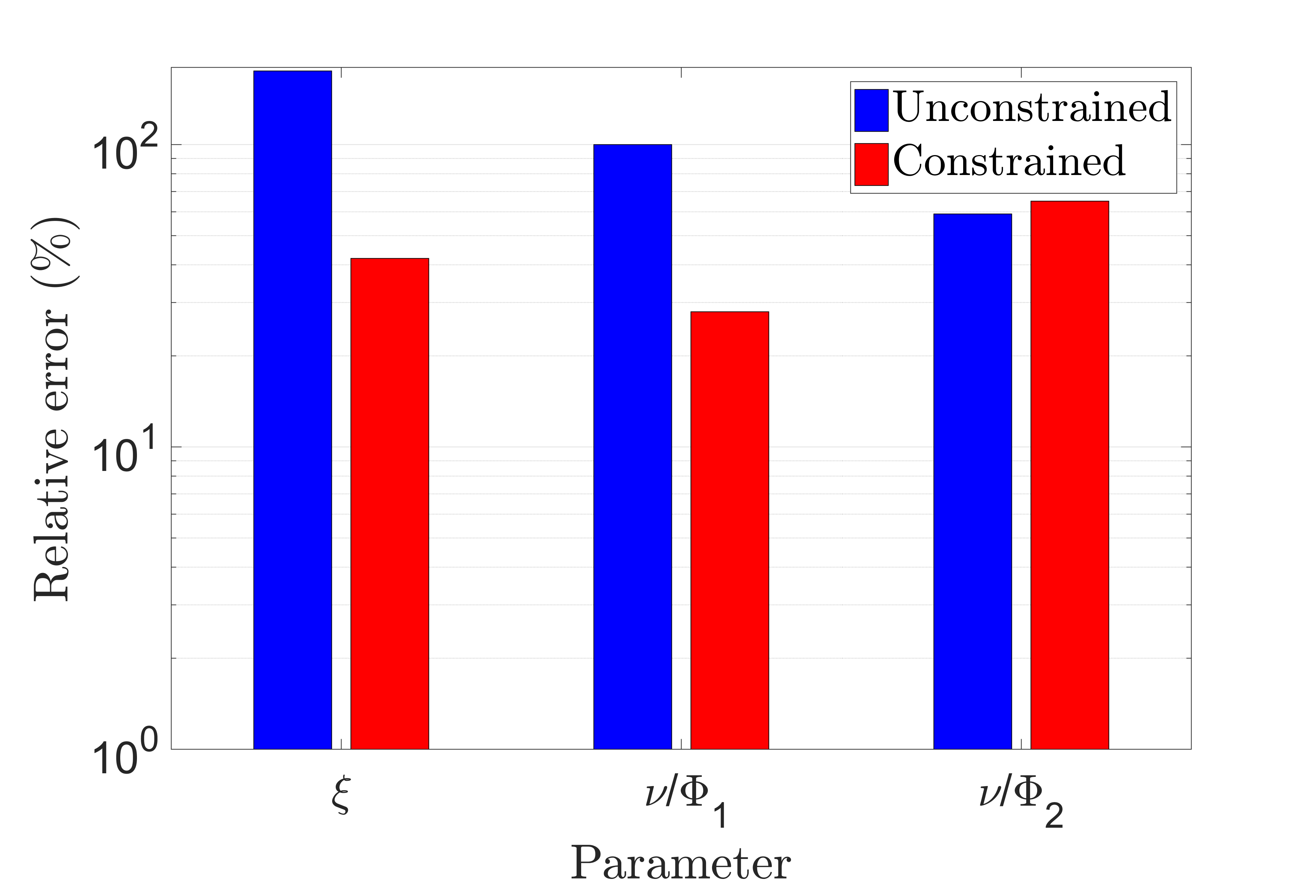}
  \caption{Darcy-Weisbach model-based.}
  \label{fig::enrichment_3b}
\end{subfigure} \\
\caption{\textbf{Explanatory capacity of each model-based network.} As the data-set was generated using the Hazen-Williams model, only this NN has a perfect explanatory capacity (relative error of the order $10^{-6}$) while the Darcy-Weisbach based network has only partial explanatory capacity.}
\label{fig::enrichment_3}
\end{figure}

Figs \ref{fig::enrichment_1} and \ref{fig::enrichment_2} illustrate the predictive capacity of the different networks in estimating the internal and measurable variables respectively for different values of $q$. Once the model-based network has converged, it is possible to extract the model parameters, whose relative error is illustrated in Table \ref{table::enrichment_2} and in Fig. \ref{fig::enrichment_3}.

The conclusion drawn is clear and natural. If we want a predictive capability, a model-free neural network is always preferred except if the underlying constitutive model is perfectly known (what is, in general, a strong assumption). A wrong model assumption worsens the network accuracy with respect to a model-free one. Therefore, model-based networks can help in model identification and can shed light about the system physical and geometrical structure. The specification of an incorrect underlying model affects both correctly specified variables ($(\Delta p)_e$) and those that are not ($(\Delta p)_i$, $i=1,2$) as the error is distributed in all the predicted variables. 

\begin{table}[!htbp]
\caption{\textbf{Statistics of the relative error $\varepsilon_r$ for the different networks analyzed.} Data is presented as mean value ($\pm$ Std. error). MF: Model-Free. MB (HW): Hazen-Williams model-based. MB (DW): Darcy-Weisbach model-based. Errors behind $1\permil$ are not reported.}
\centering
{\footnotesize
\begin{tabular}{llccccc}
        &      & \multicolumn{3}{c}{Measurable variables}                          & \multicolumn{2}{c}{Internal variables}    \\ \cline{3-7} 
        &      & $(\Delta p)_1$       & $(\Delta p)_e$      & $(\Delta p)_2$       & $v_1$               & $v_2$               \\ \hline
MF      & Unc. & $-0.031 \quad (\pm 0.002)$ & $0.030 \quad (\pm 0.002)$ & $-0.070 \quad (\pm 0.002)$ & $1.405 \quad (\pm 0.001)$ & $0.211 \quad (\pm 0.007)$ \\
        & \textbf{Con.} & $-0.049 \quad (\pm 0.003)$ & $0.023 \quad (\pm 0.002)$ & $-0.03 \quad (\pm 0.02)$   & $0.059 \quad (\pm 0.002)$ & $0.096 \quad (\pm 0.003)$ \\ \hline
MB (HW) & Unc. & $-1.000 \quad (\pm 0.000)$ & $0.008 \quad (\pm 0.000)$ & $-1.000 \quad (\pm 0.000)$ & $1.723 \quad (\pm 0.000)$ & $1.666 \quad (\pm 0.004)$ \\
        & \textbf{Con.} & $0.000 \quad (\pm 0.000)$  & $0.000 \quad (\pm 0.000)$ & $0.000 \quad (\pm 0.000)$  & $0.000 \quad (\pm 0.000)$ & $0.000 \quad (\pm 0.000)$ \\ \hline
MB (DW) & Unc. & $-1.000 \quad (\pm 0.000)$ & $0.003 \quad (\pm 0.000)$  & $-0.56 \quad (\pm 0.02)$   & $1.764 \quad (\pm 0.000)$ & $0.120 \quad (\pm 0.001)$ \\
        & \textbf{Con.}. & $-0.37 \quad (\pm 0.01)$   & $0.021 \quad (\pm 0.001)$ & $-0.165 \quad (\pm 0.004)$ & $0.097 \quad (\pm 0.002)$ & $0.165 \quad (\pm 0.006)$ \\ \hline
\end{tabular}
}
\label{table::enrichment_1}
\end{table}

\begin{table}[!htbp]
\caption{\textbf{Relative error $\varepsilon_r$ of the model parameters for the different model-based networks.} MB (HW): Hazen-Williams Model-Based. MB (DW): Darcy-Weisbach Model-Based. Relative error below $10^{-2}$ (MB (HW) constrained) is marked as $0$ in the table because it is of the order $10^{-6}$.}
\centering
\begin{tabular}{llccc}
        &      			& \multicolumn{3}{c}{Parameter}           \\ \cline{3-5} 
        &      			& $\lambda_1$ & $\lambda_2$ & $\lambda_3$ \\ \hline
MB (HW) & Unc. 			& $1.68$      & $1.00$      & $0.99$      \\
        & \textbf{Con.} & $0.00$      & $0.00$      & $0.00$      \\ \hline
MB (DW) & Unc. 			& $1.75$      & $1.00$      & $0.58$      \\
        & \textbf{Con.} & $0.42$      & $0.27$      & $0.65$      \\ \hline
\end{tabular}
\label{table::enrichment_2}
\end{table}

\subsection{Results of the characterization approach}
\label{sec::characterization_results}

Recall that the aim is now to characterize some of the parameters of the pipe segments for a given set of values $(q,p_0,p_1,p_2)$, that is, the relationship to be learned now is $(q,p_0,p_1,p_2) \rightarrow (\kappa_1,\kappa_2)$.

As Eq. (\ref{eq::true_kappa_i}) is complex and highly nonlinear, it is expected that the number of hidden layers required will be large. Thus, we refer to this part of the network as a \emph{Deep Learning Box} with its own internal number of layers, neurons and connectivity.

\begin{figure}
\centering
\begin{tikzpicture}
\centering

\foreach \m [count=\y] in {1,2,3,4}
  \node [every neuron/.try, neuron \m/.try] (input-\m) at (0,5-\y*2) {};
\foreach \m [count=\y] in {1,2,3,4}
  \node [every neuron/.try, neuron \m/.try,fill=white] (hidden-\m) at (3,5-\y*2) {};
\foreach \m [count=\y] in {1,2}
  \node [every neuron/.try, neuron \m/.try,fill=white] (output-\m) at (9,3-\y*2) {};
\draw[red,thick,dashed] (3.5,0.5) rectangle (2.5,4);
\draw[red,thick,dashed] (3.5,-4) rectangle (2.5,0);

\foreach \l [count=\i] in {1}
  \draw [<-] (input-\i) -- ++(-1,0)
    node [above, midway] {$Q$};
\draw [<-] (input-2) -- ++(-1,0)
node [above, midway] {$p_0$};
\foreach \l [count=\i] in {3,4}
  \draw [<-] (input-\l) -- ++(-1,0)
    node [above, midway] {$p_\i$};

\foreach \l [count=\i] in {1,2}
  \node [above] at (hidden-\l.north) {$v_\l$};
\foreach \l [count=\i] in {3,4}
  \node [above] at (hidden-\l.north) {$(\Delta p)_\i$};  
\foreach \l [count=\i] in {1,2}
  \node [above] at (output-\i.north) {$\kappa_{\l}$};
\draw[black,thick] (5,-4) rectangle (7,4);

\foreach \i in {1}
  \foreach \j in {1,2}
    \draw [->] (input-\i) -- (hidden-\j);
\draw [->] (input-2) -- (hidden-3);
\draw [->] (input-3) -- (hidden-3);
\draw [->] (input-3) -- (hidden-4);
\draw [->] (input-4) -- (hidden-4);
\foreach \l [count=\i] in {1,2,3,4}
	\draw [->] (hidden-\i) -- (5,5-\i*2);
\draw [->] (7,2) -- (output-1);
\draw [->] (7,-2) -- (output-2);

\foreach \l [count=\x from 0] in {Input, State variables, Deep Learning Box, Output}
  \node [align=center, above] at (\x*3,4.5) {\l \\ layer};
\end{tikzpicture}
\caption{\textbf{PGNNIV for the characterization problem.} Red dashed rectangles represent physical constraints on neurons. The relationship between flow and velocities is imposed in $v_1$ and $v_2$, $v_i = \frac{q}{\Sigma_i}$ and the definition of the incremental pressure drop is imposed in $(\Delta p)_1$ and $(\Delta p)_2$, $(\Delta p)_i = p_{i} - p_{i-1}$.}
\label{fig::kappadNN}
\end{figure}
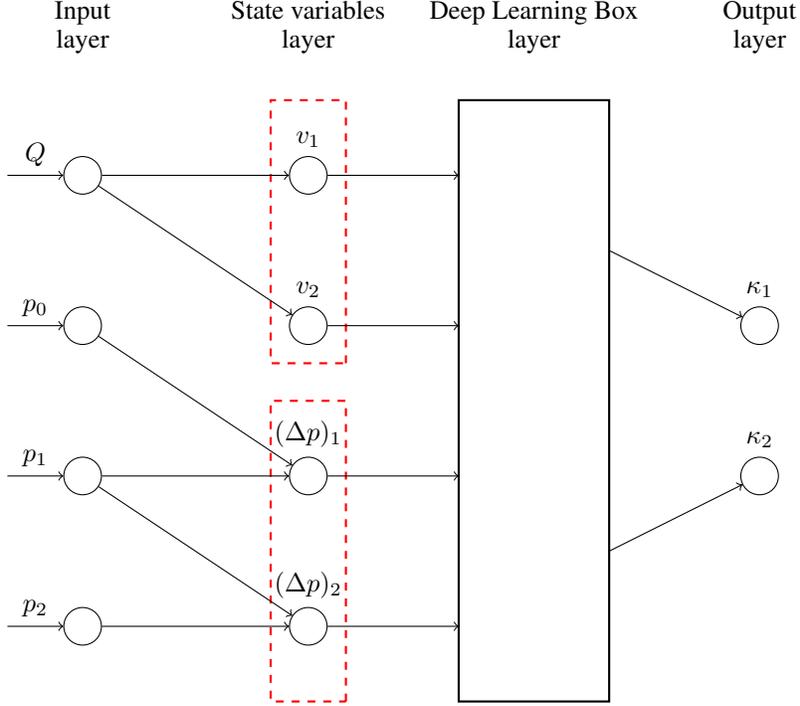

For the problem presented, the PGNNIV topology is shown in Fig. \ref{fig::kappadNN}. The Deep Learning Box is a multilayer perceptron with 5 dense layers of 20, 40, 80, 40 and 20 neurons respectively, with activation functions of ReLU type. The network performance is compared to the same network in which the physical constraint has not been included. The difference between the constrained and the unconstrained networks is that the penalty parameter is set to zero for the unconstrained network.

The training data-set was created using the analytical model, with $\kappa_1$ and $\kappa_2$ randomly generated between $\kappa = 80$ and $\kappa = 140$ (that are standard values of the roughness parameter) and a flow $q$ varying from $1 \, \mathrm{m^3/s}$ to $5 \, \mathrm{m^3/s}$. For the training process, we used batches of $n=300$, a penalty parameter of $p=0.001$ and a learning rate parameter of $\beta = 1 \times 10^{-5}$ for the gradient descent optimizer. The input $x_i$ and output $y_i$ values are normalized between their maximal and minimal value as:

\begin{equation}
\hat{x}_i = \frac{x_i - x_\mathrm{min}}{x_\mathrm{max} - x_\mathrm{min}}, \qquad
\hat{y}_i = \frac{y_i - y_\mathrm{min}}{y_\mathrm{max} - y_\mathrm{min}}
\end{equation} 
where $x_\mathrm{max}$ and $x_\mathrm{min}$ are the maximal and minimal values for the input and $y_\mathrm{max}$ and $y_\mathrm{min}$ the maximal and minimal values for the output, respectively. $N_\mathrm{test} = 100$ test values are used to evaluate the performance.

\begin{figure}[htbp]
\centering
\begin{subfigure}{.6\textwidth}
  \centering
  \includegraphics[width=\linewidth]{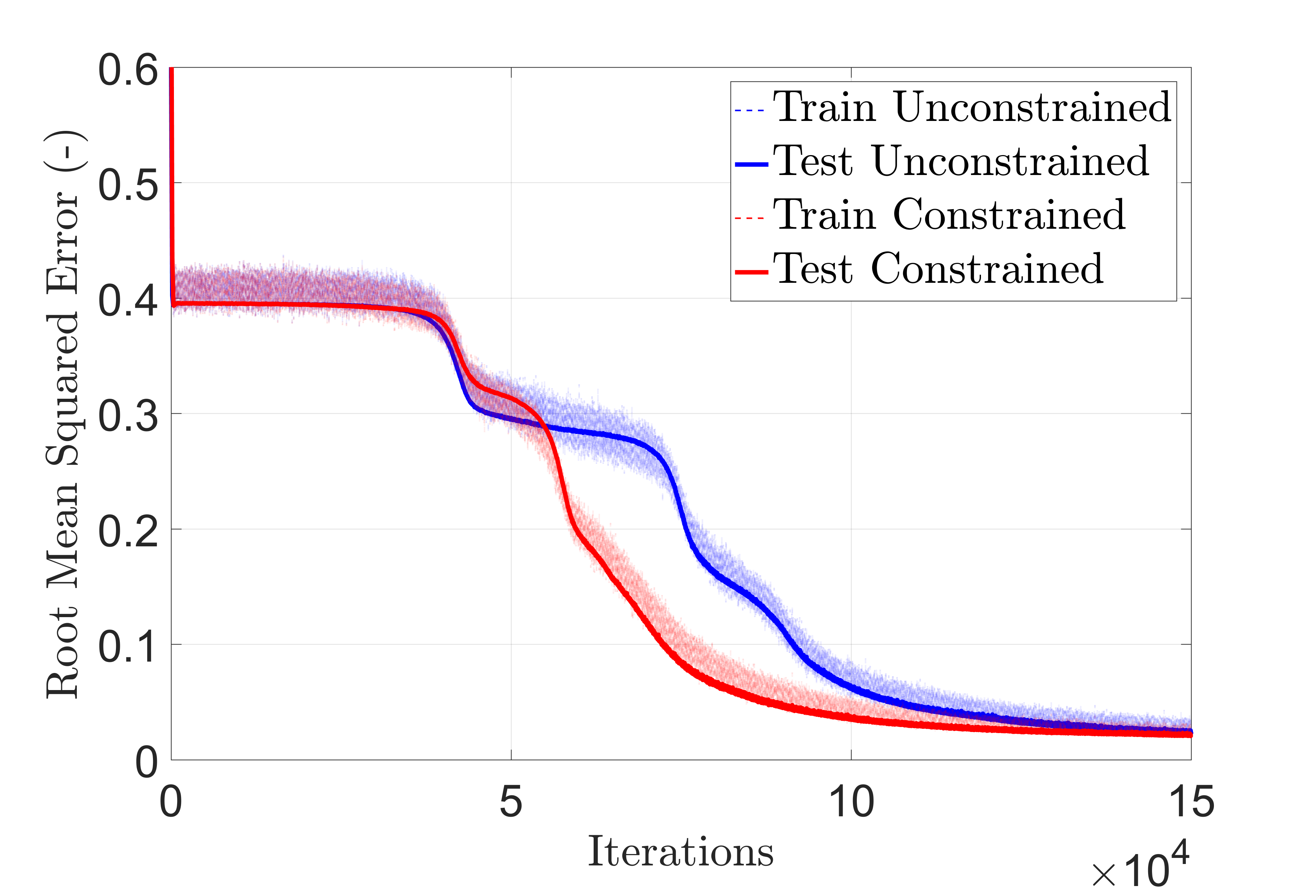}
  \caption{Root Mean Squared Error (RMSE).}
  \label{fig::characterization_1b}
\end{subfigure} \\
\begin{subfigure}{.6\textwidth}
  \centering
  \includegraphics[width=\linewidth]{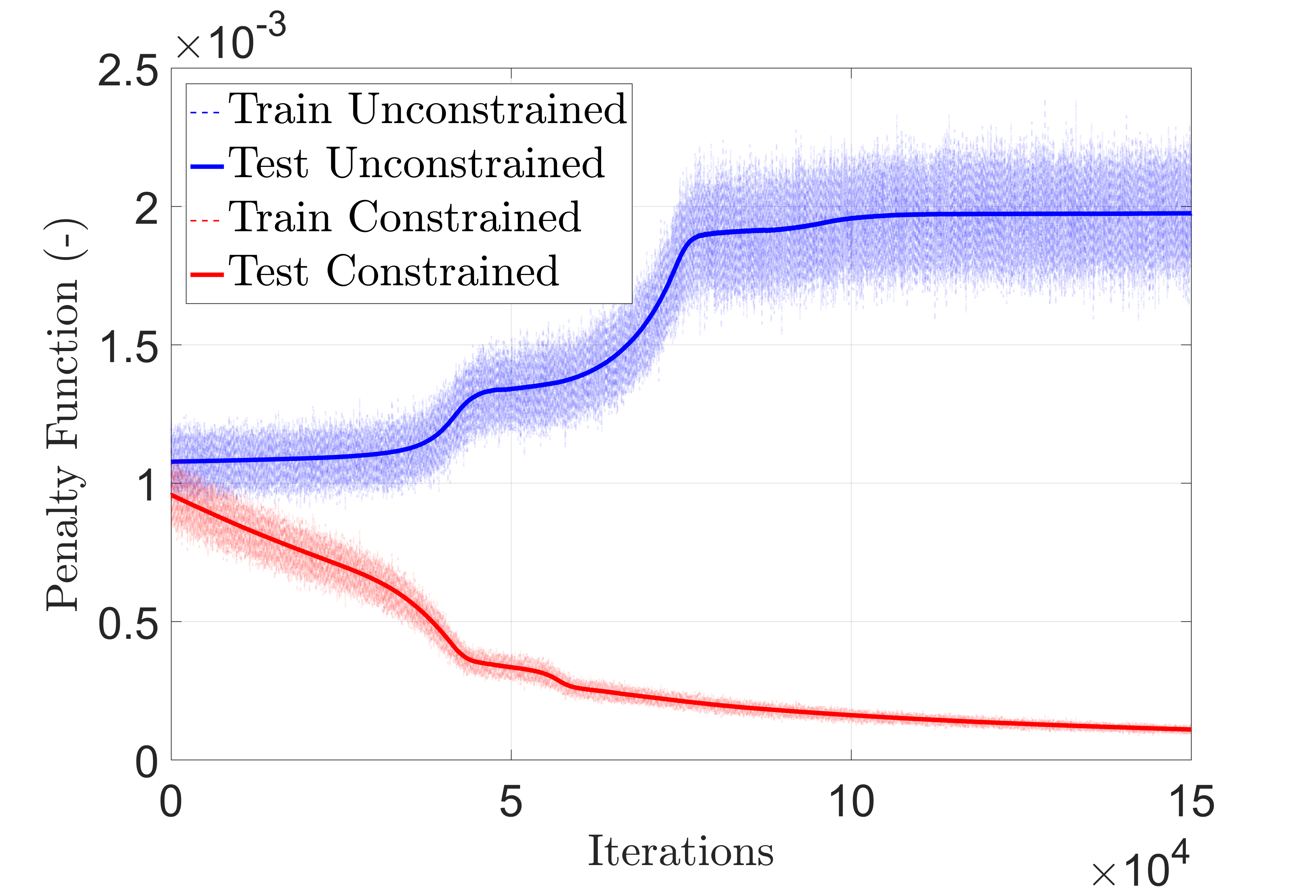}
  \caption{Penalty function (PEN).}
  \label{fig::characterization_1c}
\end{subfigure}
\caption{\textbf{Root Mean Squared Function and penalty function.}  Even if, in all cases, the Deep Learning Box has not enough power to capture the complex nonlinear model perfectly (observe that RMSE does not converge to $0$) the effect of the penalty is to speed-up the network convergence.}
\label{fig::characterization_1}
\end{figure}

As for the prediction problem, Fig. \ref{fig::characterization_1} shows the performance of both neural networks for the characterization one. As in the previous case, the constraints accelerate the convergence of the network.

\begin{figure}[htbp]
\centering
\includegraphics[width=0.8\linewidth]{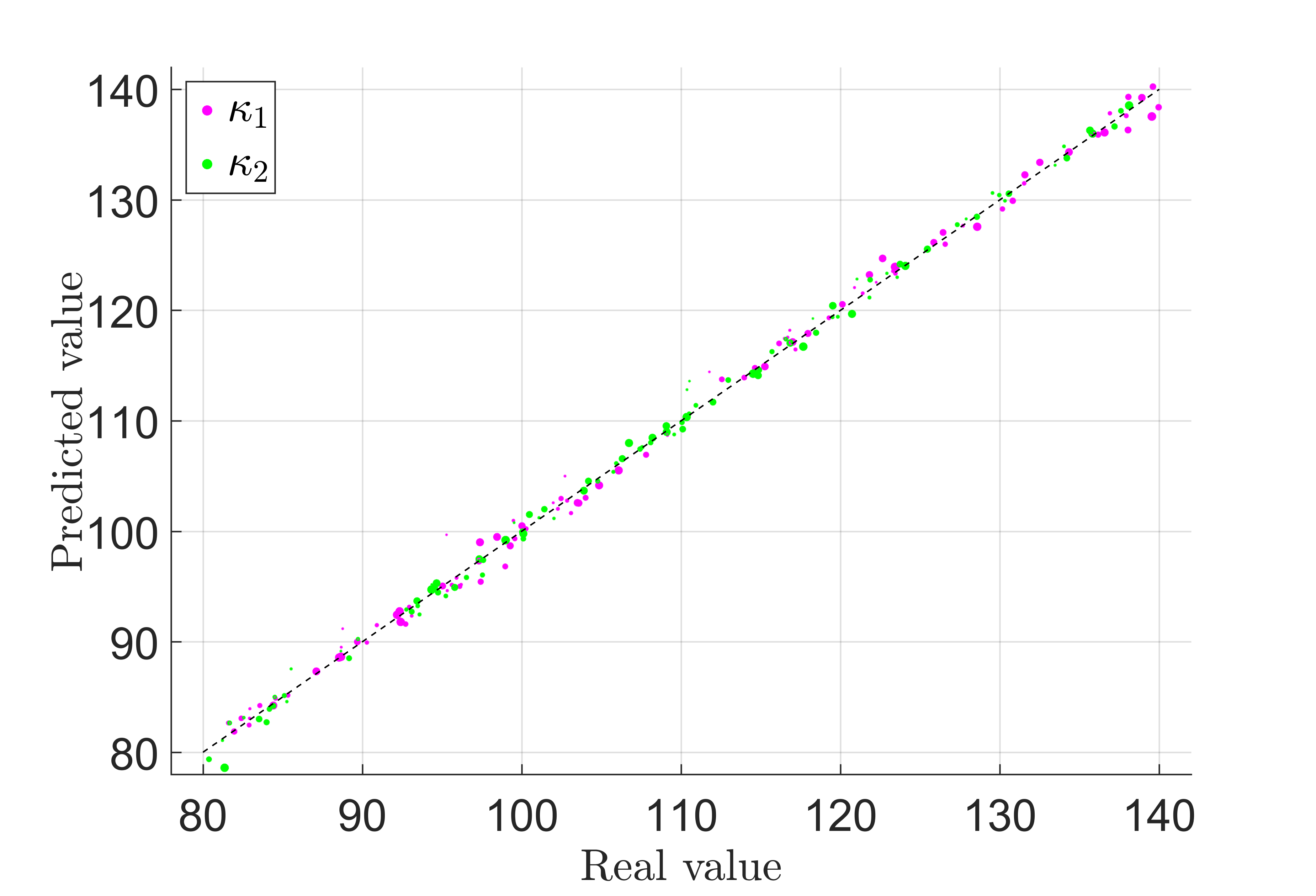}
\caption{\textbf{Results predicted by the PGNNIV.}  Both predicted values $\kappa_1$ and $\kappa_2$ are compared to the theoretical ones. Line $y=x$ identifies a perfect estimation. The size of the dot is proportional to the value of the input flow: the model tends to give worse results for smaller flows.}
\label{fig::characterization_2}
\end{figure}

The accuracy is shown in Fig. \ref{fig::characterization_2} where the predicted values of $\kappa_1$ and $\kappa_2$ are compared with the theoretical ones for $N_{\mathrm{test}}=100$ test values. The figure shows a good performance of the neural network although it decays close to the boundaries, what is natural, since the neural network has been trained with data (roughness coefficient) $\kappa \in [80;140]$.

\clearpage

\section{Discussion} \label{secc::Disc}

\subsection{Performance improvement}

The first important property of the presented methodology, beyond its explanatory capacity, is the improvement of the performance with respect to other Neural-Network based methods. As PGNNIV has internal constraints between layers (or, equivalently, a higher number of outputs) it is obvious that the search space for the optimal solution will be smaller. This observation leads to four important consequences that are quantified next.

\subsubsection{Convergence acceleration} 

\begin{figure}[htbp]
\centering
\begin{subfigure}{.49\textwidth}
		\centering
		\includegraphics[width=\linewidth, trim =0 0 0 0,clip]{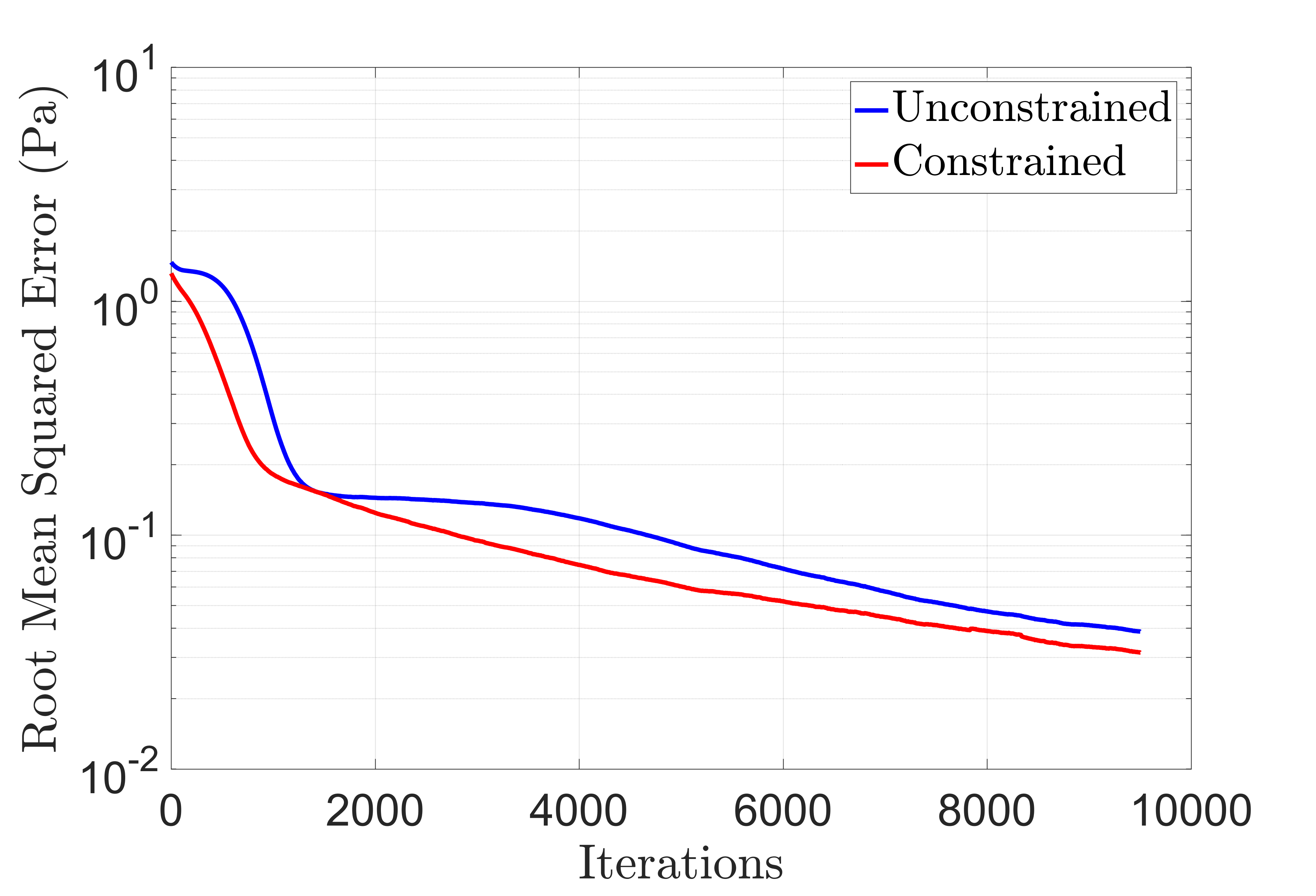}
		\caption{Prediction problem.}
		\label{fig::acceleration_1a}
\end{subfigure}
\begin{subfigure}{0.49\textwidth}
		\centering
		\includegraphics[width=\linewidth, trim =0 0 0 0,clip]{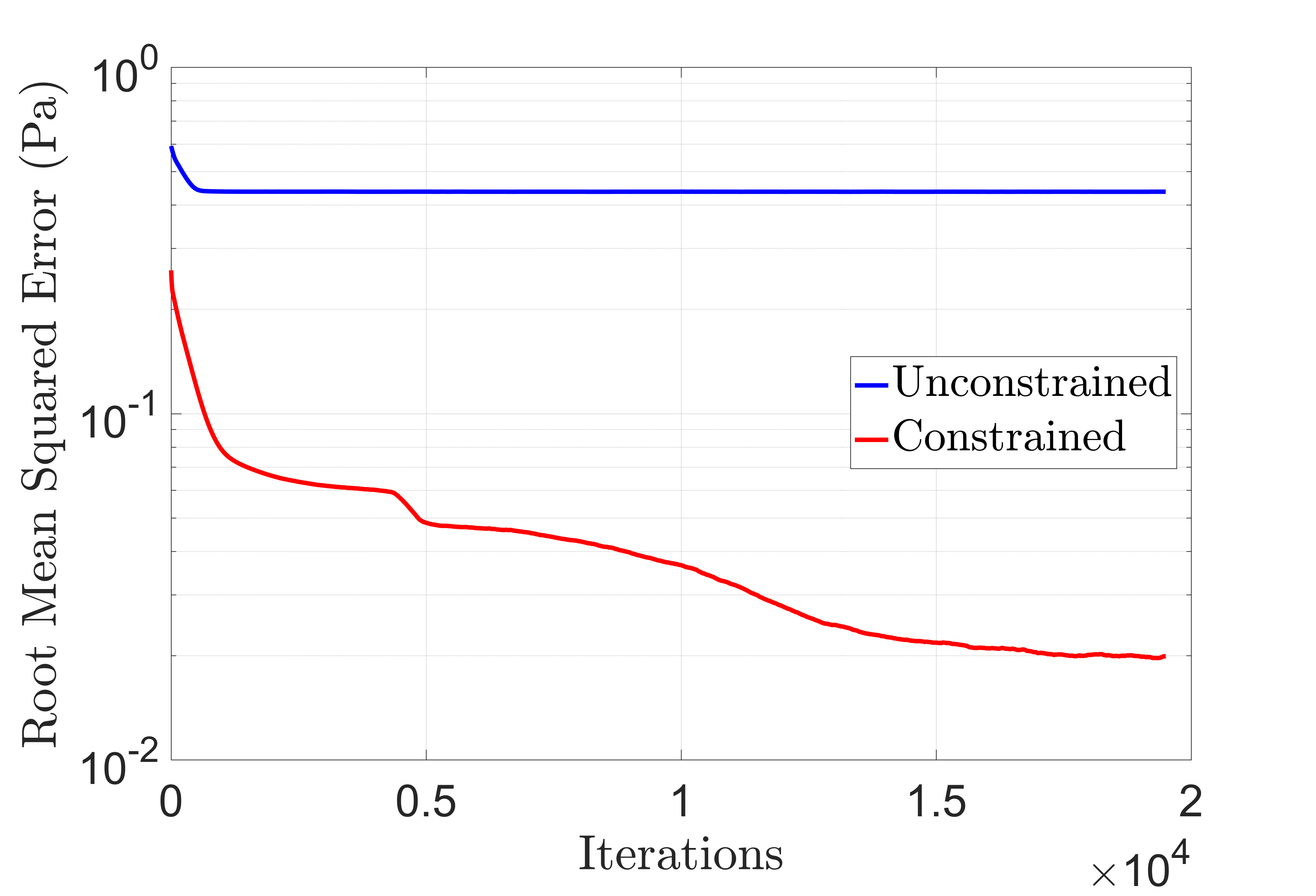}
		\caption{Problem with variable geometry.}
		\label{fig::acceleration_1b}
\end{subfigure} \\
\begin{subfigure}{0.49\textwidth}
		\centering
		\includegraphics[width=\linewidth, trim =0 0 0 0,clip]{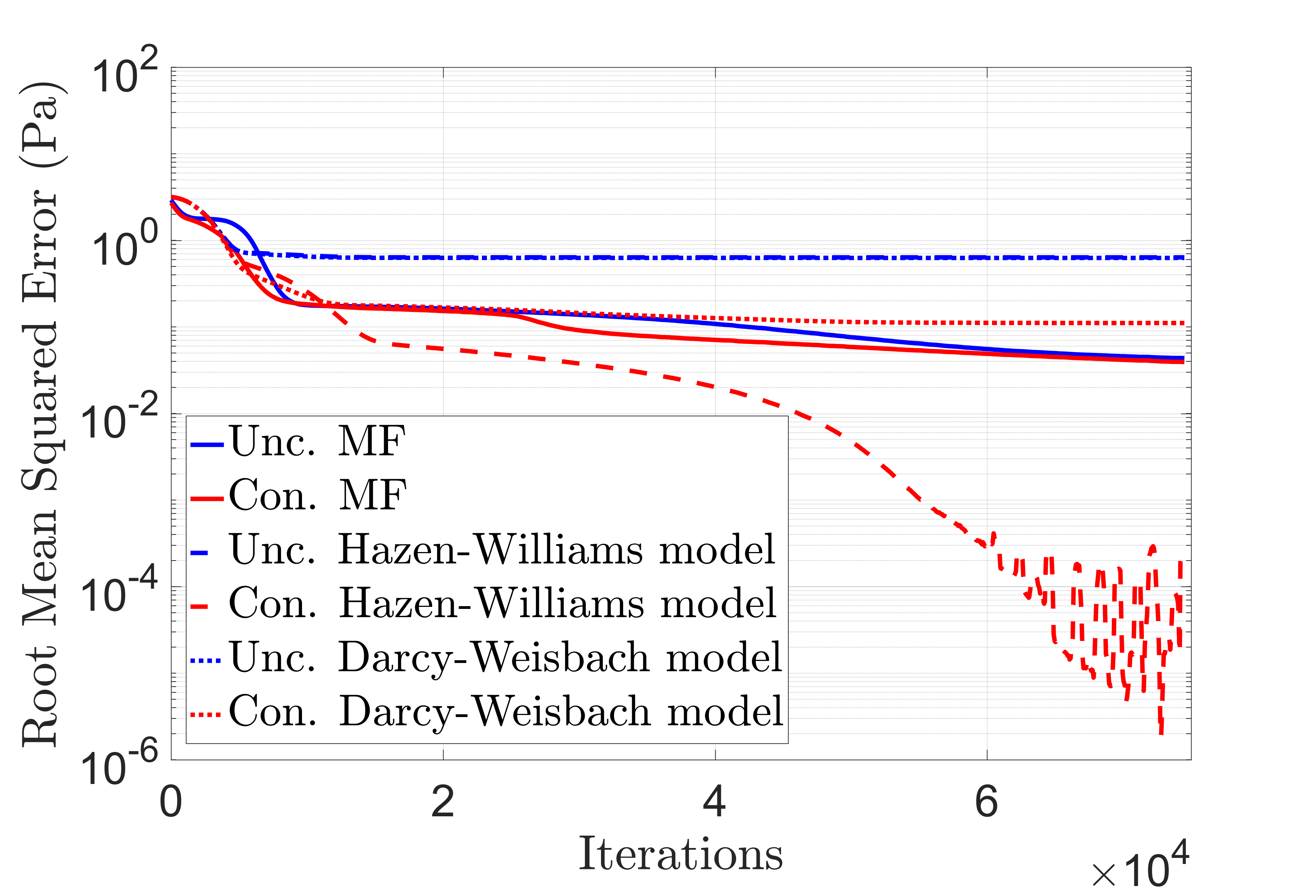}
		\caption{Problem with model inclusion.}
		\label{fig::acceleration_1c}
\end{subfigure}
\begin{subfigure}{0.49\textwidth}
		\centering
		\includegraphics[width=\linewidth, trim =0 0 0 0,clip]{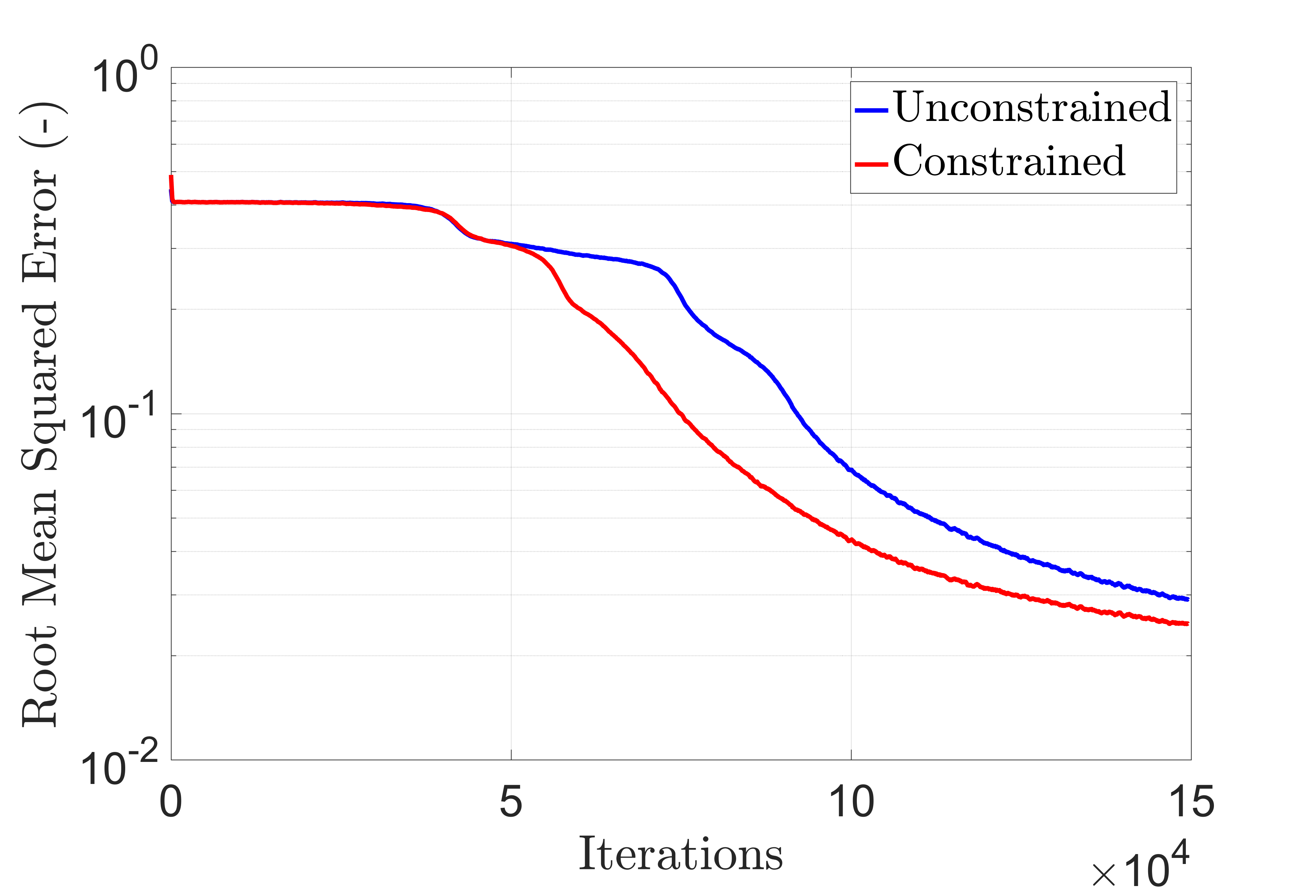}
		\caption{Characterization problem.}
		\label{fig::acceleration_1d}
\end{subfigure} \\
\caption{\textbf{Convergence comparison of constrained and unconstrained neural networks.} Convergence curves are smoothed with a moving average filter (window $W=500$) for easier comparison. For all cases, the constrained neural network shows a convergence acceleration since the search space is smaller. In general, the accuracy is not necessarily improved but the network convergence is accelerated. Adding constraints (see (c)) always speeds-up the convergence, regardless the accuracy is improved or not.}
\label{fig::acceleration_1}
\end{figure}

Fig \ref{fig::acceleration_1} shows the effect of the constraints in the network convergence for four of the presented problems: i) fundamental prediction problem (Fig. \ref{fig::acceleration_1a}), ii) problem with the geometry inclusion (Fig. \ref{fig::acceleration_1b}), iii) model inclusion (Fig. \ref{fig::acceleration_1c}) and iv) characterization problem (Fig. \ref{fig::acceleration_1d}). They show the training process for both NNs (constrained and unconstrained) in terms of the Root Mean Squared Error (RMSE).

\subsubsection{Data need decrease}
 
\begin{figure}
\centering
\includegraphics[clip=true,trim=0pt 0pt 0pt 0pt,width=0.6\textwidth]{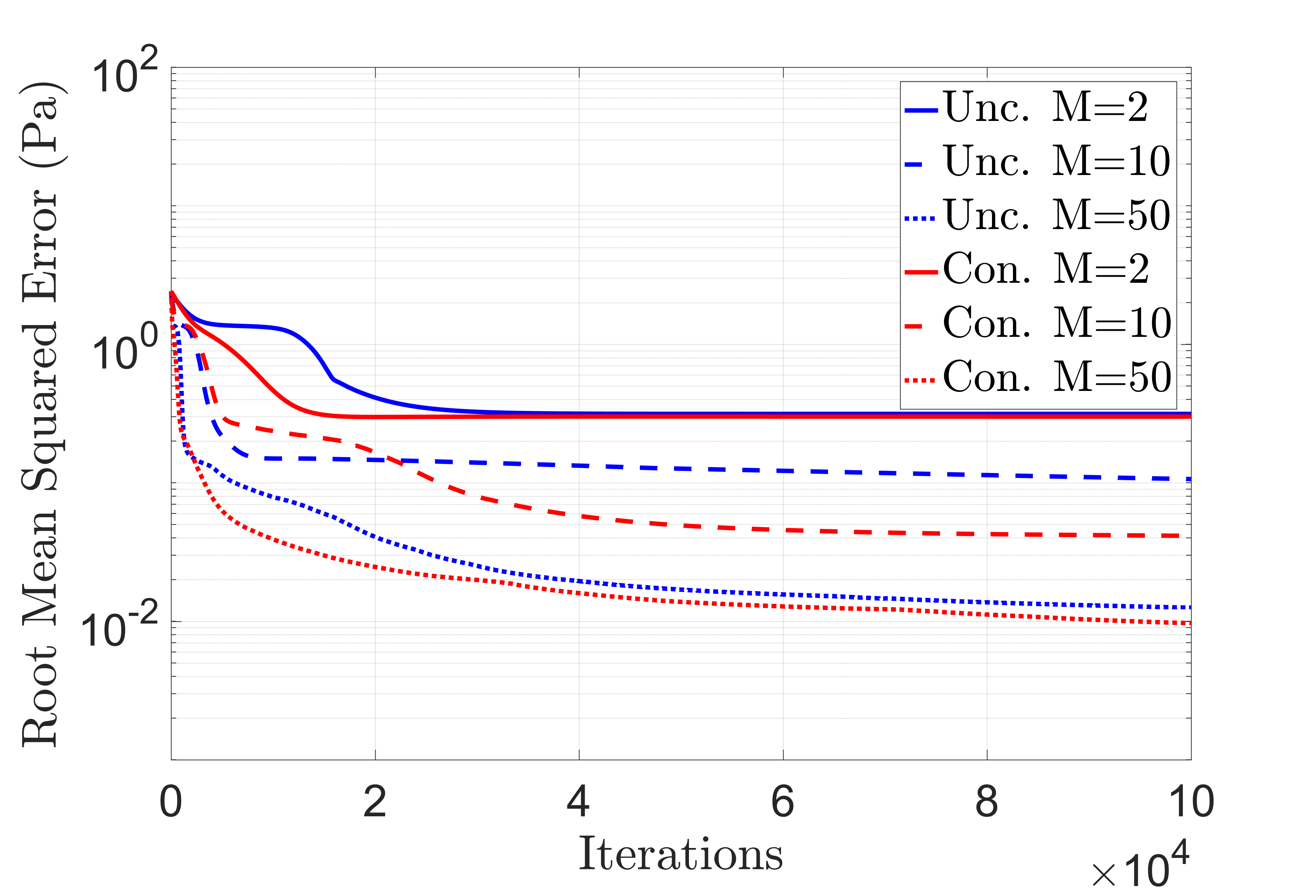}
\caption{\textbf{Learning performance for the two networks and different data-set sizes.} Convergence curves are smoothed with a moving average filter (window $W=500$) for easier comparison. For small data-sets, PGNNIV has an impact not only in the convergence acceleration, but also in the network accuracy.}
\label{fig::data_1}
\end{figure}

\begin{figure}
\centering
\begin{subfigure}{.49\textwidth}
  \centering
  \includegraphics[clip=true,trim=0pt 0pt 0pt 0pt,width=\textwidth]{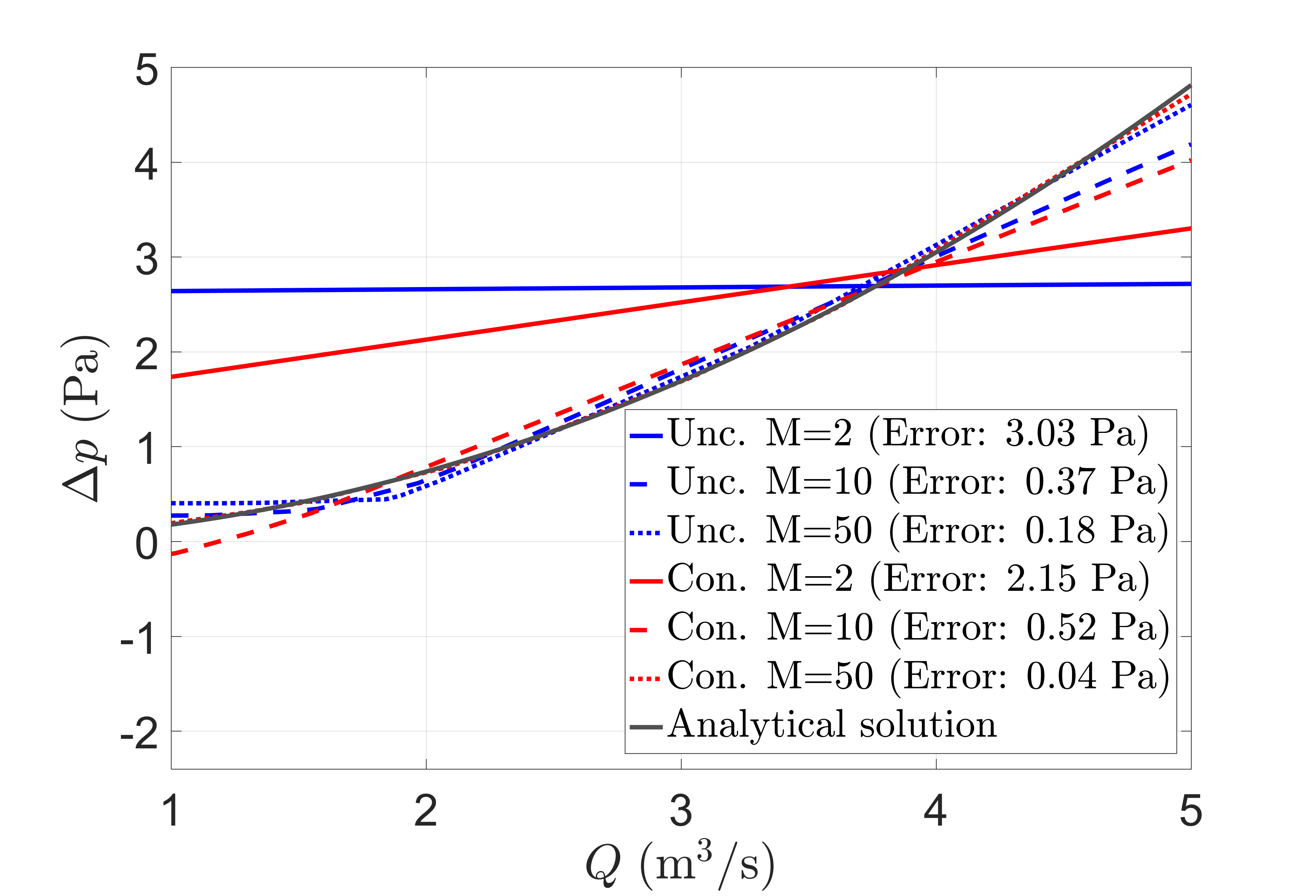}
  \caption{$N=10000$.}
  \label{fig::data_2a}
\end{subfigure}
\begin{subfigure}{.49\textwidth}
  \centering
  \includegraphics[clip=true,trim=0pt 0pt 0pt 0pt,width=\textwidth]{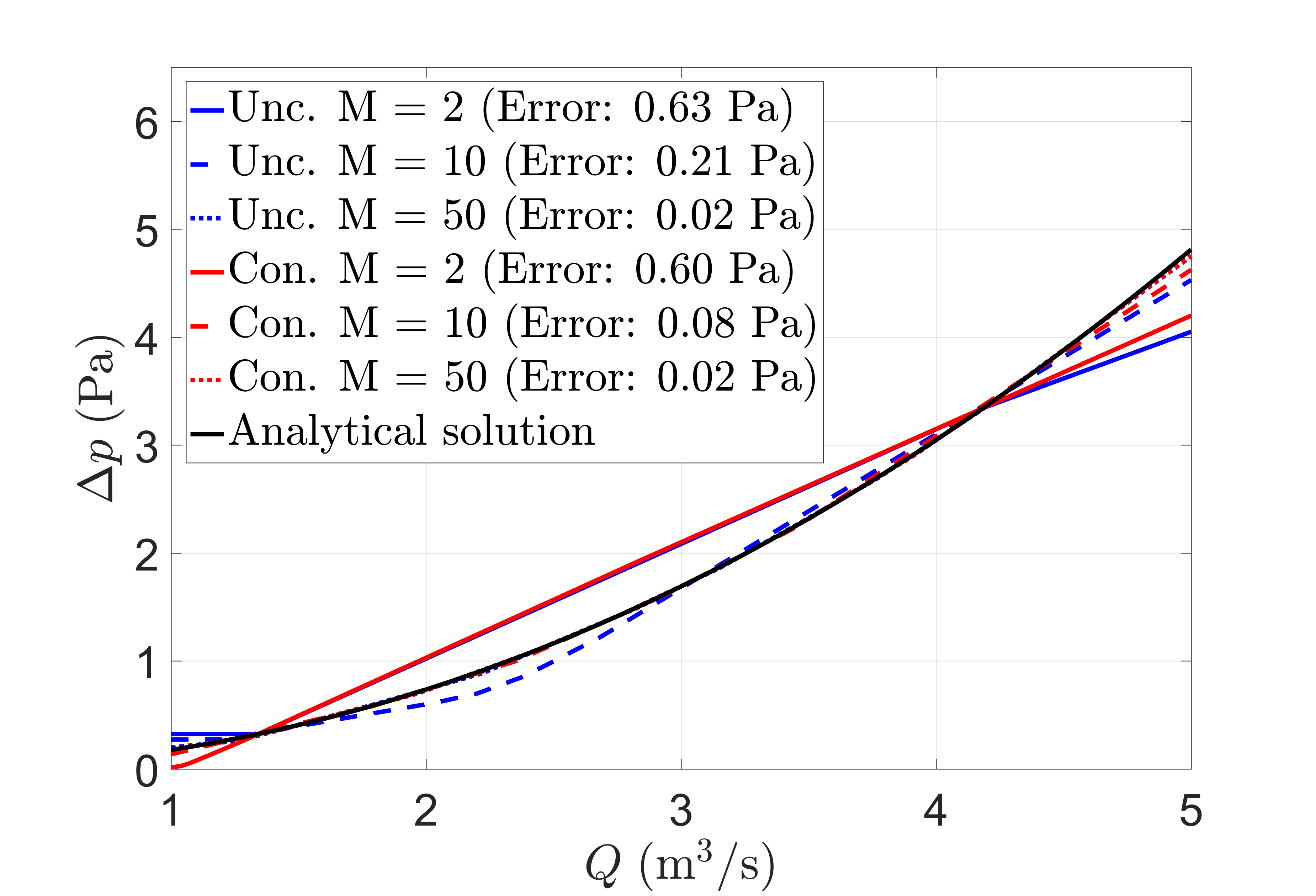}
  \caption{$N=100000$.}
  \label{fig::data_2b}
\end{subfigure}%
\caption{\textbf{Comparison of the network output with the real solution for different data-set sizes and at different convergence states.} For a small data-set ($M=2$), the network only accelerates convergence. When the data-set size is small but large enough ($M=10$), the constraints have an impact in the network accuracy. For large enough data-sets ($M=100$), we recover the big data framework, where the constraint accelerates convergence.}
\label{fig::data_2}
\end{figure}

Another way of seeing the speed-up capability of PGNNIV is to focus on data need. In many engineering problems, especially in those related to material sciences (solid and fluid mechanics, electromagnetism, optics etc.), there is a lack of experimental data due to technical or economic reasons (\emph{Small Data} framework), so, reducing the amount of training data required in the process is essential. The effect of the constraints will be evaluated in terms of the amount of data required, using the pipe flow prediction problem. For this purpose, the learning curve is evaluated for a varying size of the data-set, $M=2,10,50$. The number of iterations was set as $N=3000$ and the batch size is fixed to $n=M$ (that is, the whole data-set is evaluated in each training step). The RMSE convergence curves are depicted in Fig. \ref{fig::data_1}. Another way of analyzing these results is illustrated in Fig. \ref{fig::data_2}, where the prediction of $\Delta p = f(q)$ is shown for different data-set sizes and in different convergence steps.

From Fig. \ref{fig::data_1}, it is possible to make some particular conclusions:

\begin{itemize}
\item If both networks have the same final accuracy, given a fixed number of iterations, the physically constrained network performance is better (lower error) than the unconstrained one.
\item The impact of the data-set size is more gradual in the constrained network. Indeed, the curves associated with unconstrained networks are more step-like (specially for small data-sets), when compared to  the constrained ones. This is very important in order to detect network convergence stabilization.
\end{itemize}

In brief, constrained neural networks accelerate the learning process in such a way that they are able to discover important new features with less data (small data problems), what is extremely important in a practical engineering context.

\subsubsection{Filtering capacity improvement} 

\begin{figure}
\centering
\includegraphics[clip=true,trim=0pt 0pt 0pt 0pt,width=0.6\textwidth]{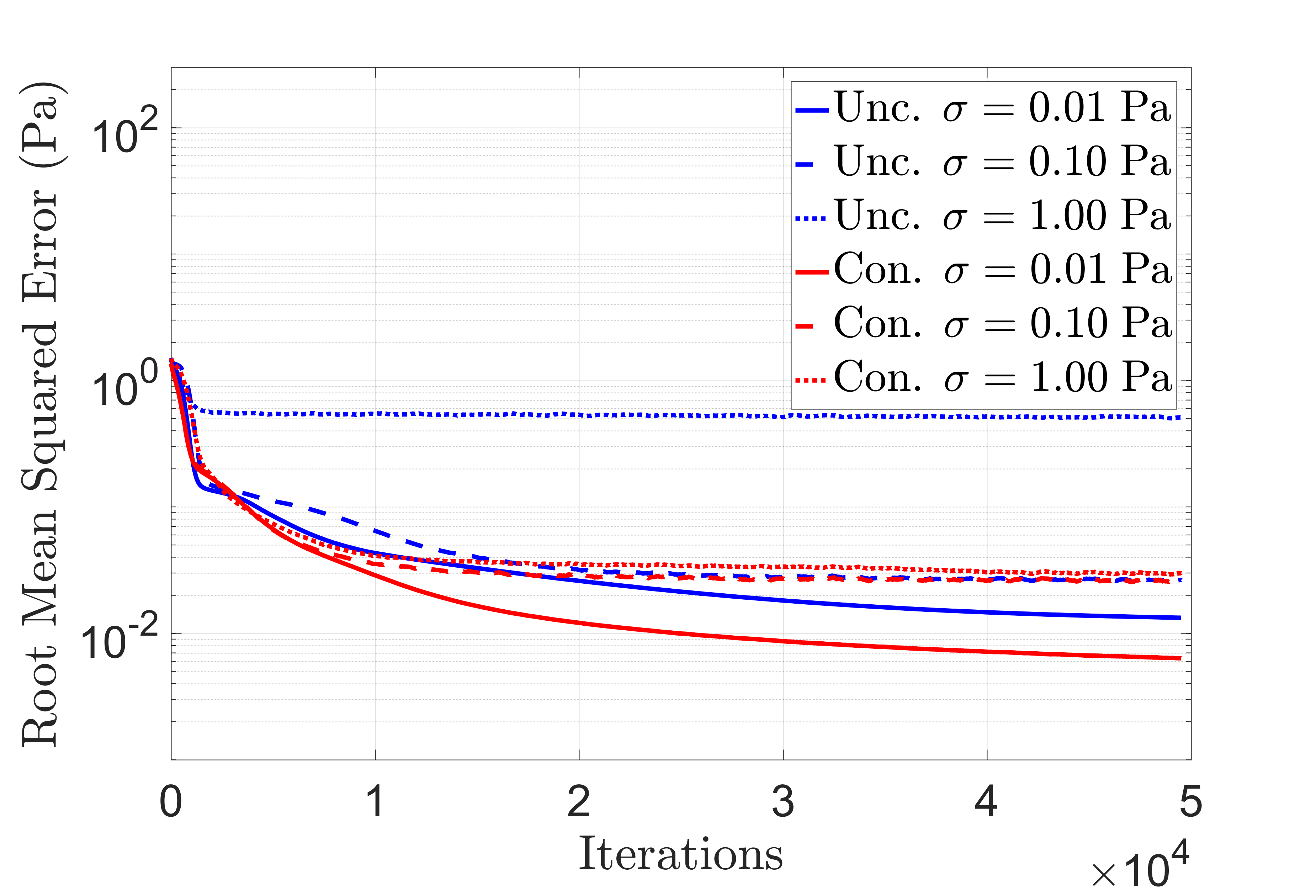}
\caption{\textbf{Learning performance for the two networks and different noise levels.} Convergence curves are smoothed with a moving average filter (window $W=500$) for easier comparison. The noise has much lower impact in the network convergence, both in speed-up and accuracy.}
\label{fig::noise_1}
\end{figure}

\begin{figure}
\centering
\begin{subfigure}{.49\textwidth}
  \centering
  \includegraphics[clip=true,trim=0pt 0pt 0pt 0pt,width=\textwidth]{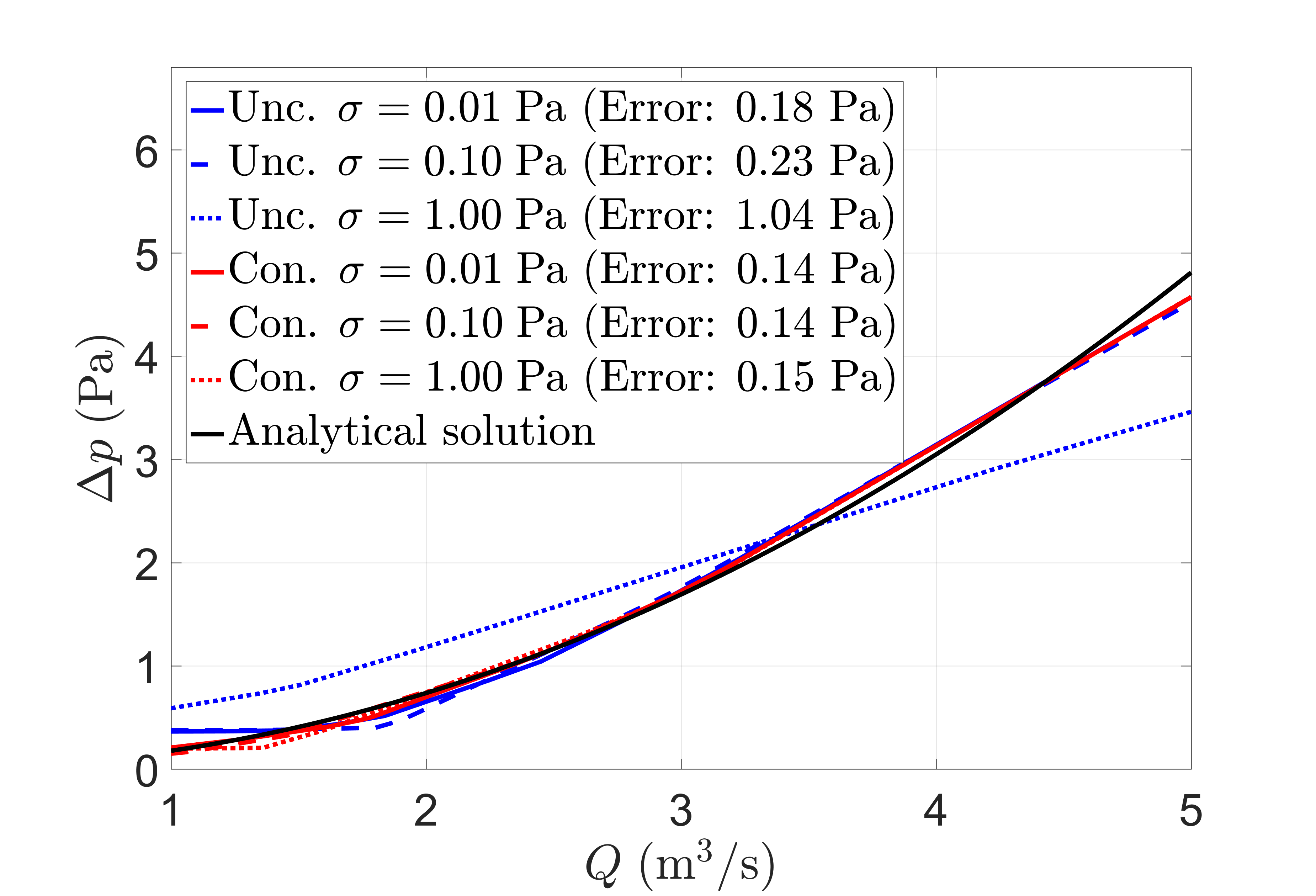}
  \caption{$N=5000$.}
  \label{fig::noise_2a}
\end{subfigure}
\begin{subfigure}{.49\textwidth}
  \centering
  \includegraphics[clip=true,trim=0pt 0pt 0pt 0pt,width=\textwidth]{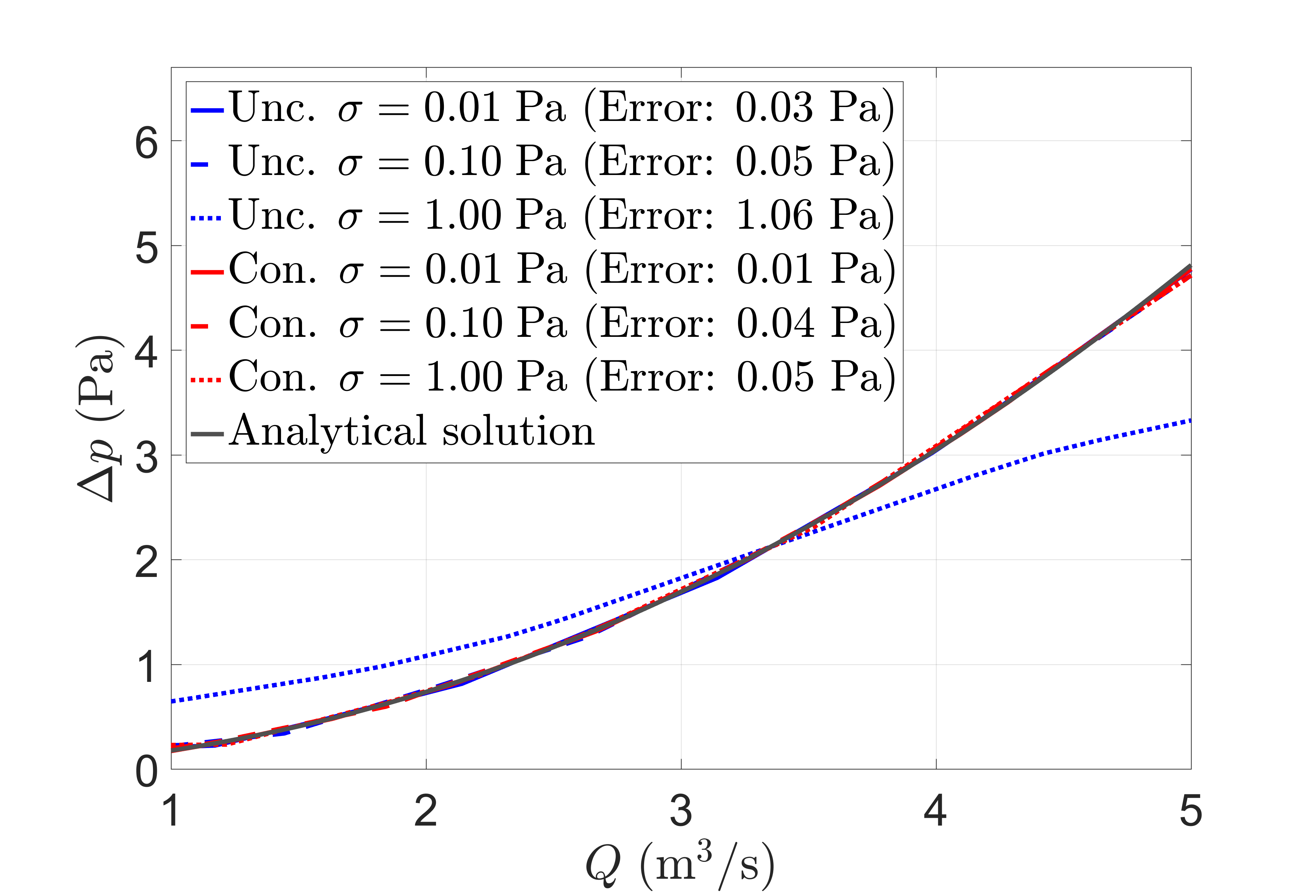}
  \caption{$N=50000$.}
  \label{fig::noise_2b}
\end{subfigure}%
\caption{\textbf{Comparison of the network output with the real solution for different noise levels and at different convergence states.} The unconstrained network has a pathological convergence when $\sigma=1.00 \, \mathrm{Pa}$. The constrained network reaches the same solution point for $\sigma = 1.00 \, \mathrm{Pa}$ than the unconstrained one with a noise lower in a factor of 10. Besides, at $N=5000$, all constrained networks provide similar results.}
\label{fig::noise_2}
\end{figure}

Here, the noise filtering capacity of the constrained network is explored. In the prediction pipe flow problem, the data-set has been considered noise free. That is, a data-set $(\bar{q}^i,\bar{(\Delta p)}^i)$ was generated directly from the equation (\ref{eq::model_pQ}). Here, we compare the performance of both neural networks when working with noisy data, a more realistic situation in experimental problems and data acquired from sensors. To show the noise impact in the learning process, let us assume a data-sets with an added Gaussian noise, i.e. $\mathrm{x} = \bar{\mathrm{x}} + Z$ with $Z \sim \mathcal{N}(0,\sigma)$ for $\mathrm{x} = q,\Delta p$ and $\sigma  =0.01,0.10,1.00 \, \mathrm{Pa}$. The RMSE convergence curves are illustrated in Fig. \ref{fig::noise_1}. Fig. \ref{fig::noise_2} represents the network accuracy in the $(q,\Delta p)$ plane for different convergence stages.

\begin{figure}
\centering
\includegraphics[clip=true,trim=0pt 0pt 0pt 0pt,width=0.6\textwidth]{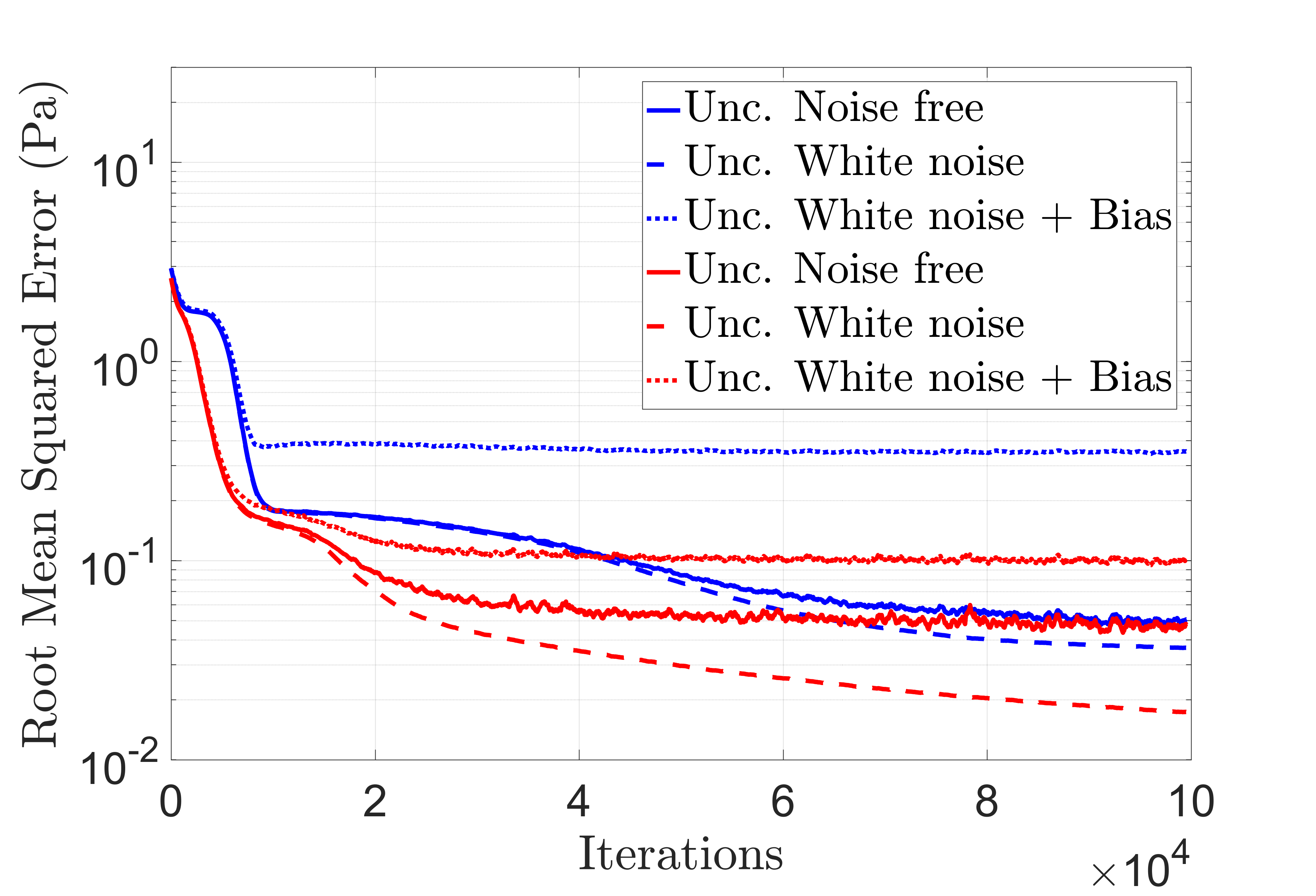}
\caption{\textbf{Effect on the learning performance of white noise and bias.} Convergence curves are smoothed with a moving average filter (window $W=500$) for easier comparison. Bias is partially corrected and noise filtered in constrained networks. Some level of noise may improve network accuracy, as reported in the literature \cite{holmstrom1992using,grandvalet1997noise,skurichina2000k}.}
\label{fig::noise_3}
\end{figure}

\begin{figure}
\centering
\begin{subfigure}{.49\textwidth}
  \centering
  \includegraphics[clip=true,trim=0pt 0pt 0pt 0pt,width=\textwidth]{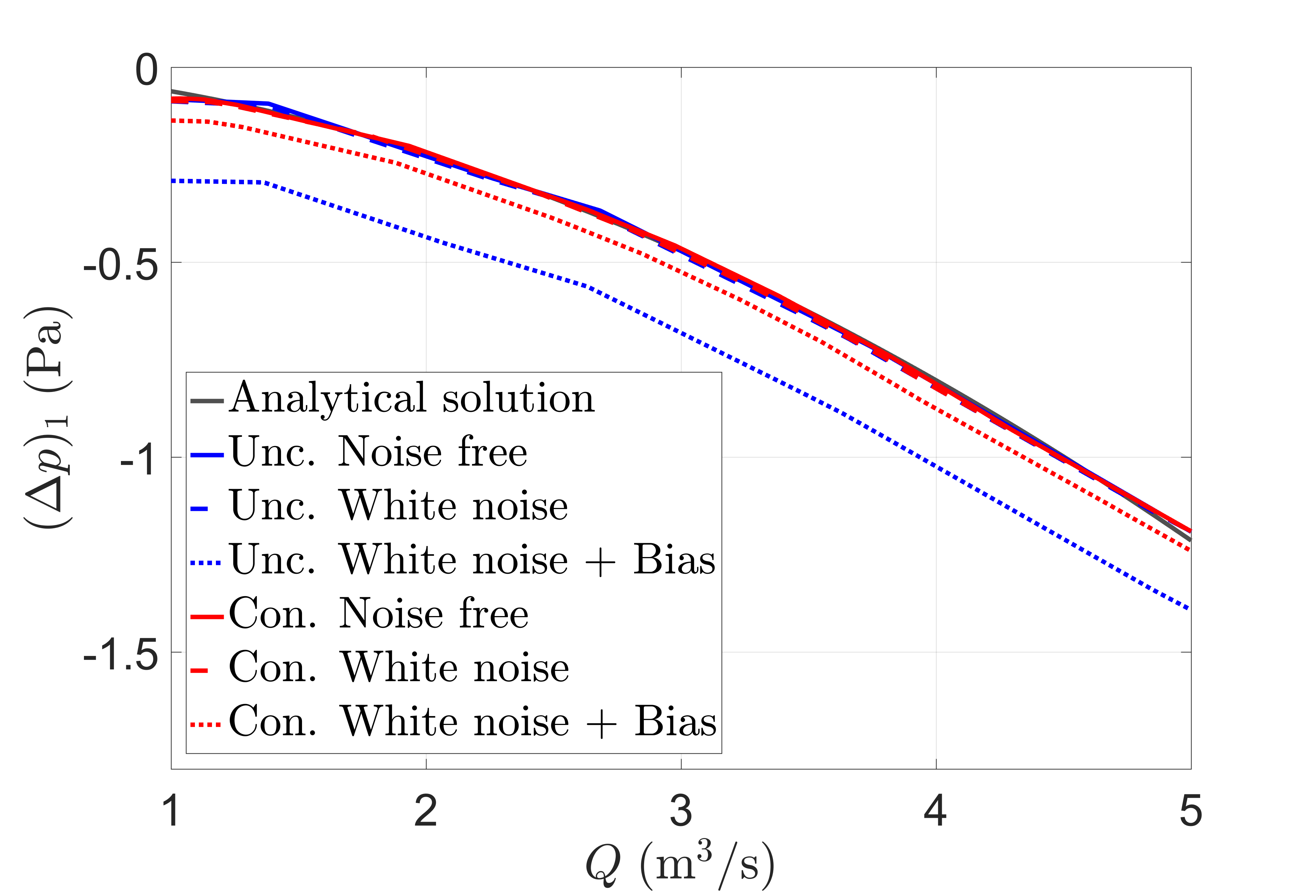}
  \caption{$(\Delta p)_1$.}
  \label{fig::noise_4a}
\end{subfigure}
\begin{subfigure}{.49\textwidth}
  \centering
  \includegraphics[clip=true,trim=0pt 0pt 0pt 0pt,width=\textwidth]{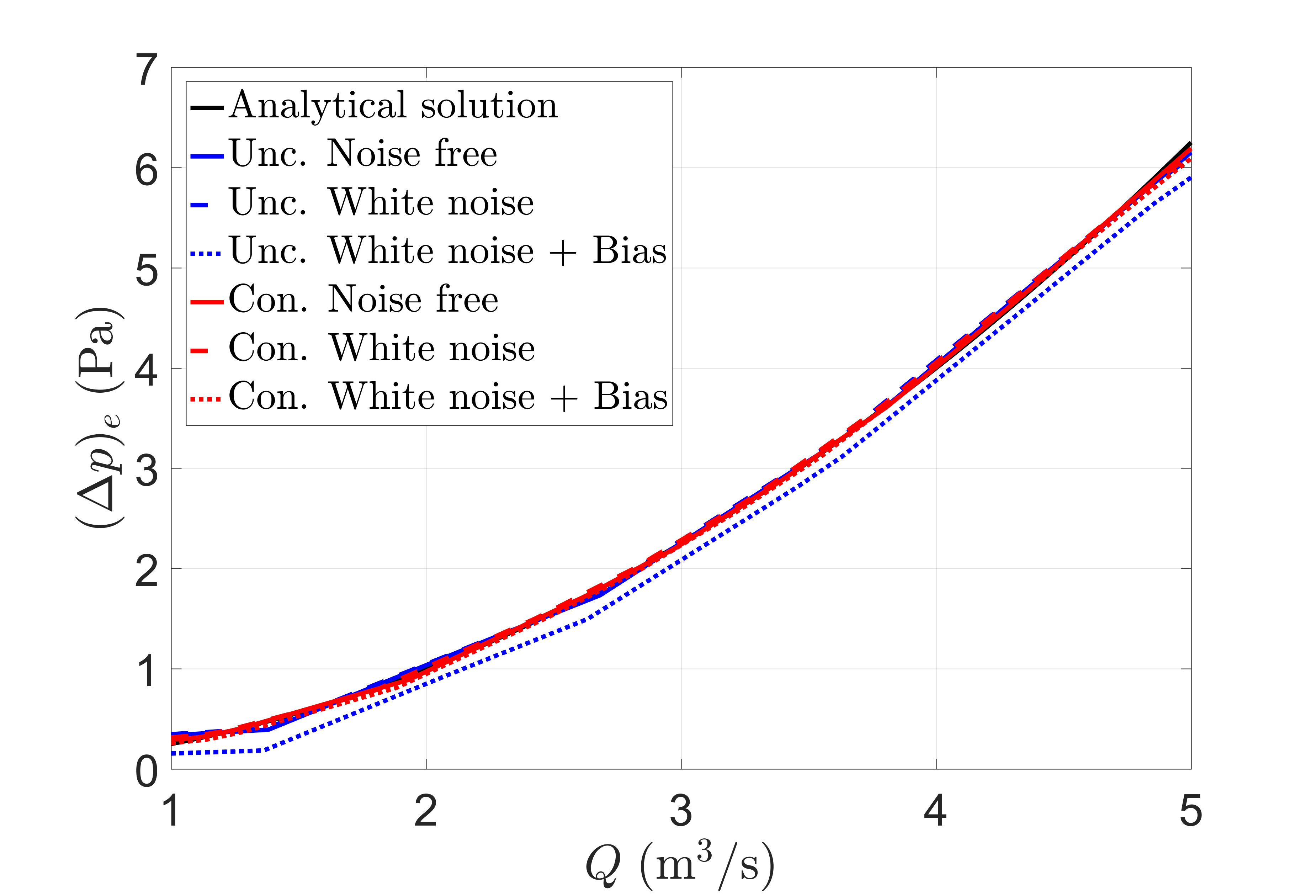}
  \caption{$(\Delta p)_e$.}
  \label{fig::noise_4b}
\end{subfigure} \\
\begin{subfigure}{.49\textwidth}
  \centering
  \includegraphics[clip=true,trim=0pt 0pt 0pt 0pt,width=\textwidth]{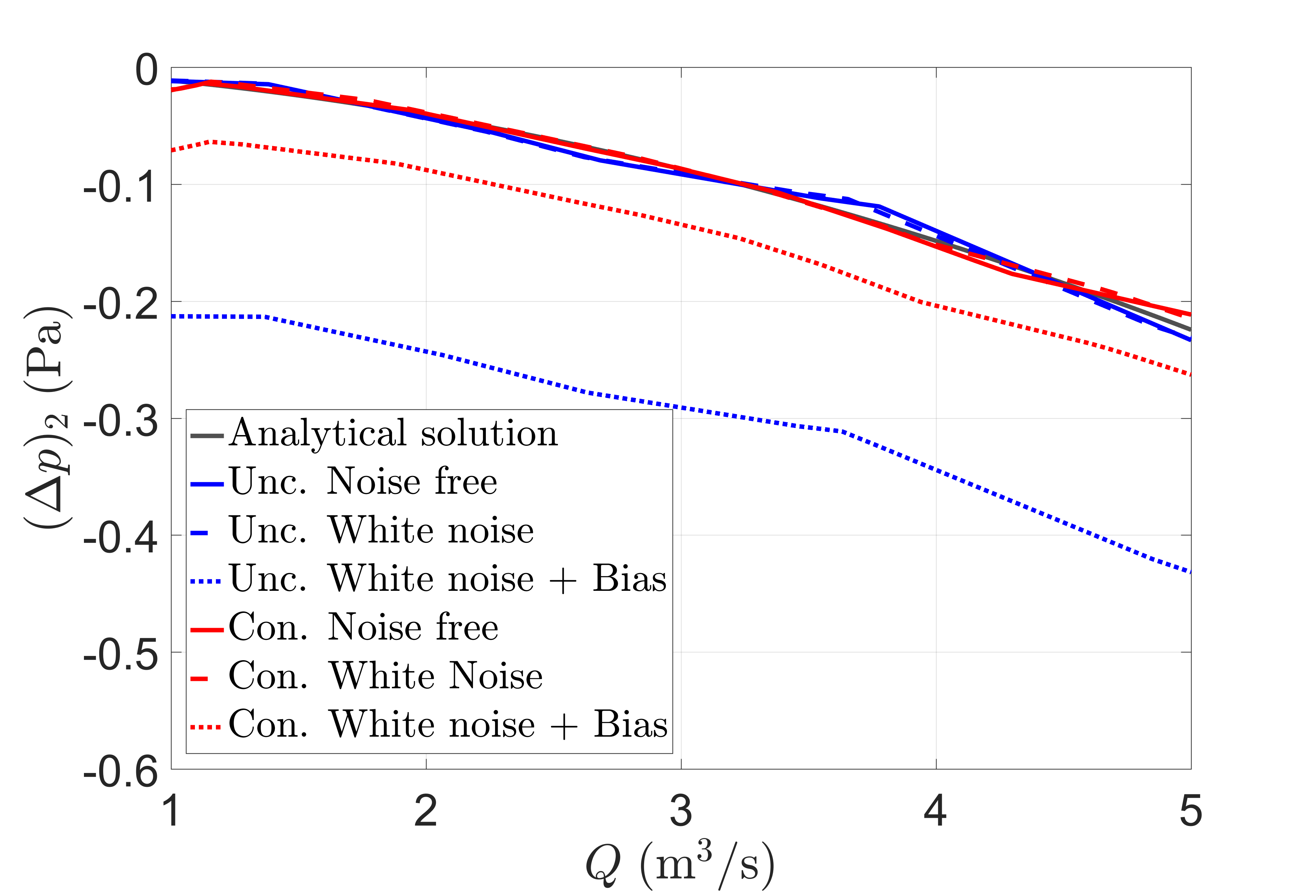}
  \caption{$(\Delta p)_2$.}
  \label{fig::noise_4c}
\end{subfigure}%
\caption{\textbf{Effect of white noise and bias in the network prediction.} The effect of adding the constraints in the output layer is a bias reduction in the pressure drop estimation.}
\label{fig::noise_4}
\end{figure}

The effect is even stronger if the constraint acts on the output layer. For instance, let us consider the network with output $((\Delta p)_1, (\Delta p)_e, (\Delta p)_2)$ and let us add the constraint $(\Delta p)_1 + (\Delta p)_e + (\Delta p)_1 = \Delta p$, where $\Delta p$ is another measured variable (the total pressure drop). In addition to a noise of $\sigma = 0.1 \, \mathrm{Pa}$, we consider also the possibility of adding a systematic bias of $-0.2 \, \mathrm{Pa}$ to all the measured variables. Fig \ref{fig::noise_3} shows the convergence curves and \ref{fig::noise_4} the network accuracy in the $(q,\Delta p)$ plane.

As main results:

\begin{itemize}
\item The performance and learning capacity of both networks decrease with noise. This is in agreement with other works \cite{jim1996analysis,kalapanidas2003machine,rolnick2017deep}.
\item The impact of the noise is lower in the network convergence, as the RMSE curves are closer for the PGNNIV networks.
\item The noise has an impact not only in the network convergence rate, but also in the network accuracy, as the curves associated with constrained networks are strictly under the curves associated with the unconstrained ones for $\sigma = 0.01 \, \mathrm{Pa}$ and $\sigma = 1.00 \, \mathrm{Pa}$. In other words, the physical constraints are able to partially filter the noise.
\item Other systematic errors as bias may be corrected partially by the addition of the constraints to the network. That is PGNNIV presents bias correction capability.
\end{itemize}

\subsubsection{Extrapolation capability} 

Let us consider the problem with the constitutive model for head loss estimations. We evaluate the network performance in predicting values of the pressure drops out of the learning data-set, that is, for $q \geq 5$.

\begin{figure}
\centering
\begin{subfigure}{.49\textwidth}
  \centering
  \includegraphics[width=\linewidth]{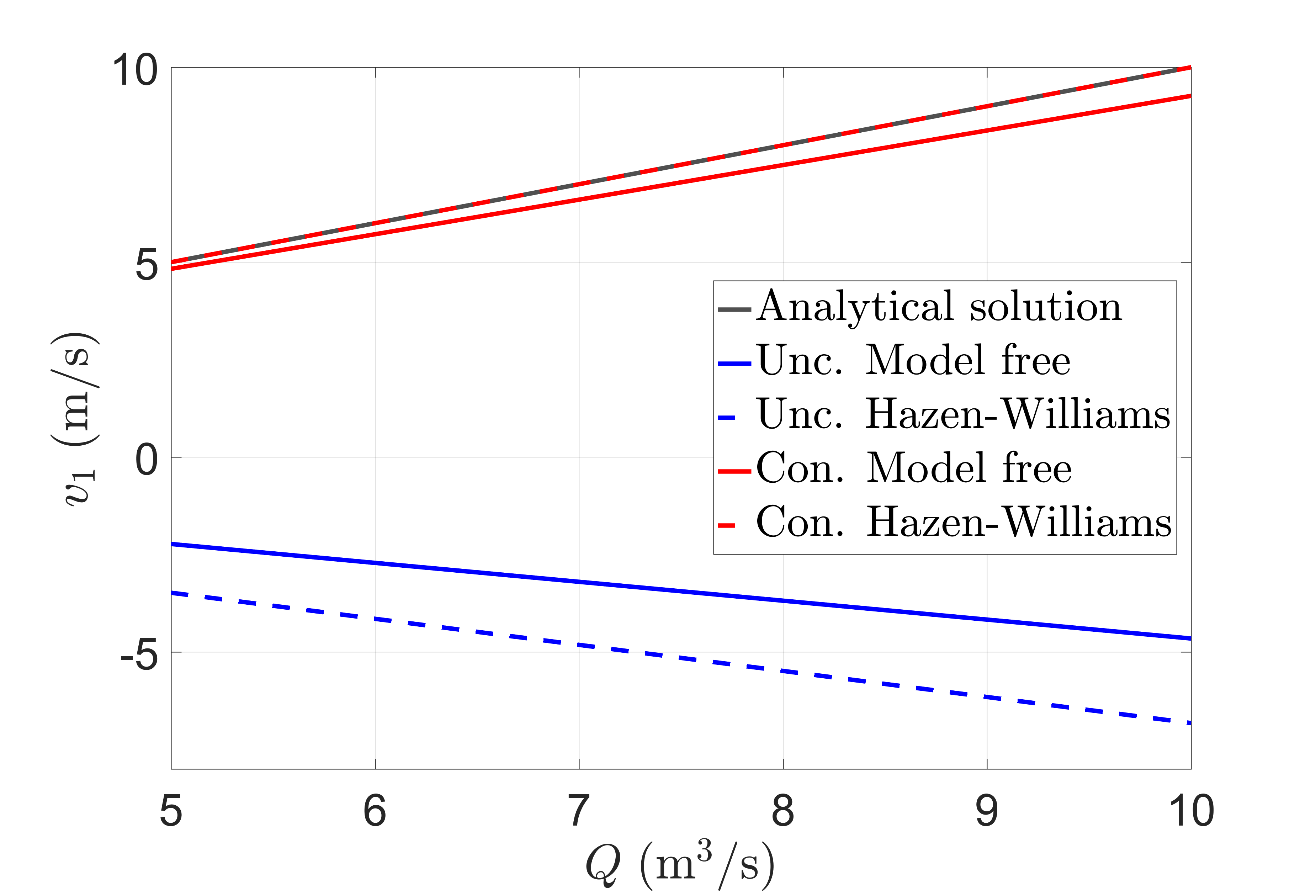}
  \caption{$v_1$.}
  \label{fig::extrapolation_1a}
\end{subfigure}
\begin{subfigure}{.49\textwidth}
  \centering
  \includegraphics[width=\linewidth]{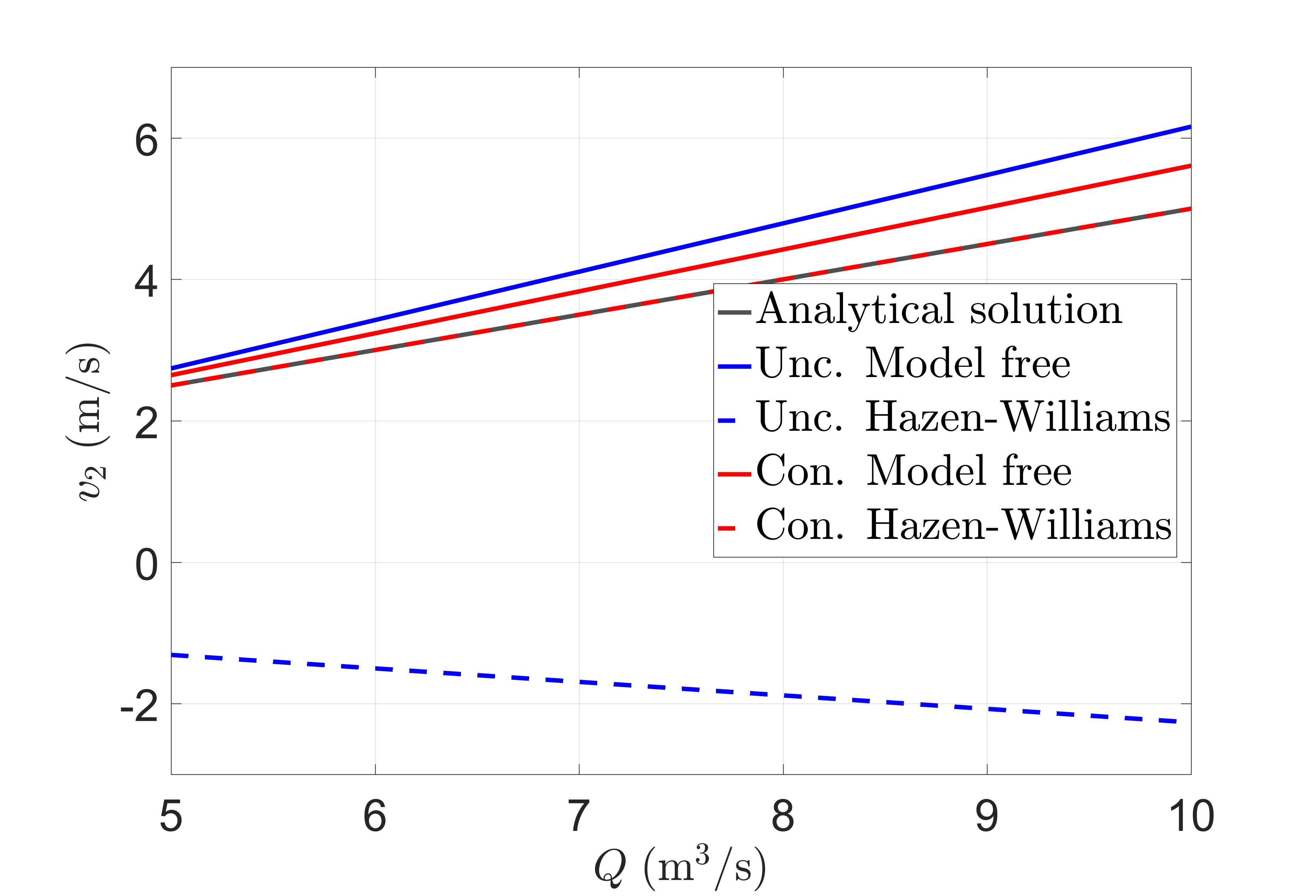}
  \caption{$v_2$.}
  \label{fig::extrapolation_1b}
\end{subfigure} \\
\caption{\textbf{Extrapolation capacity of each neural network in estimating the internal variables.} The constrained network improves always the estimation of the internal variables. For accurate model specifications, the extrapolation capacity is, obviously, total.}
\label{fig::extrapolation_1}
\end{figure}

\begin{figure}
\centering
\begin{subfigure}{.49\textwidth}
  \centering
  \includegraphics[width=\linewidth]{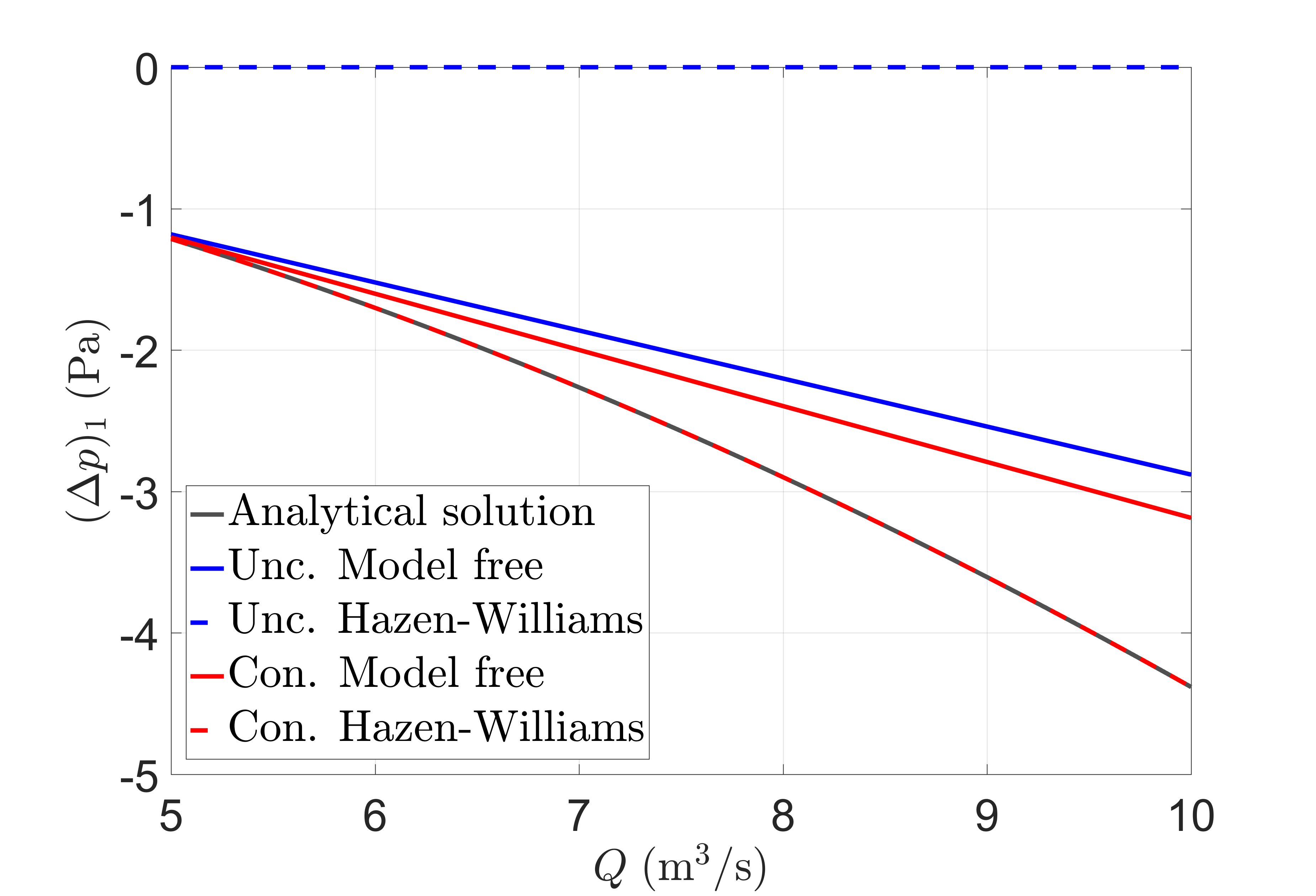}
  \caption{$(\Delta p)_1$.}
  \label{fig::extrapolation_2a}
\end{subfigure}
\begin{subfigure}{.49\textwidth}
  \centering
  \includegraphics[width=\linewidth]{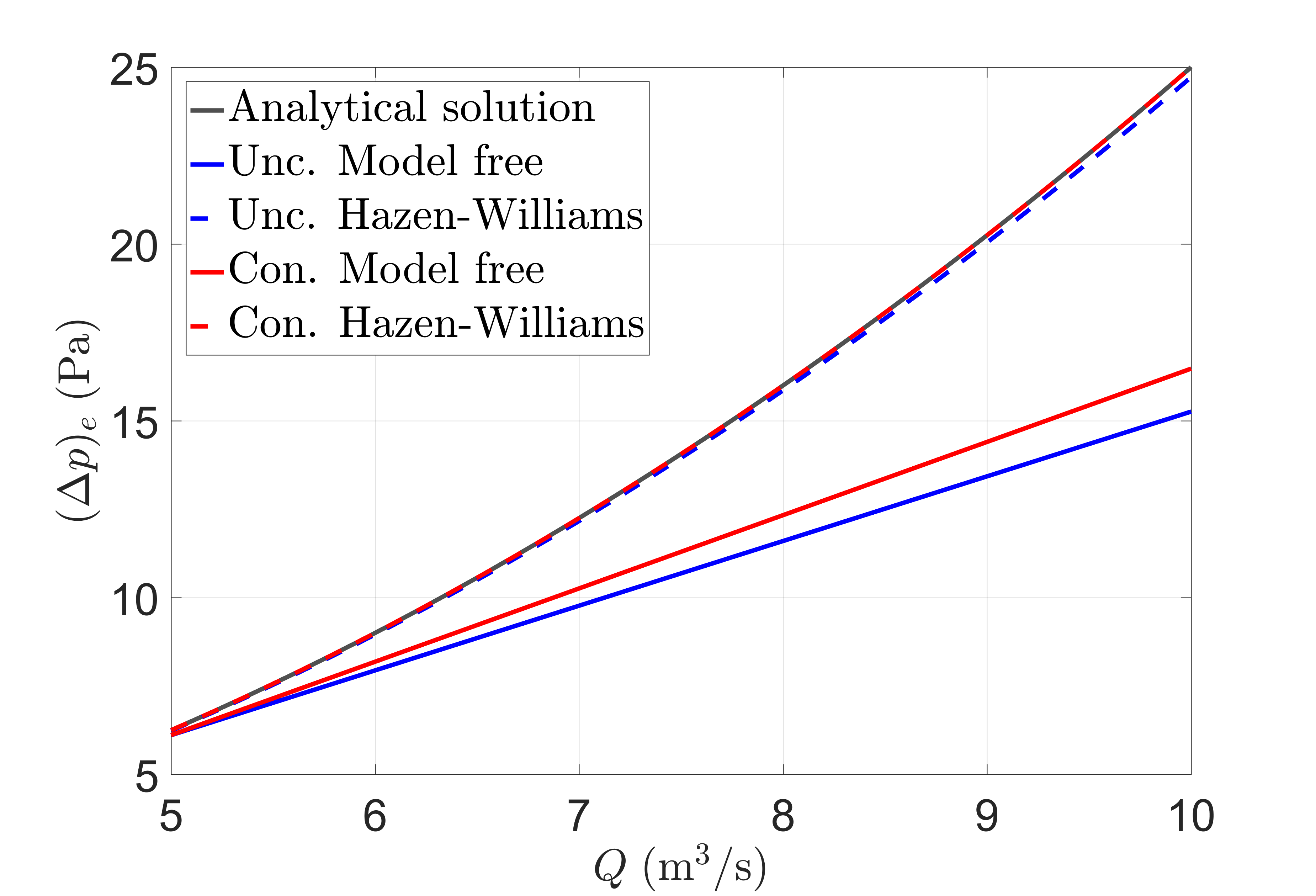}
  \caption{$(\Delta p)_e$.}
  \label{fig::extrapolation_2b}
\end{subfigure} \\
\begin{subfigure}{.49\textwidth}
  \centering
  \includegraphics[width=\linewidth]{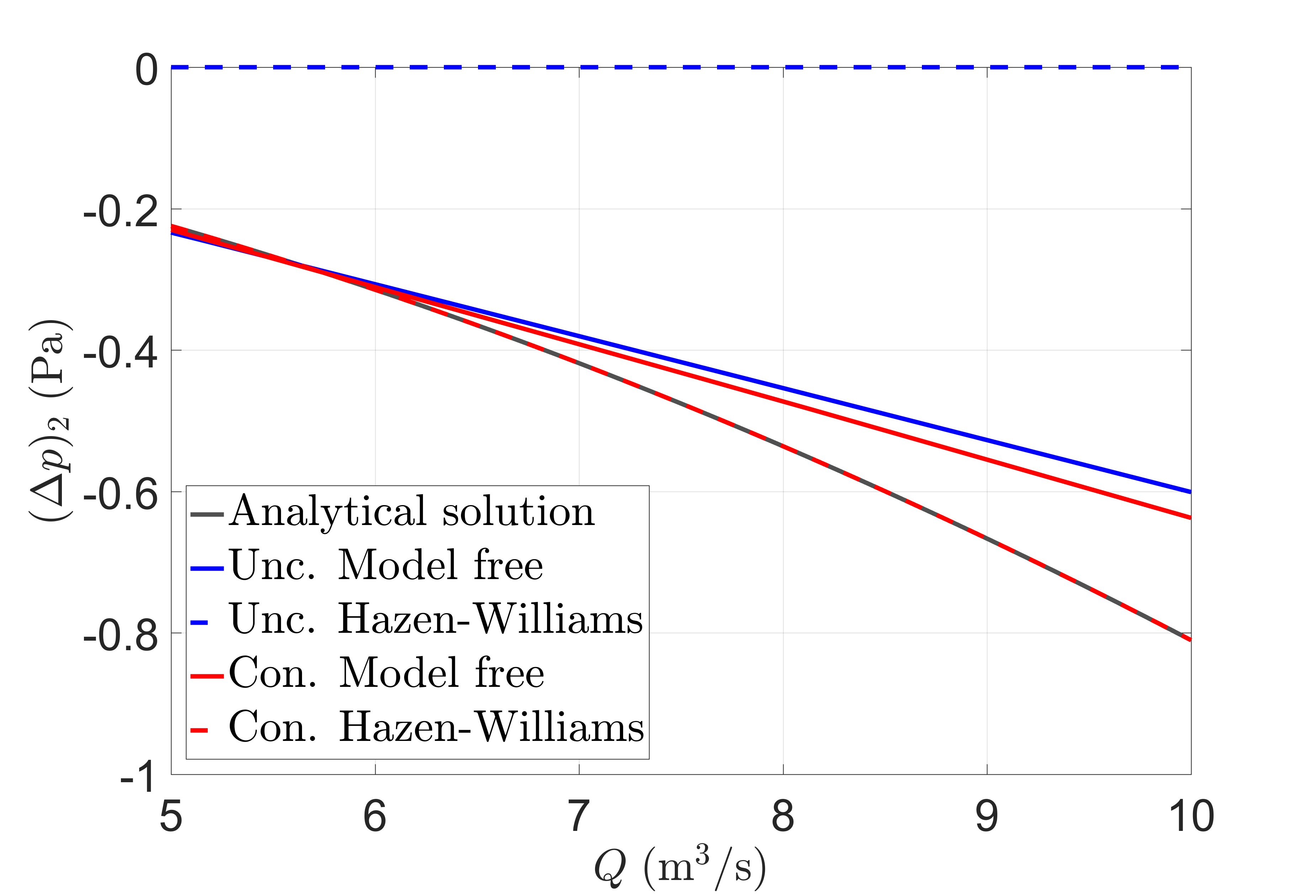}
  \caption{$(\Delta p)_2$.}
  \label{fig::extrapolation_2c}
\end{subfigure} \\
\caption{\textbf{Extrapolation capacity of each neural network in estimating the measurable variables.} The constrained network improves slightly the estimation of the measurable variables. Again, for accurate model specifications, the extrapolation capacity is, obviously, total.}
\label{fig::extrapolation_2}
\end{figure}

\begin{figure}
\centering
\includegraphics[clip=true,trim=0pt 0pt 0pt 0pt,width=0.7\textwidth]{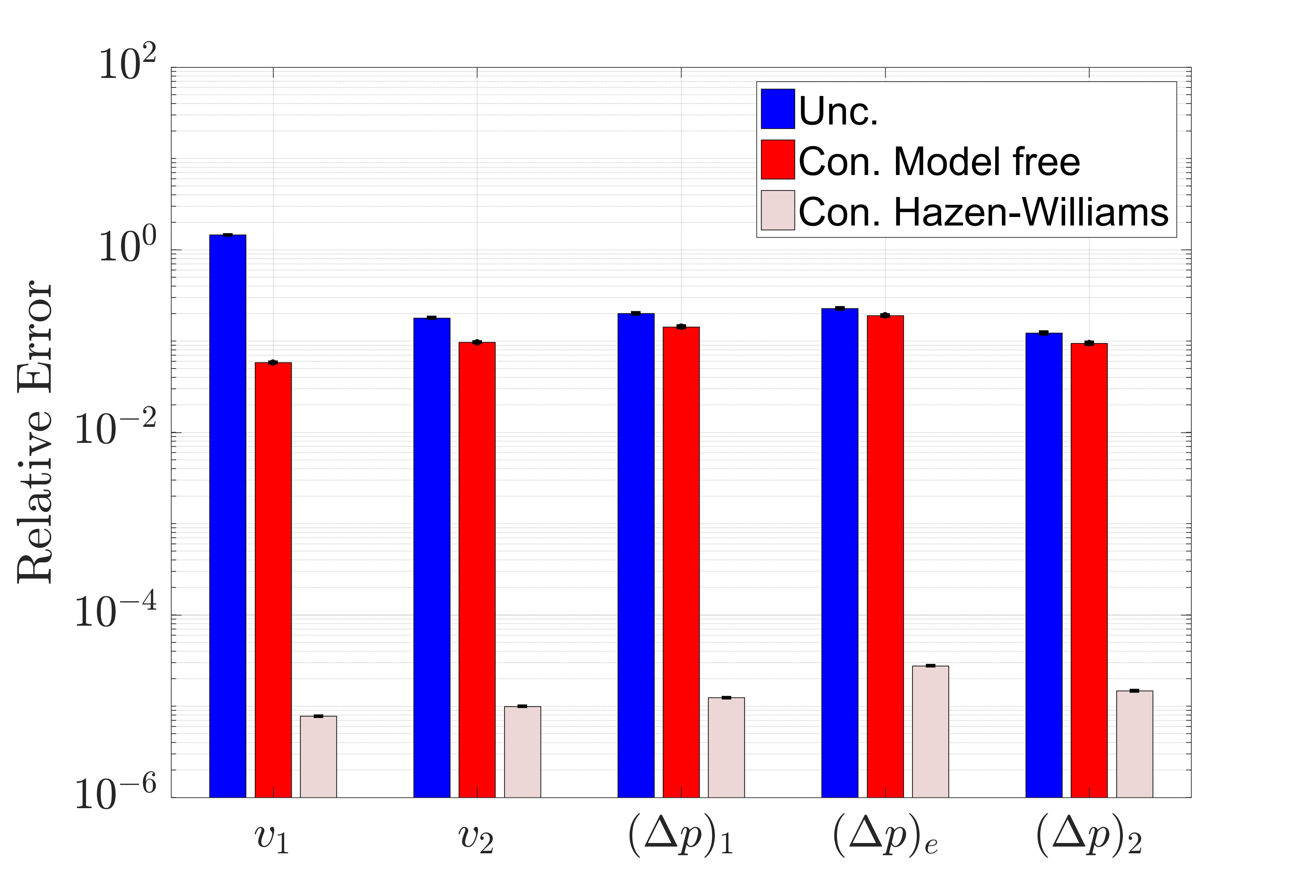}
\caption{\textbf{Evaluation of the extrapolation capacity.} PGNNIV improves always the extrapolation capacity, when compared to standard NN. This improvement is significant even for the measurable variables, although the internal variables are of course much better estimated. For model-based networks, the extrapolation capacity is extremely increased.}
\label{fig::extrapolation_3}
\end{figure}

Fig \ref{fig::extrapolation_1} and Fig. \ref{fig::extrapolation_2} illustrate the extrapolation capacity of the different networks in estimating the internal and measurable variables respectively for different values of $q \in [5;10] \, \mathrm{m^3/s}$. Fig \ref{fig::extrapolation_3} shows the relative errors statistics (mean and standard error bar) of the different variables when extrapolating to the new values of $q$.

We conclude:

\begin{itemize}
\item Even if PGNNIV are designed, among other purposes, for a good estimation of the internal variables, their physically-based nature enables the estimation of the measurable variables out of the learning data-set, which increases its generalization character.
\item Model-based PGNNIV networks are, in a certain sense, similar to standard parameter fitting algorithms so if the model is assumed to be known, the converged network has no error in predicting the values following the model assumptions, even for data out of the training range. If we compare PGNNIV, however, to classical fitting procedures, the former has the advantage of all specific hardware and software relative to ANN technology, as explained in Section \ref{sssecc::comp}.
\end{itemize}

\subsection{Internal state discovering}

Despite the characteristics that make PGNNIV faster, less data-demanding and more robust than common NN, these are not the main causes that justify its use in scientific and engineering problems. Indeed, it is because of their physical explanatory capacity. This can be exploited in two ways: i) accurate prediction of non-measurable internal variables, such as velocities, fluxes or viscous losses and ii) discovery of the hidden physics in a internal state equation such as the Hazen-Williams'. 

The network explanatory capability may be further explored in systems characterization. First, under convergence assumptions, the PGNNIV may be used for predicting the quantitative relation between the different internal variables. But also, the network may provide major features of the structure of the empirical  model, for instance, structural dependence and separability. 

\begin{figure}
\centering
	\begin{subfigure}{.49\textwidth}
		\centering
		\includegraphics[clip=true,trim=10pt 10pt 30pt 10pt, width=\linewidth]{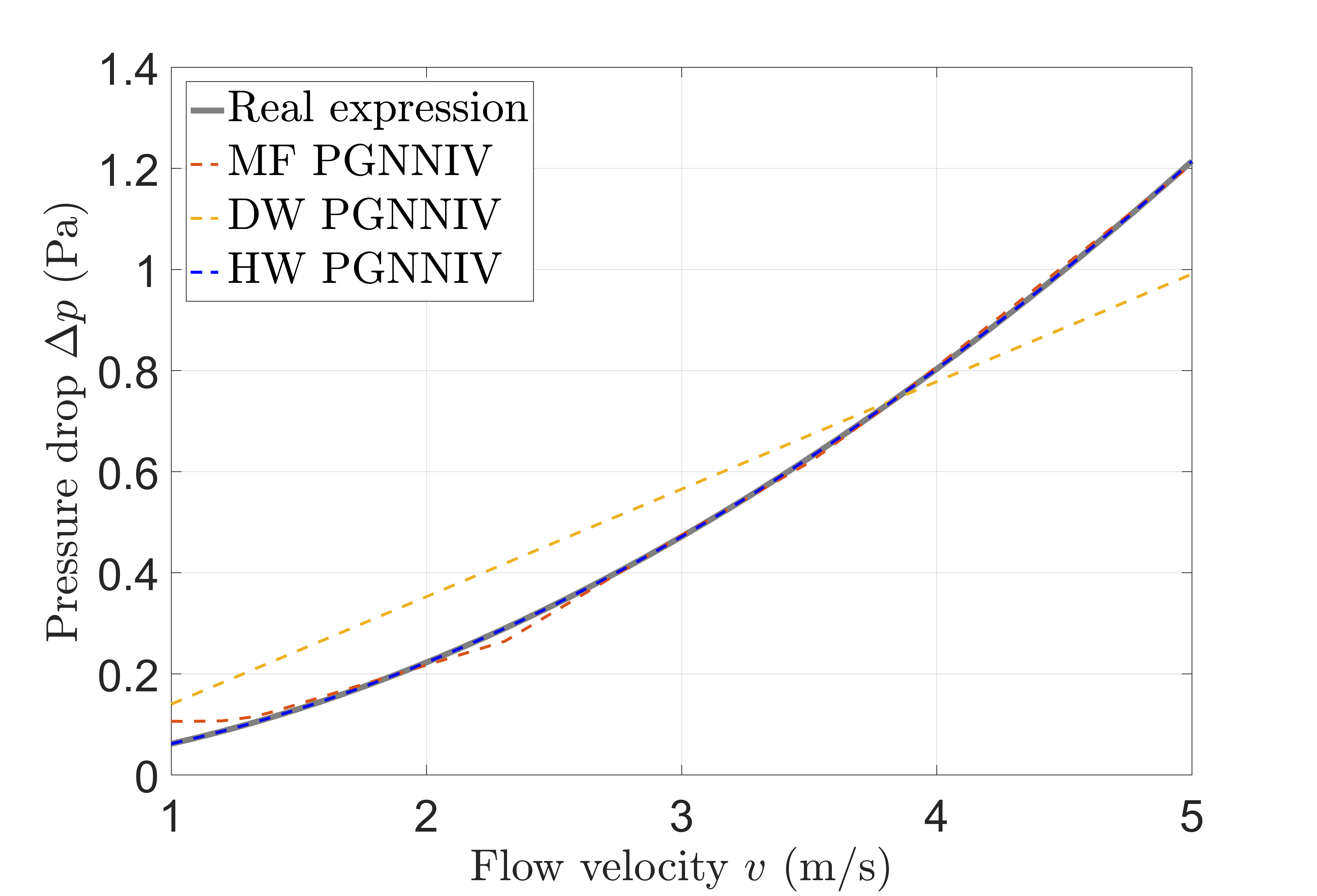}
		\caption{Hazen-Williams model.}
		\label{fig::model_learning_1}
	\end{subfigure}%
	\begin{subfigure}{.49\textwidth}
		\centering
		\includegraphics[clip=true,trim=10pt 10pt 30pt 10pt, width=\linewidth]{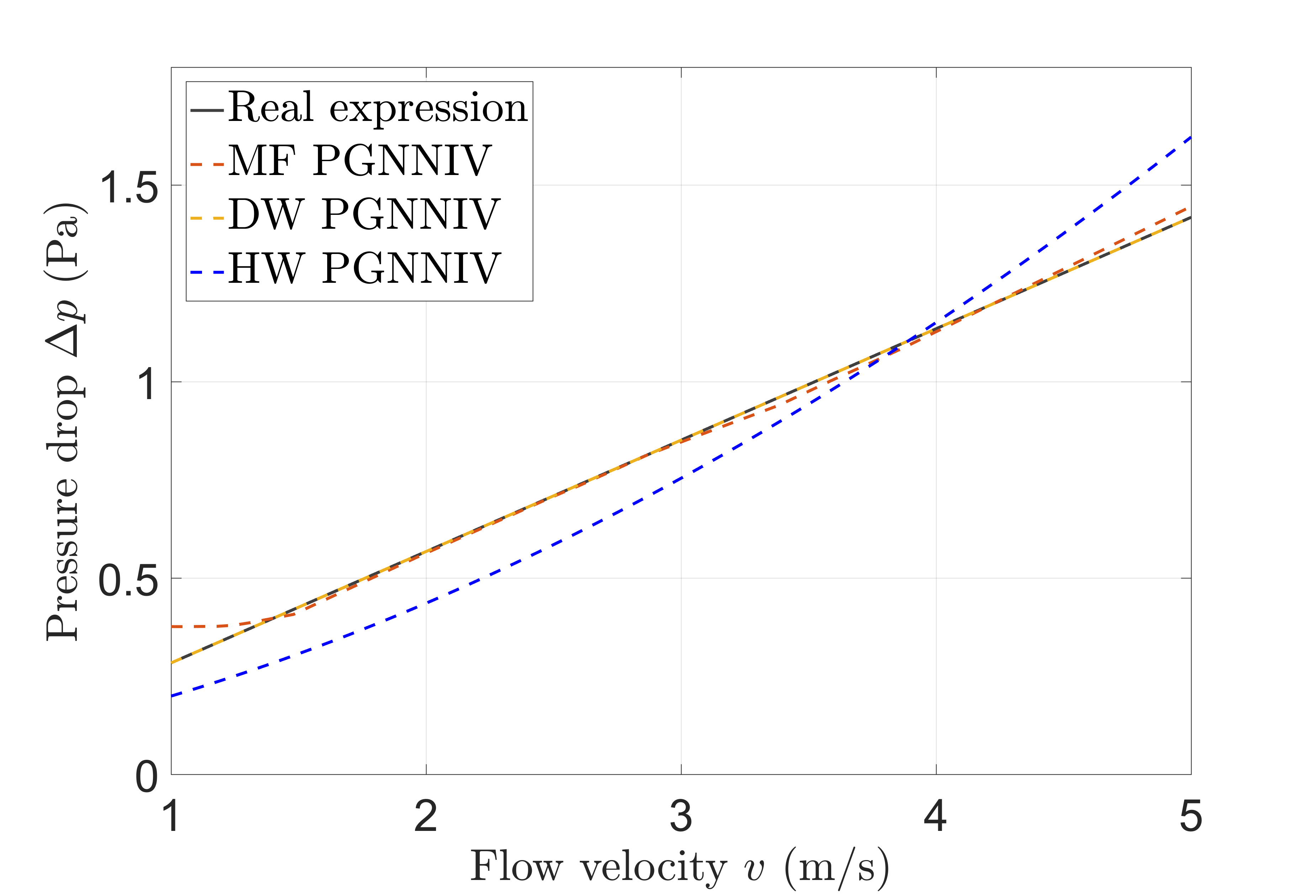}
		\caption{Darcy-Weisbach model.}
		\label{fig::model_learning_2}
		\end{subfigure}
	\caption{\textbf{Model explanatory capacity}. MF-PGNNIV has a better explanatory capacity as it is able to provide good results for data-set following different models and to discriminate between them. MB-PGNNIV give perfect fit when the model is trully the expected one, but give us worse results if not.}
	\label{fig::model_learning}
\end{figure}

For example, in the prediction problem, we have defined three PGNNIVs: a model-free (MF) PGNNIV, a Hazen-Williams model-based (MB) PGNNIV and a Darcy-Weisbach model-based (MB) PGNNIV. The presented methodology allows to discover the internal state equation, and to select the best model among several ones. In fig. \ref{fig::model_learning}, we illustrate the relationship $\Delta p = H(v)$, exported from the network after reaching convergence, for the three PGNNIVs and two data-sets, with data derived from Hazen-Williams and Darcy-Weisbach models respectively. It is clear that MF-PGNNIV gives good results for both models, although the two MB-PGNNIV are better suited for the two specific cases. MF-PGNNIV has therefore more explanatory capacity, while MB-PGNNIV have more predictive capacity for the specific considered cases. This is another illustration of the trade-off between explanatory and predictive capacity.

\begin{figure}
\centering
	\begin{subfigure}{.45\textwidth}
		\centering
		\includegraphics[width=\linewidth]{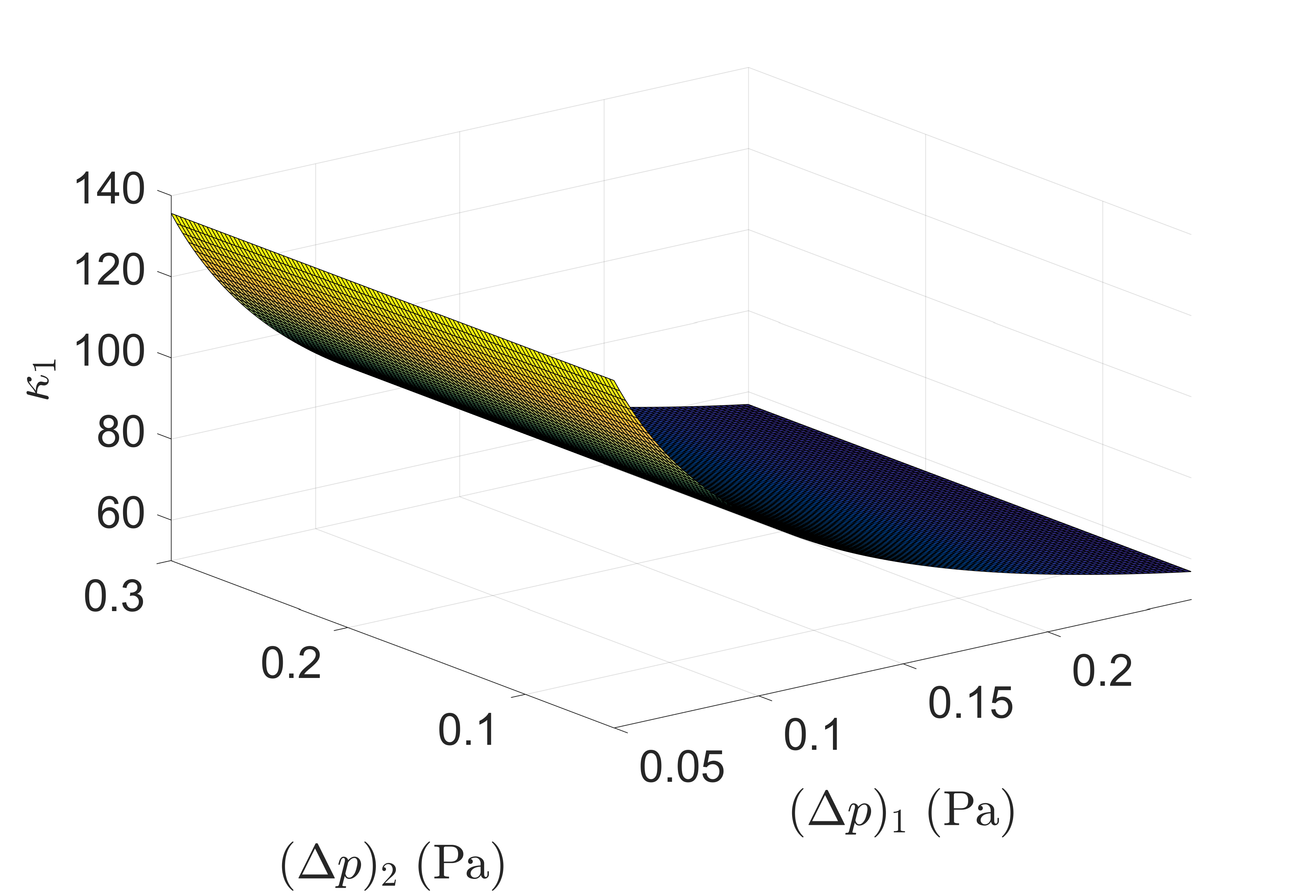}
		\caption{$\kappa_1$ real.}
		\label{fig::internal_a}
	\end{subfigure}%
	\begin{subfigure}{.45\textwidth}
		\centering
		\includegraphics[width=\linewidth]{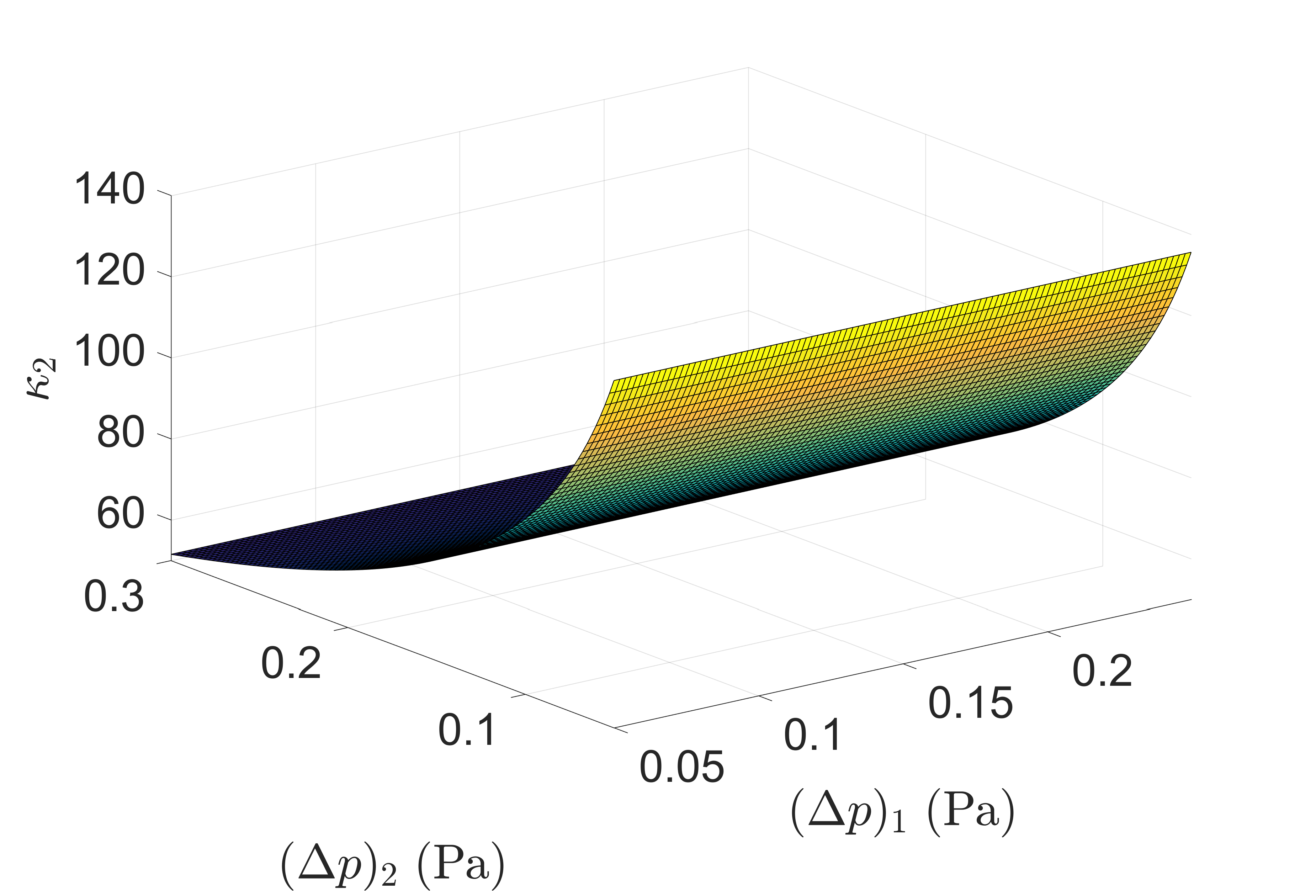}
		\caption{$\kappa_2$ real.}
		\label{fig::internal_b}
	\end{subfigure} \\
	\begin{subfigure}{.45\textwidth}
		\centering
		\includegraphics[width=\linewidth]{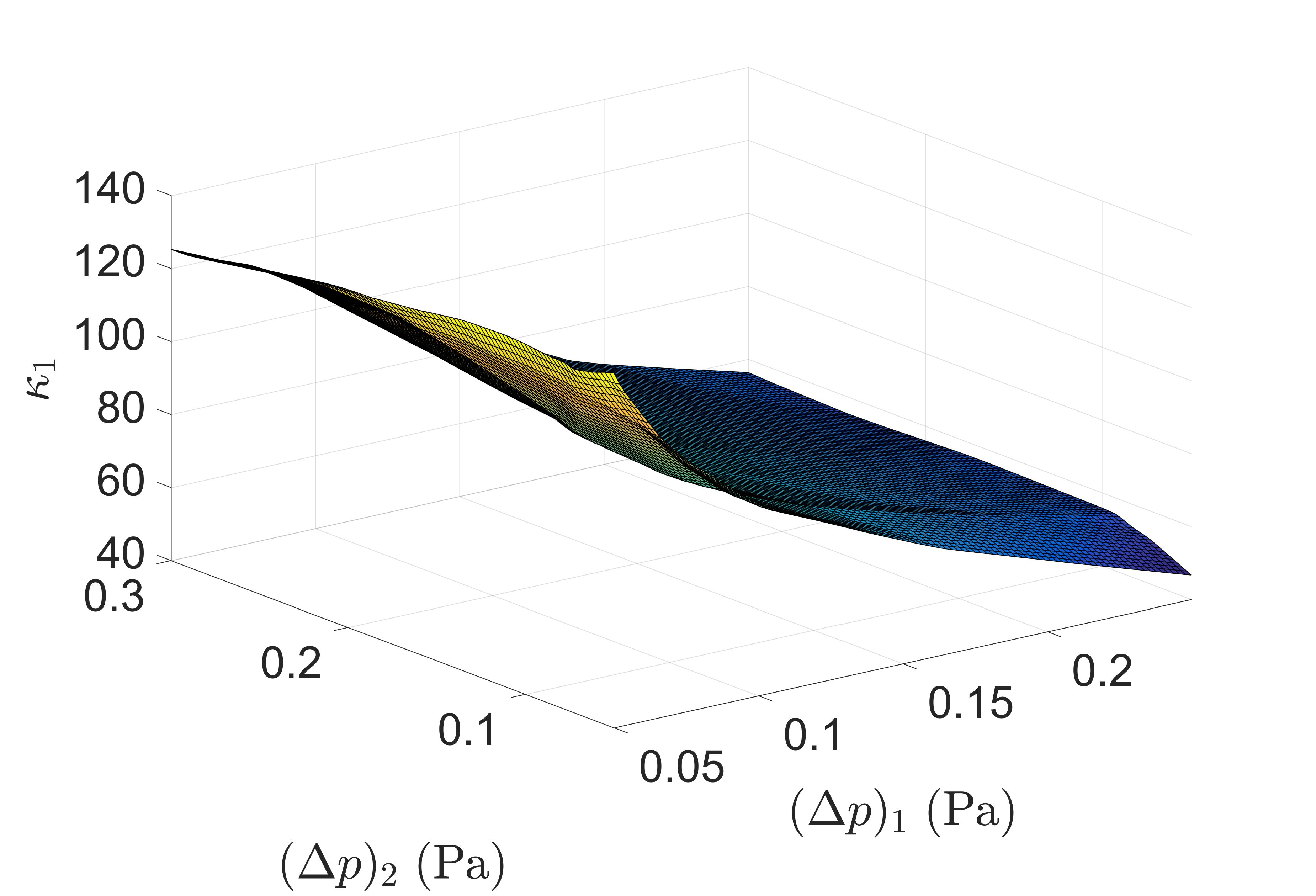}
		\caption{$\kappa_1$ predicted.}
		\label{fig::internal_c}
	\end{subfigure}
	\begin{subfigure}{0.45\textwidth}
		\centering
		\includegraphics[width=\linewidth]{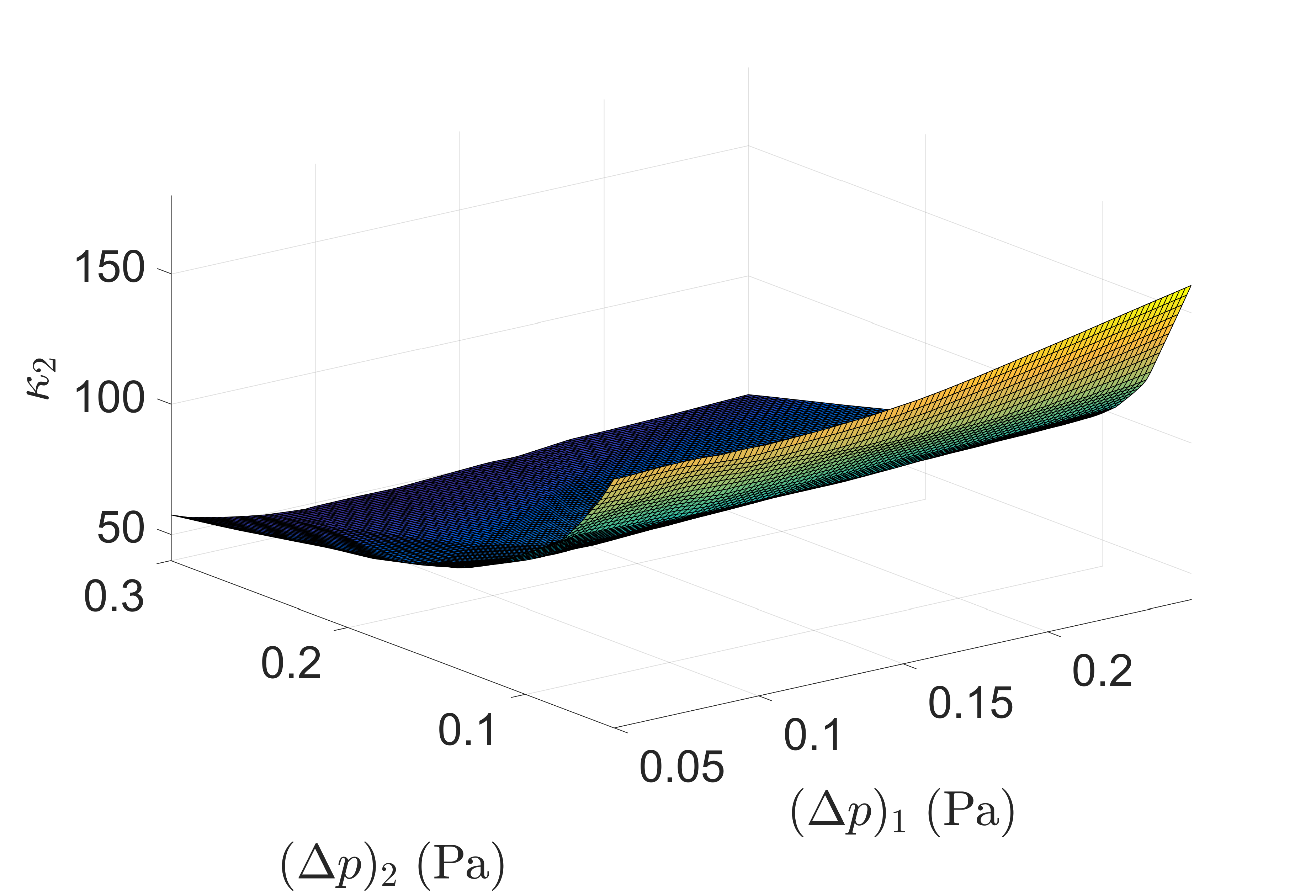}
		\caption{$\kappa_1$ predicted.}
		\label{fig::internal_d}
	\end{subfigure}
	\caption{\textbf{Model explanatory capacity}. Values of $\kappa_1$ and $\kappa_2$ predicted compared to the real ones for different values of $(\Delta p)_1$ and $(\Delta p)_2$ and $q = 3 \; \mathrm{m^3}{s}$.}
	\label{fig::internal}
\end{figure}

This network explanatory capability may be further explored in characterization also in these two ways: for accurately predicting the $(q, p_1, p_2, p_3) \rightarrow (\kappa_1, \kappa_2)$ relation and also for learning about the model separability. Fig. \ref{fig::internal} shows both the real and predicted values for $\kappa_1$ and $\kappa_2$ for $q = 3 \; \mathrm{m^3/s}$ and different values of $(\Delta p)_1$ and $(\Delta p)_2$. As explained in Section \ref{sec::characterization_results} and may be seen again in Fig. \ref{fig::internal}, the predicted values are close to the real ones, but a more important fact is that the PGNNIV, thanks to its topology, is able to separate the dependency between variables, that is $\kappa_i = H_i((\Delta p)_i)$ for $i=1,2$ instead of the general case $\boldsymbol \upkappa = \bs H(\boldsymbol \Delta p)$. Thus, some features about the model become explainable.

\subsection{Link to other methods}

In recent years, many Data-Driven methods have been applied to solve problems where some of the physics is known and other has to be discovered. PGNNIV may be compared to the different classes of methods existing in the literature. 

F. Chinesta and co-workers use Manifold Learning (ML) to establish the internal state equation $\boldsymbol \varepsilon \leftrightarrow \boldsymbol \sigma$ (\cite{lopez2016manifold,ibanez2018manifold}). In their approach, first the constitutive relationship is computed using Machine Learning techniques in the space $(\boldsymbol \varepsilon, \boldsymbol \sigma)$ (note that $\boldsymbol \sigma$ is a non-measurable variable), represented as a low-dimensional manifold. This manifold is then used, instead of the constitutive equation in the problem resolution. Our presented approach is similar in the sense that the model network $\bs H$ may be formulated using the ML framework. Indeed, methods such as kernel Principal Component Analysis (kPCA), Non-Linear Principal Component Analysis (NLPCA), Locally Linear Embedding (LLE) and t-distributed stochastic neighbor embedding (t-SNE) may be formulated in terms of appropriate weights, biases, activation functions and associated hyperparameters and network connectivity.

A very recent idea uses the GENERIC algorithm in time-dependent problems for model identification and evolution prediction \cite{gonzalez2019thermodynamically,gonzalez2019learning}. This may be seen as a particular PGNNIV, where $\bs H$ is defined using the Poisson and Dissipation operators, $\boldsymbol{L}$  and $\boldsymbol{D}$, together with the discrete version of other differential operators, if necessary. Constraints on many variables may be established, in order to ensure universal physics (the first and second laws of thermodynamics), by means of the degeneracy conditions. Combining this approach with the previous one leads to accurate solutions even while maintaining a reduced computational cost \cite{moya2019learning}.

Two other approaches have been proposed with the same \emph{model-free} philosophy. Other researchers define the constitutive manifold using interpolation instead of regression. The first one is called the \emph{What You Prescribe is What You Get} (WYPiWYG) strategy \cite{sussman2009model,latorre2014you,amores2019average} and is based on spline interpolation. The second one is based on nearest neighbor interpolation, what is totally model-free \cite{kirchdoerfer2016data}. Both strategies have demonstrated  good performance provided that we have the variables sampled at the space $(\boldsymbol \varepsilon, \boldsymbol \sigma)$. However, these two approaches suffer from extrapolating capacity if the data-set provided has not a broad enough coverage, what is faced in the PGNNIV framework by making flexible the network associated with $\bs H$.

At last and as mentioned, when $\bs H$ is defined via a parametrization of a classical model, $\bs H(\cdot) = \bs H(\cdot,\boldsymbol \lambda)$, we recover the classical fitting framework (but using neural network tools and algorithms). If, in addition, $\boldsymbol \lambda$ is completely specified and the number of weights and biases is less than or equal to the number of parameters $\boldsymbol \lambda$, PGNNIV performs merely as a dimensionality reduction.

In a certain sense, PGNNIV framework may be seen as a generalization of all the former approaches. However, only the proposed approach is able to deal with measurable variables, albeit performing the data discovering in the state space, where both measurable and non-measurable variables are present. This is possible thanks to the network constraints $\bs R$, from which the state space is built and the internal model $\bs H$, which is unknown, is learned. 

\section{Conclusions} \label{secc::Ccl}

We have presented a framework in which we use the technology and methods of Artificial Neural Networks (ANN) for solving physically-based problems. This approach allows us both to predict the evolution of a physical system and to explain its structure in the language of Physics.

Of course, it suffers from the typical pros and cons of neural networks. As pros:

\begin{itemize}
\item Once the PGNNIV is trained, it allows us to make predictions in an evaluation cost (that is, in real-time). No linear operator inversion, tangent-based operators computation, or iterative procedure is necessary when predicting the state or the evolution of a system.
\item The network is trained offline in a very time-consuming process. However, ANN is a mature and hot field in continuous development, and specific hardware and software tools (parallel, cloud and distributed computation, GPUs and TPUs software, mathematical optimization algorithms, among other software and hardware solutions) are accelerating more and more the training steps.
\end{itemize}

As main cons, we can mention the following:
\begin{itemize}
\item For Deep Neural models, the data used in the training process must be large and varied. This explains why topics such as the Internet of Things (IoT) and the Big Data paradigm are becoming so important in this context.  
\item Defining neural network successful models needs a complex and time-consuming process of network topology definition and hyperparameter tuning. Although some efforts have been made in the last years for developing appropriate tools \cite{j2008kriging,snoek2012practical,bardenet2013collaborative,maclaurin2015gradient}, this remains a hard task that strongly depends on the researcher knowledge, experience or intuition. 
\item Even if the ANN converges, it is difficult to expect a prediction to be as accurate as when using full prescribed mathematical models. PGNNIV are therefore strongly recommended for problems where qualitative explanations or major trends are searched, without entering in fine quantitative details.
\end{itemize}

In addition to these general characteristics of ANN methods, the presented hybrid formulation has shown extra advantages with respect to other existent methods:

\begin{itemize}
\item It allows working only with measurable variables. This is crucial as all the Data-Driven approaches, in one way or another, make hypotheses and assumptions about the relationships between measurable and non-measurable (internal) variables.
\item The presented method ranges from model-free (pure prediction) to model-based (explanatory) approaches. In this sense, we talk about model-guided methods.
\item As this methodology is physically guided, it allows the explainability of the different phenomena investigated, so it can be framed within the scope of the eXplainable Artificial Intelligence (XAI) \cite{samek2017explainable,dovsilovic2018explainable}.
\item It allows obtaining the whole field of internal variables, without any post-processing of the output variable. This is impossible in any other ANN method without additional assumptions.
\item As shown in the presented examples, PGNNIV has both predictive and explanatory capacity. Depending on the aim of the scientist, they can emphasize one capability or the other, depending on their interest, by easily adding/removing constraints, changing the penalty parameters or modifying the network topology. 
\item Last, but not least, PGNNIV has shown better performance than ANN in aspects such as convergence speed-up, data needs, noise filtering, and extrapolation capacity.
\end{itemize}

PGNN have just emerged in the last years, but many researchers have come to the conviction that it is the combination of physical knowledge and machine learning tools the appropriate way to adapt the Big Data paradigm to Simulation-Based Engineering and Sciences, overcoming the growing distrust of physical scientists with artificial intelligence.

\section*{Acknowledgements}
The authors gratefully acknowledge the financial support from the Spanish Ministry of Economy and Competitiveness (PGC2018-097257-B-C31 and MINECO MAT2016-76039-C4-4-R, AEI/FEDER, UE), the Government of Aragon (DGA-T24\_17R) and the Biomedical Research Networking Center in Bioengineering, Biomaterials and Nanomedicine (CIBER-BBN). CIBER-BBN is funded by the Instituto de Salud Carlos III with assistance from the European Regional Development Fund.

\bibliographystyle{unsrt} 
\bibliography{references}

\end{document}